\documentclass[10pt]{article}
\usepackage{tgpagella} 

\usepackage[left=2cm,right=2cm,top=2cm,bottom=2cm]{geometry}
\usepackage[utf8]{inputenc} 

\usepackage{definitions}

\date{}

\begin{document}

\begin{center}{{\bf \LARGE
Learning-Augmented Decentralized Online Convex Optimization in Networks}}\footnote{Pengfei Li, Jianyi Yang, and Shaolei Ren were supported in part by the NSF under grants CNS-2007115 and CCF-2324941. Adam Wierman was supported by NSF grants CCF-2326609, CNS-2146814, CPS-2136197, CNS-2106403, and NGSDI-2105648 as well as funding from the Resnick Sustainability Institute.}
\end{center}

\begin{table}[!h]
\centering
\begin{tabular}{m{0.2\textwidth} m{0.2\textwidth} m{0.2\textwidth} m{0.2\textwidth}}
\centering Pengfei Li\\ \emph{UC Riverside}
& \centering Jianyi Yang\\ \emph{UC Riverside} & \centering Adam Wierman\\ \emph{Caltech} & \centering Shaolei Ren\\
 \emph{UC Riverside} 
\end{tabular}
\end{table}

\begin{center}
    \textbf{Abstract}
\end{center}

This paper studies learning-augmented decentralized online convex optimization in
a  networked multi-agent system, a challenging setting that has remained under-explored.
We first consider a linear learning-augmented
decentralized online algorithm (\ouralglin) that
combines a machine learning (ML) policy
with a baseline expert policy in a linear manner.
We show that, while \ouralglin can exploit the potential of ML predictions to improve
the average cost performance, it cannot have guaranteed
worst-case performance. To address this limitation,
we propose a novel online algorithm (\ouralg)
that adaptively combines the ML policy and expert policy
to safeguard the ML predictions to achieve strong
competitiveness guarantees.
We also prove 
the average cost bound for \ouralg,
revealing the tradeoff between average performance
and worst-case robustness and demonstrating
the advantage of training the ML policy by explicitly considering the robustness requirement. Finally,
we run an experiment on decentralized battery management. Our results highlight the potential of ML augmentation
to improve the average performance as well as the guaranteed worst-case performance
of \ouralg.

\section{Introduction}

This paper studies the problem of decentralized online convex optimization in networks, where inter-connected agents 
must individually select actions with sequentially-revealed local online information and delayed feedback from their neighboring agents.  We consider a setting where, at each step, agents must decide on an action using local information while collectively seeking to minimize a global cost consisting of the sum of (i) the agents' node costs, which capture the local instantaneous effects of the individual actions; (ii) temporal costs, which capture the (inertia) effects of local temporal action changes; and (iii) spatial costs, which characterize the loss due to unaligned actions of two connected neighboring agents in the network.
This problem models
a wide variety of networked systems with numerous applications, such as
decentralized control in power systems \cite{decentralized_voltage_control_shahbazi2020decentralized,stability_constrained_RL_voltage_control_shi2021stability,online_distriubted_convex_optimization_networks_hosseini2016online}, 
spectrum management in multi-user wireless networks \cite{tutorial_MARL_wireless_feriani2021single,MARL_power_allocation_wireless_nasir2019multi,collaborative_MARL_wireless_yao2019collaborative}, 
multi-product pricing in revenue management \cite{multiproduct_pricing_caro2012clearance,multiproduct_pricing_candogan2012optimal}, among many others.

\revise{While centralized algorithms can effectively minimize the global cost, decentralized optimization offers advantages including resilience to single-point failure and lower computational complexity. Despite the recent progress (e.g., \cite{lin2017collab,decentralized_delay_cao2021decentralized}), online optimization in decentralized settings is inherently more challenging due to limited information availability. Agents must coordinate their actions across the network to minimize the global cost, while they usually lack complete knowledge of future costs or the actions of their neighbors. This information gap presents significant challenges for decentralized online optimization, compared to its centralized counterparts \cite{SOCO_ML_ChasingConvexBodiesFunction_Adam_COLT_2022, SOCO_OBD_R-OBD_Goel_Adam_NIPS_2019_NEURIPS2019_9f36407e}.}

To address these challenges, decentralized online convex optimization
has been studied under various settings.
For example, online algorithms for the special single-agent case \cite{OnlineOpt_Convex_LongTermConstraint_10.5555/2503308.2503322,SOCO_OBD_LQR_Abstract_Goel_Adam_Caltech_2019_10.1145/3374888.3374892,SOCO_OBD_R-OBD_Goel_Adam_NIPS_2019_NEURIPS2019_9f36407e,SOCO_Revisiting_Nanjing_NIPS_2021_zhang2021revisiting,competitive_control_memory_shi2020online,SOCO_Memory_FeedbackDelay_Nonlinear_Adam_Sigmetrics_2022_10.1145/3508037,SOCO_OBD_Niangjun_Adam_COLT_2018_DBLP:conf/colt/ChenGW18}  have been utilized as the basis for decentralized optimization to minimize the worst-case regret or competitive ratio in the multi-agent case
\cite{Decentralized_NetworkedOnlineOptimization_TSP_7131577,lin2017collab,decentralized_delay_cao2021decentralized, distributed_OCO_aggregative_li2021distributed,decentralized_OCO_relative_states_cao2021decentralized,distributed_OCO_dynamic_networks_hosseini2016online}. However, because these algorithms must make conservative decisions to mitigate potentially adversarial uncertainties, they often do not perform well in terms of the average cost. In contrast, online optimizers based on machine learning (ML) can improve the average performance by exploiting the distributional information for various problems, e.g., \cite{L2O_NewDog_OldTrick_Google_ICLR_2019,L2O_Combinatorial_Reinforcement_AAAI_2020,L2O_Scheduling_DRL_Infocom_2019_8737649,Shaolei_LearningRobustCombinatorial_Zhihui_Infocom_2022}, including multi-agent networked
systems \cite{zhang2018drl_collab,multi-agent_RL_overview_zhang2021multi,cooperative_MARL_review_oroojlooy2022review}.
But, ML-based optimizers typically lack robustness guarantees and can
result in a very high cost in the worst case 
(due to, e.g., out-of-distribution inputs),
which makes
them unsuitable for mission-critical applications.

The field of learning-augmented algorithms has emerged in recent years with the goal of
providing ``best of both worlds'' guarantees: 
\revise{near-optimal performance with accurate ML predictions and guaranteed robustness with inaccurate predictions. These algorithms have demonstrated success in various online settings e.g., \cite{Shaolei_SOCO_RobustLearning_OnlineOpt_MemoryCosts_Infocom_2023,SOCO_ML_ChasingConvexBodiesFunction_Adam_COLT_2022,OnlineOpt_Learning_Augmented_RobustnessConsistency_NIPS_2020,OnlineOpt_ML_Adivce_Survey_2016_10.1145/2993749.2993766,primal_dual_learning_augmented_bamas2020primal,OnlineOpt_ML_Advice_CompetitiveCache_Google_JACM_2021_10.1145/3447579}. However, existing learning-augmented algorithms \cite{OnlineOpt_Learning_Augmented_RobustnessConsistency_NIPS_2020, SOCO_OnlineOpt_UnreliablePrediction_Adam_Sigmetrics_2023_10.1145/3579442, SOCO_ML_ChasingConvexBodiesFunction_Adam_COLT_2022} primarily focus on centralized scenarios, making their adaptation to decentralized setups technically challenging due to spatial uncertainties arising from limited information availability. Moreover, these algorithms predominantly focus on worst-case performance guarantees, with less emphasis on average cost performance.
}

\textbf{Contributions.} 
We study the challenging and under-explored setting of decentralized online optimization in networks. 
We first consider a linear learning-augmented decentralized
online optimization algorithm (\ouralglinear), which
linearly combines a potentially untrusted ML policy with a trusted
baseline policy (called ``expert''). We show that \ouralglinear can
exploit the power of ML predictions to improve the average cost performance when the ML predictions are of sufficiently high quality, but it \emph{cannot} offer guaranteed competitiveness or robustness in the worst case
when ML predictions are of arbitrarily low quality. 

To overcome \ouralglinear's lack of guaranteed competitiveness, we introduce and analyze a novel algorithm, \ouralg, that adaptively combines an ML policy with an expert policy based on the actual online costs. The key idea behind \ouralg is to leverage the baseline expert policy to safeguard online actions to avoid too greedily following ML predictions that may not be robust. \revise{In a decentralized setting, the primary design challenge is managing spatial information inefficiency, as agents lack prior knowledge of their neighbors' actions. 
More concretely, the spatial cost is dependent on the actions of neighboring agents, making it difficult for a single agent to evaluate it in isolation.
To address this and associated spatial cost uncertainties, we propose a novel spatial cost decomposition that adaptively splits the shared spatial cost between connected agents, enabling each agent to safeguard its own actions based on local information.
To ensure non-empty action sets and maintain robustness when deviating from the trusted expert policy, we also introduce temporal reservation costs to address worst-case future cost uncertainties.}

Our main results provide 
 worst-case and average cost bounds for \ouralg (see Theorems~\ref{thm:robustness},~\ref{thm:avg_cost_blackbox}, and~\ref{thm:avg_cost_e2e}). 
We also show the worst-case robustness and consistency
of \ouralglinear and \ouralg (see Corollary~\ref{coro:robust_consistency}
and Theorem~\ref{theorem:robust_consistency_lado}).
 Importantly, unlike most of the prior work
 that assumes a black-box ML model trained as a standalone optimizer,
 our results also provide average cost bounds for 
 training ML by explicitly considering how the ML policy will be used. Our results quantify the improvement obtained by explicitly accounting
for the robustness step in ML training.

\revise{
To evaluate the effectiveness of \ouralglinear and \ouralg, we conduct experiments on decentralized battery management for sustainable data centers, 
with a networked battery system
of up to {120 nodes}. Our results demonstrate the empirical benefits of our algorithms over existing baselines across various network topologies. Both \ouralglinear and \ouralg, when augmented with ML, consistently achieve strong average cost performance under various network topologies. Moreover,  \ouralg 
offers guaranteed robustness even when the ML predictions have low quality.
}

To summarize, the main contributions of our work are as follows.
\revise{First, unlike existing learning-augmented algorithms, we focus on the more challenging setting of decentralized optimization, where agents make online decisions with delayed information about their neighbors' actions.}
Second, to guarantee worst-case robustness of \ouralg against
a given policy in our decentralized setting,
our design of robust action sets includes novel adaptive spatial cost splitting, which is a novel technique and differs from the design in
a centralized setting.
Last but not least, 
we rigorously analyze and also empirically show the performance of 
\ouralglinear and \ouralg in terms of their average-case and worst-case costs.

\section{Problem Formulation} \label{sec:problem}

We study the setting introduced in \cite{Decentralized_SOCO_YihengLin_Adam_ICML_2022_pmlr-v162-lin22c}
(where there is no ML policy augmentation)
and consider decentralized online convex optimization in a network with $V=|\mathcal{V}|$ agents/nodes belonging to the set $\mathcal{V}$. If two agents have interactions with each other, there exists an edge between them.  Thus, the networked system can be represented by an undirected graph $ ( \mathcal{V}, \mathcal{E})$, with $\mathcal{E}$ being the set of edges. 
Each problem instance (a.k.a. episode) consists
of $T$ sequential time steps.

At step $t=1,\cdots,T$, 
 each agent $v$ selects an irrevocable action $x_t^v\in\mathbb{R}^n$.  We denote $x_t=[x_t^1,\cdots, x_t^V]$ as the action vector for all agents at step $t$, where
the superscript $v$ represents the agent index whenever applicable. After
$x_t$ is selected 
for step $t$, the network generates a global cost $g_t(x_t)$ which consists of the following three parts. 

$\bullet$ \textbf{Node cost $f_t^v(x_t^v)$:} Each individual agent incurs
a node cost  $f_t^v(x_t^v)$, which only relies on the action of a single agent $v$ at step $t$ and measures the effect of the agent's decision on itself.

$\bullet$  \textbf{Temporal cost $c_t^v(x_t^v,x_{t-1}^v)$:} It couples the two temporal-adjacent actions of a single agent $v$ and represents the effect of temporal interactions to smooth actions over time.

$\bullet$ \textbf{Spatial cost $s^{(v,u)}_t(x_t^v,x_t^u)$:} 
It is incurred if an edge exists between two agents $v$ and $u$,
capturing the 
 loss due to unaligned actions of two connected agents.

This formulation applies directly to many real-world applications \cite{Decentralized_SOCO_YihengLin_Adam_ICML_2022_pmlr-v162-lin22c}. For example, in geo-distributed cloud resource management,
each data center is an agent whose server provisioning decision (i.e.,
the number of on/off servers) incurs a node cost that captures its local operational cost \cite{SOCO_DynamicRightSizing_Adam_Infocom_2011_LinWiermanAndrewThereska}. The temporal cost penalizes frequent servers on/off to avoid excessive wear-and-tear (a.k.a., switching costs) \cite{SOCO_DynamicRightSizing_Adam_Infocom_2011_LinWiermanAndrewThereska}.
Meanwhile, each data center's decision 
 results in an environmental footprint
(e.g., carbon emission and water consumption)  \cite{Shaolei_Water_SpatioTemporal_GLB_TCC_2018_7420641}. Thus, the added spatial cost mitigates
inequitable environmental impacts in different locations to achieve environmental
justice, which is a crucial consideration in many corporates' 
Environmental, Social, and Governance (ESG) strategies \cite{Facebook_SustainabilityReport_2021}.
In Section~\ref{sec:application_example}, we provide more modeling details, and  explore two other applications: decentralized battery management for sustainable computing and multi-product dynamic pricing.

Next, we make the following common assumptions for online optimization, e.g.,  \cite{goel2019beyond,Decentralized_SOCO_YihengLin_Adam_ICML_2022_pmlr-v162-lin22c}.
\begin{assumption}\label{assumption:node_cost}
    The node cost $f_t^v$:  $\mathbb{R}^n \to \mathbb{R}_{\geq 0}$ is $\beta$-strongly convex and $\ell_f$-smooth.
\end{assumption}
\vspace{-.1in}
\begin{assumption}
    The temporal interaction cost $c_t^v$:  $\mathbb{R}^n \times \mathbb{R}^n \to \mathbb{R}_{\geq 0}$ is convex and $\ell_T$-smooth.
\end{assumption}
\vspace{-.1in}
\begin{assumption}\label{assumption:spatial_cost}
    The spatial interaction cost $s_t^{v,u}$:  $\mathbb{R}^n \times \mathbb{R}^n \to \mathbb{R}_{\geq 0}$ is convex and $\ell_S$-smooth.
\end{assumption}
The convexity assumption is needed for analysis,
while smoothness (i.e., Lipschitz-continuous gradients) ensures that the costs will not vary unboundedly when the actions change \cite{Decentralized_SOCO_YihengLin_Adam_ICML_2022_pmlr-v162-lin22c}.

The networked agents collaboratively minimize the total global cost over 
$T$ time steps defined as: 
\begin{equation}\label{eqn:cost_objective}
\nonumber
\begin{aligned}
   {cost}&(x_{1:T})= \sum_{t=1}^Tg_t(x_t)=\sum_{t=1}^T   \sum_{v \in \mathcal{V}} f_t^v(x_t^v)+
    \sum_{t=1}^T   \sum_{v \in \mathcal{V}}c_t^v(x_t^v, x_{t-1}^v)  +
     \sum_{t=1}^T \sum_{(v, u) \in \mathcal{E}} s_t^{(v, u)}(x_t^v, x_t^u), 
\end{aligned}
\end{equation}
where $g_t(x_t)=\sum_{v\in\mathcal{V}}f_t^v(x_t^v)+\sum_{v\in\mathcal{V}}c_t^v(x_t^v,x_{t-1}^v)+\sum_{(v,u)\in\mathcal{E}}s^{(v,u)}_t(x_t^v,x_t^u)$
is the total cost at time $t$.
In practice, \revise{we consider a weighted sum of the node, temporal, and spatial costs. These weights assigned to each cost component reflect their relative importance in the overall cost metric. For the convenience of presentation, we normalize the weight of the node cost to 1 and incorporate the remaining weights directly into the individual cost terms.}
With a slight abuse of notation, we also denote $g_t=\{f_t^v,c_t^v,s_t^{(u,v)}, v\in\mathcal{V}, (u,v)\in\mathcal{E}\}$ as the cost function information for step $t$, and $g_{1:T}=[g_1,\cdots, g_T]\in\mathcal{G}$ as all the exogenously-determined information
for the entire problem instance where
 $\mathcal{G}$ is the set of all possible $g_{1:T}$.

Our goal is to find a decentralized learning-augmented online policy $\pi_v$ for each agent $v$
that maps the local available information
(to be specified in Section~\ref{sec:local_information})
to its action $x_t^v$
at time $t$.
For notational convenience,
we also denote $\pi=[\pi_1, \cdots,\pi_V]$ as the combined
policy for the network. 

\subsection{Performance Metrics}\label{sec:model_metric}


We consider the following two performance metrics --- average cost and $\lambda$-competitiveness.  

\begin{definition}[Average cost]\label{def:avg} 
Given a decentralized online policy $\pi=[\pi_1,\cdots,\pi_V]$,
the average cost is $AVG(\pi)=\mathbb{E}_{g_{1:T}}\left[cost(\pi,g_{1:T} )\right]$,
where the  information $g_{1:T}$ follows a distribution $\mathcal{P}_{g_{1:T}}$.
\end{definition}

\begin{definition}[$\lambda$-competitive to $\pi^{\dagger}$]\label{def:robustness} 
For $\lambda>0$, an online policy $\pi=[\pi_1,\cdots,\pi_V]$
is  $\lambda$-competitive against a  baseline policy
$\pi^{\dagger}$ if
 $cost(\pi,g_{1:T} )\leq (1+\lambda) cost(\pi^{\dagger},g_{1:T} ))$ holds
for any 
$g_{1:T}\in\mathcal{G}$.
\end{definition}

The average cost measures the decision quality of the decentralized
policy $\pi$ in typical cases, whereas the 
$\lambda$-competitiveness shows the worst-case competitiveness
in terms of the cost ratio of 
the global cost of $\pi$ to a given trusted baseline policy 
$\pi^{\dagger}$ (which is also referred to as an \emph{expert} policy). Our definition of 
$\lambda$-competitiveness against $\pi^{\dagger}$
is both general and common in the literature on learning-augmented online algorithms as well as online control \cite{SOCO_ML_ChasingConvexBodiesFunction_Adam_COLT_2022,Control_Regularization_Reduced_Prior_Yisong_ICML_2019_pmlr-v97-cheng19a,Yisong_SmoothImitation_ICML_2016_10.5555/3045390.3045463}, where
 competitiveness is defined against a given baseline policy $\pi^{\dagger}$
\cite{SOCO_ML_ChasingConvexBodiesFunction_Adam_COLT_2022,Shaolei_Learning_OBM_EdgeWeighted_ICML_2023,Shaolei_SOCO_RobustLearning_OnlineOpt_MemoryCosts_Infocom_2023}.
Importantly, for our problem, there exist various expert policies $\pi^{\dagger}$ (e.g.,
localized prediction control \cite{Decentralized_SOCO_YihengLin_Adam_ICML_2022_pmlr-v162-lin22c})
with bounded cost ratios against the oracle policy $OPT=\pi^*$ that minimizes the global cost with all offline information. As a result, by considering an expert policy
with a competitive ratio of $\rho_{\pi^{\dagger}}$
, our policy $\pi$ is also competitive against the optimal oracle, i.e.,
$cost(\pi,g_{1:T} )\leq \rho_{\pi^{\dagger}}(1+\lambda) cost(OPT,g_{1:T} ))$
for any $g_{1:T}\in\mathcal{G}$.
Alternatively, 
the expert policy $\pi^{\dagger}$ can be viewed
as a policy prior currently in use \cite{Conservative_RL_Bandits_LiweiWang_SimonDu_ICLR_2022_yang2022a}, while the new learning-augmented policy
$\pi$ must no worse than  $(1+\lambda)$-times the policy prior in terms
of the cost for any problem instance.

The average cost and worst-case competitiveness metrics are different and complementary to
each other \cite{OnlineOpt_ML_Augmented_RobustCache_Google_ICML_2021_pmlr-v139-chledowski21a,SOCO_ML_ChasingConvexBodiesFunction_Adam_COLT_2022}. 
 Here,
 we take a competitiveness-constrained approach.
 Specifically,
given both an ML-based optimizer and an expert algorithm as advice,
we aim to find a learning-augmented policy $\pi=[\pi_1,\cdots,\pi_V]$
to minimize the average cost subject to the $\lambda$-competitiveness constraint:
\begin{equation}\label{eqn:objective_function}
        \min_{\pi} \mathbb{E}_{g_{1:T}}\left[cost(\pi,g_{1:T} )\right],\;
        s.t., \; cost(\pi,g_{1:T} )\leq (1+\lambda) cost(\pi^{\dagger},g_{1:T}), \;  \forall g_{1:T}\in\mathcal{G}. 
\end{equation}

While offline-trained ML-based policies (e.g., based on 
multi-agent reinforcement learning  \cite{zhang2018drl_collab,multi-agent_RL_overview_zhang2021multi,cooperative_MARL_review_oroojlooy2022review})
can potentially minimize the average cost, they may not satisfy
 $\lambda$-competitiveness in the worst case.
In fact, it is well-known that
due to the statistical nature,  ML-based policies can have 
arbitrarily bad performance in certain (possibly rare) cases,
especially when the testing problem instance is very distinct from those 
training instances \cite{SOCO_OnlineOpt_UnreliablePrediction_Adam_Sigmetrics_2023_10.1145/3579442}. 
Thus, we adopt a learning-augmented approach where we 
integrate an ML-based policy into decision-making
while using a trusted expert policy to safeguard our online decisions.

\subsection{Application Examples}\label{sec:application_example}
To make our model concrete, we present the following application examples.
 Readers are also referred to \cite{Decentralized_SOCO_YihengLin_Adam_ICML_2022_pmlr-v162-lin22c} for additional examples.

\textbf{Decentralized battery management for sustainable computing.}
 Traditionally, data centers rely on fossil fuels such as coal or natural gas to power their operation. Thus, with the proliferating demand for  cloud computing and artificial intelligence services, there have been increasing environmental
concerns with data centers' growing carbon emissions.
 As such, it is important to find ways to reduce data centers' carbon footprint and mitigate their environmental impact --- decarbonizing data centers. While renewable energy sources, such as solar and wind, are natural alternatives for sustainable data centers, their availability can be highly fluctuating
 subject to weather conditions, thus imposing significant challenges to meet data centers' energy demands. 
Consequently, large energy storage consisting of multiple battery units has become essential to leverage intermittent renewable energy to power data centers for sustainable computing. Nonetheless, it is challenging
to manage a large energy storage system to achieve optimal efficiency.
Specifically, while each battery unit is responsible for its own charging/discharging decisions 
to keep the energy level within a desired range (e.g., 20-80\%) in decentralized battery management, the state-of-charge (SoC) levels
across different battery units should also be maintained as uniform as possible to extend the overall battery lifespan and energy efficiency 
\cite{Battery_UniformSOC_IEEE_TransSustainableEnergy_2016_7305808}.
This problem can be well captured by converting a canonical form into
our model:
each battery unit decides its SoC level by charging/discharging
and  incurs a node cost (i.e., SoC level deviating from
the desired range) and a temporal cost (i.e., SoC changes due to charging/discharging), and meanwhile there is a spatial cost due
to SoC differences across different battery units.

More concretely, we consider an energy storage system that includes a set of battery units $\mathcal{V}$ interconnected through physical connections $\mathcal{E}$. 
For a battery unit $v \in \mathcal{V}$, the goal is to minimize the difference between the current SoC and a nominal value $\bar{x}_{v}$ plus a power grid's usage cost, which can be defined as a local objective: $\min_{u_{v, 1:T}} \sum_{t=1}^{T} \|x_{v,t} - \bar{x}_v \|^2 + \sum_{t=1}^{T}b\| \xi_{v,t} \|^2$, where $\xi_{v,t}$ is the charging/discharging schedule from the power grid
(i.e., $\xi_{v,t}>0$ means drawing energy from the grid and $\xi_{v,t}<0$ means returning
energy to the grid)  and $b$ is the power grid's usage penalty cost. The time index for the first term starts at
$t=2$ as we assume a given initial state $x_{v,1}$ (i.e., the SoC cost at $t=1$ is already given). 
The canonical form of the battery SoC dynamics follows by $x_{v,t} = A_v x_{v,t-1} + B_v \xi_{v,t} + C_v w_{v,t}$, 
where $A_v$ denotes the self-degradation coefficient, 
$B_v$ denotes the charging efficiency, $w_{v,t}$ is the data center's net energy demand 
from battery unit $v$ (i.e., $w_{v,t}>0$ means the data center's energy
demand exceeds the available renewables and $w_{v,t}<0$ otherwise), $C_v$ denotes the conversion coefficient (inversely proportional to the capacity of battery unit $v$), which translates the net energy demand to the change in battery SoC.

Based on the physical connection $(u,v) \in \mathcal{E}$, the SoC difference 
between battery units $u$ and $v$ can lead to reduced performance and lifespan. 
For instance, the battery voltage difference caused by different SoCs may cause overheating problems or even battery damage \cite{Battery_UniformSOC_IEEE_TransSustainableEnergy_2016_7305808}. 
Thus, to penalize the SoC difference between two interconnected battery units,
we add a spatial cost $\sum_{(v, u) \in \mathcal{E}} c\cdot \| x_t^v - x_t^u\|^2$, where $c$ is the SoC difference penalty coefficient. Thus, the total control cost  is 
\begin{equation}\label{eqn:total_cost_battery_original}
\begin{aligned}
   \min_{ \{u_{v, 1:T}, \forall v \in \mathcal{V}\}  } \sum_{t=1}^{T} \sum_{v \in \mathcal{V}}  \|x_{v,t} - \bar{x}_v \|^2 + \sum_{t=1}^{T} \sum_{v \in \mathcal{V}}  b\| \xi_{v,t} \|^2 + \sum_{t=1}^{T}\sum_{(v, u) \in \mathcal{E}} c \| x_t^v - x_t^u\|^2.
\end{aligned}
\end{equation}

Next, we convert \eqref{eqn:total_cost_battery_original} into
our formulation decentralized online convex optimization.
At time $t$, we define  $y_{v,t} = \bar{x}_v - A^t_v x_{v,1} - \sum_{i=1}^t A_v^{t-i}C_v w_{v,i}$ as the context parameter determined by all the previous states and online inputs, and $a_{v,t} = \sum_{i=1}^t A^{t-i}_v B_v \xi_{v,i}$ as the corresponding node $v$'s online action in our model. 
Then, we define the node cost for $v$  as $f_{t}^v(a_{v,t}) = \| a_{v,t} - y_{v,t}\|^2 = \|x_{v,t} - \bar{x}_v\|^2$, the temporal cost for $v$ as $c_{t}^v(a_{v,t}, a_{v,t-1}) = \frac{b}{B_v^2}\| a_{v,t} - A_v a_{v,t-1} \|^2 = b\| \xi_{v,t} \|^2$, and the spatial cost for edge $(v,u)$ as $s_{t}^{(v,u)} = c\|  (a_{v,t} - a_{u,t}) - (y_{v,t} - y_{u,t}) + \bar{x}_v - \bar{x}_u\|^2 = c \| x_t^v - x_t^u\|^2$. By combining these three costs together, the total global cost becomes
\begin{equation}
\begin{aligned}\label{eqn:decen_objective}
   \min_{ \{a_{v, 1:T}, \forall v \in \mathcal{V} \} }  \sum_{t=1}^{T}   \biggl( &\sum_{v \in \mathcal{V}} \| a_{v,t} - y_{v,t}\|^2 +  \sum_{v \in \mathcal{V}} \frac{b}{B_v^2}\| a_{v,t} - A_v a_{v,t-1} \|^2 +   \\
   &\sum_{(v, u) \in \mathcal{E}} c\|(a_{v,t} - a_{u,t}) - (y_{v,t} - y_{u,t}) + \bar{x}_v - \bar{x}_u\|^2 \biggr), 
\end{aligned}
\end{equation}
which has the same form as our formulation \eqref{eqn:cost_objective}
if we view $a_{v,t}$ as node $v$'s online action at time $t$.

\textbf{Geographic server provisioning with environmental equity.}
Online service providers commonly rely on geographically distributed data
centers in the proximity of end users to minimize service latency. 
Nonetheless, data centers are notoriously energy-intensive.
Thus, given time-varying workload demands, the data center capacity (i.e., the number of active servers)
needs to be dynamically adjusted 
to achieve energy-proportional computing
 and minimize the operational cost
\cite{SOCO_DynamicRightSizing_Adam_Infocom_2011_LinWiermanAndrewThereska}.
More specifically, each data center  dynamically provisions its servers 
in a decentralized manner, based on which 
the incoming workloads
are scheduled \cite{Google_CarbonAwareComputing_PowerSystems_2023_9770383}.
Naturally, turning on more servers in a data center can provide better service quality
in general, but it also consumes more energy and hence negatively results in a higher environmental footprint (e.g., carbon  and water, which both
roughly increase with the energy consumption proportionally
\cite{Facebook_SustainabilityReport_2021,Shaolei_Water_SpatioTemporal_GLB_TCC_2018_7420641}).
 
While it is important to reduce the total environmental footprint across
 geo-distributed data centers, addressing
 environmental \emph{inequity} --- mitigating
 locational disparity in terms of negative environmental consequences 
 caused by data center operation
 --- is also crucial
 as inequity can create significant business risks and unintended societal impacts 
\cite{Shaolei_Equity_GLB_Environmental_AI_eEnergy_2024}.
Indeed, the emergence of data centers' environmental inequity
has been recently compared to ``historical practices of settler colonialism and racial capitalism'' \cite{Justice_AINowInstitute_ConfrontingTechPower_2023} and calls for attention
from various environmental groups and policy think tanks
\cite{Justice_AI_EnvironmentalEquity_Think20_Policy_2023,Justice_Policy_Ethical_AI_Recommendation_UNESCO_2022}.
 
To address environmental inequity, we view
each data center 
as a node $v$ in our model. The data center $v$ makes its own dynamic server provisioning decision $x_t^v$ (i.e., the number of active servers, which can be treated as a continuous variable
due to tens of thousands of servers in data centers), and incurs a node cost $f_t^v(x_t^v)$ that captures
the local energy cost, environmental footprint, and service quality \cite{SOCO_DynamicRightSizing_Adam_Infocom_2011_LinWiermanAndrewThereska}.
The temporal cost $c_t^v(x_t^v, x_{t-1}^v)=\|x_t^v-x_{t-1}^v\|^2$ captures the negative impact of switching servers on and
off (e.g., wear-and-tear), which is also referred to as the switching cost in the data center literature \cite{SOCO_DynamicRightSizing_Adam_Infocom_2011_LinWiermanAndrewThereska}.
Additionally, the spatial cost $s_t^{(v, u)}(x_t^v, x_t^u)$
can be written as $s_t^{(v, u)}(x_t^v, x_t^u)=\|e_{t}^vx_t^v-e_{t}^ux_t^u\|^2$ where $e_{t}^v$ is the weighted environmental ``price''
 (e.g., water usage efficiency scaled
 by the average per-server energy)
in data center $v$. Thus, the spatial cost
addresses environmental justice concerns by penalizing difference between
data center $v$ and data center $u$ in terms of their environmental footprint. As a result, by considering weighted
sums of the node costs, temporal costs, and spatial costs,
our model applies to
the problem of geographic server provisioning with environmental justice, which
is emerging as a critical concern in the wake of increasingly hyperscale
data centers that may 
leave certain local communities to disproportionately bear the negative environmental consequences.

\textbf{Multi-product dynamic pricing.}
As digital marketplaces continue to grow, offering a diverse range of products or services becomes inevitable for businesses that seek to cater to diverse consumer preferences. 
As such, 
 a dynamic multi-product pricing policy is vital for revenue management. 
For instance, the online platform may modify product prices multiple times within a single day, considering the estimated user demand, competitor prices and inventory dynamics.
Nevertheless, due to the intricate relationships products share within the marketplace, it is a challenging task to set prices. For example, 
for complementary products (e.g. laptop vs headphones), a special offer on a certain laptop may stimulate the demand for headphones and other accessories. 
Additionally, customers may also observe historical prices, which affects their willingness to buy. In other words, the current demand for a certain product can be temporally coupled with the previous prices \cite{Decentralized_SOCO_YihengLin_Adam_ICML_2022_pmlr-v162-lin22c}. 
In our framework, the temporal interaction and spatial interaction costs model the effects of multiple product relationships and user behaviors, respectively. More specifically, at time $t$, suppose that the price of product $v$ is $x_t^v$. Then, under a linear demand model, the total revenue is represented by
\begin{equation}
    \sum_{t=1}^T \biggl[ \sum_{v\in \cV}x_t^v(a_t^v - k_t^v x_t^v) + \sum_{v\in \cV}x_t^v(b_t^v x_{t-1}^v)  + \sum_{(u,v) \in \cE} x_t^v ( \xi^{(u,v)}_t x_t^{u} )  \biggr]
\end{equation}
where $a_t^v - k_t^v x_t^v$ models the nominal demand under price $x_t$, $b_t^v$ quantifies the 
effect of the previous price, the coefficient $\xi^{(u,v)}_t$ denotes the spatial relationships between a product pair $(u,v)$. 
Under realistic parameter settings, this problem can be converted into our model
of decentralized online convex optimization  (see \cite{Decentralized_SOCO_YihengLin_Adam_ICML_2022_pmlr-v162-lin22c} for details).

\section{\ouralglinear: Linearly Combining ML Advice and Expert Advice}\label{sec:lado-linear}

To begin, we study a simple approach toward designing a learning-augmented algorithm, which uses a fixed linear combination of ML-based untrusted policy and the trusted expert policy, i.e., Linear Learning-Augmented Decentralized Online Optimization (\ouralglinear). We analyze the performance of \ouralglinear
and highlight its key limitation: the lack of guaranteed worst-case competitiveness.

\subsection{Local Information Availability}\label{sec:local_information}

Our goal is to effectively use both ML-based advice and expert advice
to solve \eqref{eqn:objective_function} in a decentralized online manner.
In our setting,
each agent $v\in\mathcal{V}$ has access to a decentralized online ML policy
$\tilde{\pi}_v$ and a decentralized online expert policy $\pi^{\dagger}_v$,
which produce actions $\tilde{x}_t^{v}$
and $x_t^{v,\dagger}$ at time $t=1,\cdots,T$, respectively, based on local online information. 
Then, given $\tilde{x}_t^{v}$ and $x_t^{v,\dagger}$,
the agent $v$ chooses its actual action $x_t^v$ using \ouralglinear.

More specifically, 
the following online information is revealed
to each agent $v$ at step $t$:
node cost function $f^v_{t}$, temporal cost 
function $c^v_{t}$, spatial cost function $s_{t-1}^{(v,u)}$, 
connected agents' actions $x_{t-1}^u$ 
and their corresponding expert actions $x_{t-1}^{u,\dagger}$ for  $(v,u)\in\mathcal{E}$.
That is, at the beginning of step $t$, 
each agent $v$ receives its own node cost
and temporal cost functions for time $t$, and also 
the spatial cost along with the actual/expert actions
from the neighboring agents connected to agent $v$ for time $t-1$.
Thus, before choosing an action at time $t$,
all the local information available to agent $v$ can be summarized  as
\begin{equation}\label{eqn:information_set}
   I^v_t=\{f^v_{1:t},c^v_{1:t},s_{1:t-1}^{(v,u)},
x_{1:t-1}^u, x_{1:t-1}^{u,\dagger},
Z^v_t, (v,u)\in\mathcal{E}\},
\end{equation}
 where $Z^v_t$ captures the other applicable information (e.g.,
agent $v$'s own actual/ML/expert actions in
the past). Moreover, knowledge of cost functions over the next $k$ temporal steps
and/or $r$-hop agents in the network
can further improve the competitiveness of expert policies \cite{Decentralized_SOCO_YihengLin_Adam_ICML_2022_pmlr-v162-lin22c} and, if available, be included in $Z^v_t$. Without
loss of generality, we use $I^v_t$ as the locally available information for agent $v$ at time $t$.
Additionally, the smoothness parameters
$\ell_f$, $\ell_c$, and $\ell_s$ and robustness parameter $\lambda$ are known to the agents as shared information.

Our information availability setting is in line with that considered by the prior literature on decentralized
online convex optimization \cite{Decentralized_SOCO_YihengLin_Adam_ICML_2022_pmlr-v162-lin22c}, except that each agent has access to both the ML advice
and expert advice in our setting. Note also that  there is a 
\emph{separate} line of research of online convex optimization that assumes the node cost function
$f^v_{t}$ is only revealed to the agent at the end of time $t$ \cite{OnlineConvexOpt_Book_2016_hazan2016introduction}, but they 
often have different design goals (e.g., sublinear regret compared
to a \emph{static} baseline policy) than our worst-case competitiveness guarantees against a \emph{dynamic} baseline policy specified in Definition~\ref{def:robustness}.

Most importantly, unlike in a centralized setting,
an agent $v$ must
individually choose its irrevocable action $x_t^v$ on its own
--- it cannot communicate its action $x_t^v$ or its expert action $x_t^{v,\dagger}$ to its connected
agent $u$ until the next time step $t+1$.  
The one-step delayed feedback of
the spatial costs and  the actual/expert actions
from the connected agents 
is commonly studied in decentralized online convex optimization \cite{Decentralized_SOCO_YihengLin_Adam_ICML_2022_pmlr-v162-lin22c}
and crucially differentiates our work
 from the prior \emph{centralized} learning-augmented algorithms, adding
challenges for ensuring the satisfaction of the $\lambda$-competitiveness requirement.

\begin{algorithm}[t]
\caption{Learning-Augmented Online Decentralized Optimization for Agent $v\in\mathcal{V}$}
\begin{algorithmic}[1]\label{alg:RP-OBD}
\REQUIRE Expert policy $\pi^{\dagger}_v$, and ML policy $\tilde{\pi}_v$
\FOR {$t=1,\cdots, T$}
\STATE Collect local online information $I^v_t$.
\STATE Obtain ML prediction $\tilde{x}_t^{v}$
and expert action $x_t^{v,\dagger}$ 
based on $I_t^{v}$, respectively. 
\STATE Choose the action $ x_t^v = \gamma \tilde{x}_t^v + (1 - \gamma) x_t^{v, \dagger}$ in \textbf{\ouralglinear},
and $x_t  =  \psi_{\lambda}(\tilde{x}_t^v)$ by ~\eqref{eqn:projection}
in \textbf{\ouralg}.\;
\ENDFOR
\end{algorithmic}
\end{algorithm}

\subsection{Algorithm Design}

In our problem, each individual agent $v \in \cV$ is provided with the potentially untrusted ML advice $\tilde{x}^v_t$
and the trusted expert advice $x_t^\dagger$ at time  time $t \in [1, T]$ based on its local online information $I_t^v$ specified in \eqref{eqn:information_set}.
The assumption of an offline-trained
predictor (i.e., ML policy in our case) 
is standard in learning-augmented algorithms \cite{L2O_LearningMLAugmented_Regression_CR_GeRong_NIPS_2021_anand2021a,OnlineOpt_ML_Advice_Improving_Google_Ravi_NeurIPS_2018_NEURIPS2018_73a427ba,OnlineOpt_ML_Adivce_Survey_2016_10.1145/2993749.2993766,Shaolei_SOCO_RobustLearning_OnlineOpt_MemoryCosts_Infocom_2023} as well as general learning-based optimizers \cite{L2O_NewDog_OldTrick_Google_ICLR_2019,L2O_Adversarial_Robust_GAN_arXiv_2020,Shaolei_LearningRobustCombinatorial_Zhihui_Infocom_2022}.
For our problem, approaches such as
multi-agent reinforcement learning  \cite{zhang2018drl_collab,multi-agent_RL_overview_zhang2021multi,cooperative_MARL_review_oroojlooy2022review} can be used to train ML policies for each agent.
When the context is clear, we also interchangeably use ML \emph{prediction} to refer to the ML action or advice.

Had we known which policy --- the ML policy $\tilde{\pi}$
or the expert policy $\pi^{\dagger}$ --- would be better for a problem
instance in advance, the problem would become trivial and we just need to choose the better policy. But, this is not possible in an online setting.
To exploit the potential of ML predictions
by augmenting the expert advice $x_t^\dagger$
with the ML advice $\tilde{x}^v_t$, out first attempt is 
to construct a linear combination of the two advice for each agent, 
which is defined as follows:
\begin{equation}
    x_t^v = \gamma \tilde{x}_t^v + (1 - \gamma) x_t^{v, \dagger} , \quad \forall v \in \cV
\end{equation}
where $\gamma \in [0,1]$ is the hyperparameter that reflects our confidence in ML predictions:
the larger
$\gamma \in [0,1]$, the more we trust the ML advice. 
We refer to this algorithm as \ouralglinear, which is also described in Algorithm~\ref{alg:RP-OBD}.
Note that, in Algorithm~\ref{alg:RP-OBD},
we run the expert policy 
(e.g., the localized policy proposed in \cite{Decentralized_SOCO_YihengLin_Adam_ICML_2022_pmlr-v162-lin22c}) independently as if it
is applied alone.
Thus, the expert policy $\pi^{\dagger}_v$
does not need to use all the information in 
$I^v_t$.

\subsection{Performance Analysis}

We first analyze the performance of \ouralglinear in terms of its average cost bound as follows.

\begin{theorem}[Average cost of \ouralglinear]\label{thm:linear_bound}
   For any $\gamma \in [0,1]$, the average cost of \ouralglinear is upper bounded by
    \begin{equation}
        \begin{aligned}
        \text{AVG}(\ouralglinear) \leq \min \Bigg\{& \gamma \text{AVG}(\tilde{\pi}) +  (1-\gamma)\text{AVG}(\pi^\dagger) ,\\
        &\left( \sqrt{\text{AVG}(\tilde{\pi})} +  (1-\gamma)\sqrt{\mathbb{E}_{g_{1:T}} \Bigl[ \sum_{t=1}^T \sum_{v\in \cV} \frac{\ell_f + 2\ell_T +  \ell_S D_v}{2} \| \tilde{x}_t^v - {x}^{v, \dagger}_t \|^2 \Bigr]} \right)^2 \Bigg\}
        \end{aligned}  
    \end{equation}
    where   $\text{AVG}(\tilde{\pi})$ and
    $\text{AVG}(\pi^\dagger)$ 
    are the average costs of the ML policy and expert over the distribution $g_{1:T}\sim\cP_{g_{1:T}}$, respectively.
\end{theorem}

In Theorem~\ref{thm:linear_bound}, the cost bound for \ouralglinear is given by the minimum of two terms: the first term is based on the convex property of the cost functions in our networked system, and the second term is derived in terms of the expected total distance between 
the ML and expert actions based on the smoothness of the costs. The proof is available in Appendix~\ref{appendix:proof_linear}.

Naturally, with a larger  $\gamma\in[0,1]$, 
the cost of \ouralglinear is more determined by the cost of the ML policy $\tilde{\pi}$. 
In practice, the ML policy is often trained to minimize the average cost,
while the expert policy is conservatively designed to address the worst case. Thus, for a well-trained ML policy, we typically have 
$\text{AVG}(\tilde{\pi})<\text{AVG}(\pi^\dagger)$.
This means that to minimize the average cost, we should 
choose $\gamma=1$, i.e., purely following the ML advice. 

While \ouralglinear can successfully exploit the potential of ML predictions by setting a large $\lambda\in[0,1]$, it hardly meets the $\lambda$-competitiveness constraint
(Definition~\ref{def:robustness}).
Indeed, 
unless the ML policy itself is sufficiently close
to the optimal policy for \emph{any} problem instance $g_{1:T}\in\mathcal{G}$,
\ouralglinear cannot meet the $\lambda$-competitiveness constraint. This is formalized as follows.

\begin{theorem}\label{coro:robustness_violation}
Given any problem instance $g_{1:T}\in\mathcal{G}$,
denote $\tilde{x}=[\tilde{x}_1,\cdots,\tilde{x}_T]$
and ${x}^{*}=[{x}_1^{*},\cdots,{x}_T^{*}]$
as the actions produced by the ML model and
the offline optimal policy, respectively, where we suppress
the dependency on $g_{1:T}$ for notational convenience.
Suppose that
the cost of the offline optimal policy is given by $cost(\pi^*,g_{1:T})$
For any linear combination hyperparameter $\gamma\in[0,1]$, if \ouralglinear
satisfies the $\lambda$-competitiveness constraint in Definition~\ref{def:robustness},
we must have
\begin{equation}\label{eqn:robustness_condition}
\frac{\|\tilde{x}-x^{*}\|^2}{cost(\pi^*,g_{1:T})} \leq \frac{2}{\beta}\left(\frac{1-\gamma}{\gamma}\sqrt{\rho_{\pi^{\dagger}}-1}+\frac{1}{\gamma}\sqrt{(1+\lambda)\rho_{\pi^\dagger}-1}\right)^2
\end{equation}
where $\beta>0$ is the strong convex parameter of
the node cost functions in Assumption~\ref{assumption:node_cost}, $\lambda>0$ is the competitiveness constraint parameter,
and $\rho_{{\pi}^{\dagger}}=\max_{g_{1:T}\in\mathcal{G}}\frac{cost({\pi}^{\dagger},g_{1:T})}{cost(\pi^*,,g_{1:T} )}>1$ is the competitive ratio
of the expert policy $\pi^{\dagger}$.
\end{theorem}

Theorem~\ref{coro:robustness_violation} is proved in Appendix~\ref{sec:proof_proposition3.2} and provides a necessary condition
for \ouralglinear to satisfy the $\lambda$-competitiveness constraint with respect
to the expert policy. The metric
$\frac{\|\tilde{x}-x^{*}\|^2}{cost(\pi^*,g_{1:T})}$
in \eqref{eqn:robustness_condition} measures
the distance between the ML policy and the offline optimal policy
(normalized by the optimal cost) and
is also commonly
used by prior studies \cite{SOCO_OnlineOpt_UnreliablePrediction_Adam_Sigmetrics_2023_10.1145/3579442,Shaolei_L2O_ExpertCalibrated_SOCO_SIGMETRICS_2022}
to 
characterize the ML prediction quality.
Intuitively, as $\lambda>0$ increases,
the competitiveness constraint becomes more relaxed, and so does the
requirement on the ML prediction quality. Additionally,
as \ouralglinear relies less
on the ML policy (i.e., $\gamma\in[0,1]$ becomes smaller)
and/or the expert policy itself
has a higher competitive ratio $\rho_{{\pi}^{\dagger}}$,
there is a less stringent requirement
on $\frac{\|\tilde{x}-x^{*}\|^2}{cost(\pi^*,g_{1:T})}$ for $\lambda$-competitiveness with respect to the expert.

Importantly, Theorem~\ref{coro:robustness_violation} highlights
that, unless we completely ignore the ML advice
(i.e., setting $\gamma=0$), the discrepancy between the ML policy
and the optimal policy measured in terms of $\frac{\|\tilde{x}-x^{*}\|^2}{cost(\pi^*,g_{1:T})}$
must be upper bounded by \eqref{eqn:robustness_condition} 
for $\lambda$-competitiveness given any problem instance $g_{1:T}\in\mathcal{G}$. The larger $\gamma$,
the greater dependency on ML predictions
to improve the average performance, but the more difficult
to meet the worst-case  $\lambda$-competitiveness constraint.

In practice, it is extremely challenging,
if not impossible, to ensure that the ML predictions satisfy \eqref{eqn:robustness_condition} for \emph{any} problem instance. It is well-known  that, although a trained ML model can perform well on average,
its performance in certain (possibly rare) cases can have an arbitrarily bad quality,
especially when the testing problem instance is very distinct from those 
training instances. This is also the key motivation for safeguarding ML predictions
to guarantee the worst-case competitiveness.

The performance of a learning-augmented algorithm is also analyzed
under two extreme cases when the ML policy is arbitrarily bad and when
it is perfect (i.e., robustness-consistency analysis \cite{L2O_LearningMLAugmented_Regression_CR_GeRong_NIPS_2021_anand2021a,SOCO_MetricUntrustedPrediction_Google_ICML_2020_pmlr-v119-antoniadis20a}).
Next, we show the robustness and consistency of \ouralglinear.

\begin{definition}[Robustness-consistency]\label{def:robust_consistency}
Suppose that the competitive ratios of the ML policy 
$\tilde{\pi}$
and a learning-augmented online policy ${\pi}$ are $\rho_{\tilde{\pi}}=\max_{g_{1:T}\in\mathcal{G}}\frac{cost(\tilde{\pi},g_{1:T})}{cost(\pi^*,,g_{1:T} )}$
and 
$\rho_{{\pi}}=\max_{g_{1:T}\in\mathcal{G}}\frac{cost({\pi},g_{1:T})}{cost(\pi^*,,g_{1:T} )}$, respectively, where 
$\pi^*$ is the optimal offline policy. Then,
$\rho_{{\pi}}$ is called
the robustness of the policy $\pi$ 
when $\rho_{\tilde{\pi}}\to\infty$,
and the consistency 
 when $\rho_{\tilde{\pi}}=1$.
\end{definition}

\begin{corollary}[Robustness-consistency of \ouralglinear]\label{coro:robust_consistency}
When $\rho_{\tilde{\pi}}\to\infty$, 
the robustness of \ouralglinear is $\rho_{\ouralglinear}=\infty$ for $\gamma\in(0,1]$ and 
$\rho_{\ouralglinear}=\rho_{{\pi}^{\dagger}}$ for $\gamma=0$;
when $\rho_{\tilde{\pi}}=1$,
the consistency of \ouralglinear is upper bounded by $\rho_{\ouralglinear}=\gamma + (1-\gamma)\rho_{\pi^{\dagger}}$  where  $\rho_{{\pi}^{\dagger}}=\max_{g_{1:T}\in\mathcal{G}}\frac{cost({\pi}^{\dagger},g_{1:T})}{cost(\pi^*,,g_{1:T} )}>1$ is the competitive ratio
of the expert policy $\pi^{\dagger}$.
\end{corollary}

Corollary~\ref{coro:robust_consistency}
shows that while \ouralglinear can improve the competitive ratio
over the (best) expert policy $\pi^{\dagger}$ for
$\gamma\in(0,1]$, it has an unbounded robustness when the ML policy
has an arbitrarily high cost.  
This shows the tension between following ML predictions for improving
the average cost performance and staying close to the expert policy for worst-case robustness. 
Thus, both Theorem~\ref{coro:robustness_violation} and 
Corollary~\ref{coro:robust_consistency} highlight
 the key limitation of \ouralglinear, i.e., lack of worst-case performance
 guarantees.

\section{\ouralg: Adaptively Combining ML Advice and Expert Advice}\label{sec:algorithm}

The previous section highlights that 
\ouralglinear 
 with a {fixed} linear combination of
 the ML prediction and expert advice 
\emph{cannot} offer guaranteed competitiveness or robustness in the worst case when ML predictions are of arbitrarily low quality. 
To address this limitation,
this section proposes an \emph{adaptive} approach based on a novel spatial cost decomposition and temporal reservation cost. 
Specifically, we present learning-augmented decentralized online optimization (\ouralg), an algorithm that adaptively exploits the benefits of ML
while guaranteeing $\lambda$-competitiveness against
any given expert policy $\pi^{\dagger}$ in a network.

\subsection{Algorithm Design}\label{sec:robustalgorithm}

We present our learning-augmented decentralized online algorithm, \ouralg, in Algorithm~\ref{alg:RP-OBD}, where
an ML policy is trained offline and deployed online by each agent
$v$ as in \ouralglinear.

To address the limitation of \ouralglinear
and
guarantee $\lambda$-competitiveness to the expert, the crux of \ouralg 
is to carefully leverage ML predictions while being
close enough to expert actions.
Specifically, we design a novel robust action set
that addresses the key challenge that only
local online information $I^v_t$ is available to each agent $v$ in our decentralized setting.
By choosing an action that falls
into the robust action set while staying close
to the ML prediction, \ouralg guarantees $\lambda$-competitiveness and exploits the benefits of ML predictions,
achieving the best of both worlds.

Concretely, we project the ML prediction
$\tilde{x}_t^{v}$ into the robust action
set denoted by $\mathcal{X}^v_{\lambda,t}$
as follows
\begin{equation}\label{eqn:projection}
x_t^v=\arg\min_{x\in \mathcal{X}^v_{\lambda,t}}\|x-\tilde{x}_t^{v}\|^2,
\end{equation}
where the robust action set $\mathcal{X}^v_{\lambda,t}$ 
is convex 
and will be specified
in Section~\ref{sec:robust_action_set_design}.
Thus, the projection in \eqref{eqn:projection}
can be efficiently performed by solving convex optimization at  each individual
agent $v$.

In contrast with the
{fixed} linear combination of
 the ML prediction and expert advice
in \ouralglinear
that 
is provably insufficient for competitiveness guarantees,
the novel robust action set we design for \ouralg is
adaptively chosen based on the online costs of actual actions
and the expert policy, guaranteeing $\lambda$-competitiveness for
any $\lambda>0$.

\subsection{Designing a Robust Action Set}\label{sec:robust_action_set_design}

The core of \ouralg is an action set that ``robustifies'' ML
predictions for $\lambda$-competitiveness. 
This is challenging due to the temporal and spatial
information inefficiency --- the $\lambda$-competitiveness requirement in \eqref{eqn:objective_function}
is imposed over the total global cost over $T$ steps, whereas each agent must choose its action based on
\emph{local} and \emph{online} information $I_t^v$. 

To construct a robust action set $\mathcal{X}^v_{\lambda,t}$ 
locally computable by each agent, 
we first convert the $\lambda$-competitiveness constraint 
over $T$ time steps  to an equivalent \emph{anytime} constraint below.

\begin{proposition}[Anytime $\lambda$-competitiveness]\label{theorem:anytime_constraint_equivalent}
For any $\lambda>0$, to guarantee the $\lambda$-competitiveness constraint
$cost(\pi,g_{1:t} )\leq (1+\lambda) cost(\pi^{\dagger},g_{1:t})$,   
$\forall g_{1:t}\in\mathcal{G}$,
a sufficient and necessary condition is
\begin{equation}\label{eqn:anytime_constraint}
cost(\pi,g_{1:t} )\leq (1+\lambda) cost(\pi^{\dagger},g_{1:t}),   
\;\;\;\forall t\in[1,T], 
\end{equation}
 where $cost(\pi,g_{1:t})$
is the cumulative global cost of a policy $\pi$ up to time $t\in[1,T]$.
\end{proposition}
\begin{proof}
    The sufficient part in Proposition~\ref{theorem:anytime_constraint_equivalent}
is straightforward, while the necessary part can be proved by constructing
a counter-example as follows. Suppose that there is a time $t\in[1,T-1]$
such that $cost(\pi,g_{1:t} )\geq(1+\lambda) cost(\pi^{\dagger},g_{1:t})+\epsilon$, where $\epsilon>0$.
It is possible that
 the expert's future total cost $cost(\pi^{\dagger},g_{t+1:T})<\frac{\epsilon}{1+\lambda}$.
Then, by the non-negativeness of the cost functions, the policy
$\pi$'s total cost
$cost(\pi,g_{1:T} )\geq cost(\pi,g_{1:t})
\geq(1+\lambda) cost(\pi^{\dagger},g_{1:t})+\epsilon
>(1+\lambda)\cdot\left[cost(\pi^{\dagger},g_{1:t})+cost(\pi^{\dagger},g_{t+1:T})\right]=(1+\lambda)cost(\pi^{\dagger},g_{1:T})$, violating
 $\lambda$-competitiveness.
\end{proof}

While Proposition~\ref{theorem:anytime_constraint_equivalent}
simplifies the $\lambda$-competitiveness constraint, the spatial
cost in \eqref{eqn:anytime_constraint} cannot be locally computed
by each agent in a decentralized manner without knowing its neighboring agents' actions. Moreover, due to the future uncertainties and coupling
of actions in online optimization, it is very challenging to meet
the constraints \eqref{eqn:anytime_constraint} for every $t\in[1,T]$.
To address these challenges,
we propose 
novel \emph{adaptive spatial cost decomposition} and
introduce \emph{reservation costs} to safeguard online actions for $\lambda$-competitiveness. 

\subsubsection{Spatial Cost Decomposition}\label{sec:global_cost_decomposition}
Due to the decentralized setting, we first decompose the global cost $g_t(x_t)$ at time $t$ into locally computable costs for individual agents $v\in\mathcal{V}$ expressed as
\begin{equation}\label{eqn:local_cost_agent_v}
\begin{split}
g_t^v(x^v_{t})
=&f_t^v(x_t^v)+c_t^v(x_t^v,x_{t-1}^v) +\sum_{(u,v)\in\mathcal{E}}\kappa_t^{(v,u)}s^{(u,v)}_t(x_t^v,x_t^u),
\end{split}
\end{equation}
where we use the weights $\kappa_t^{(v,u)}\geq0$ and $\kappa_t^{(v,u)}\geq0$, such that
$\kappa_t^{(v,u)}+\kappa_t^{(u,v)}=1$ for $(v,u)\in\mathcal{E}$,
to adaptively split the shared spatial cost $s^{(u,v)}_t(x_t^v,x_t^u)$ between the two connected
agents (i.e.,
$\kappa^{(u,v)}$ for agent $v$
and $\kappa^{(v,u)}$ for agent $u$). 
We specify the choice of the weight 
$\kappa_t^{(v,u)}$ in \eqref{eqn:splitting_weight} later.

Based on the cost decomposition in \eqref{eqn:local_cost_agent_v}, the anytime $\lambda$-competitiveness constraint in ~\eqref{eqn:anytime_constraint} can be guaranteed if the action of each node $v$ satisfies the following local constraint: 
\begin{equation}\label{eqn:easyconstraint}
\sum_{i=1}^t g_i^v(x^v_{i})\leq (1+\lambda) \sum_{i=1}^t g_i^v(x^{v,\dagger}_{i}),\;\;\;\;\forall t\in[1,T].
\end{equation}

At step $t$,
however, agent $v$ cannot evaluate its local cost $g_t^v(x^v_{t})$, because it has no access to the actions $x_t^u$ and expert actions $x_t^{u,\dagger}$ of its connected neighbors $u$ and hence cannot calculate the actual or expert's spatial costs for $(v,u)\in\mathcal{E}$. 

Additionally, even if agent $v$ has the knowledge of $g_t^v(x^v_{t})$, simply satisfying \eqref{eqn:easyconstraint} at time  $t$ 
cannot guarantee that a feasible action exists to satisfy the local constraints for future steps 
$t+1,\cdots,T$ due to the temporal cost. 
To see this, consider a toy example with $T=2$ and $c_t^v=\|x_t^v-x_{t-1}^v\|^2$. Assume that $x_1^v$ is selected such that the first-step local constraint is satisfied by equality, i.e., $g_1^v(x^v_{1})= (1+\lambda) g_i^v(x^{v,\dagger}_{1})$. Then, at the second step $t=2$, it can happen that the node
costs satisfy $f_2^v(x^{v,\dagger}_{1})=0$ and $f_2^v(x^v_{1})>0$, while the spatial costs are all zero. Then, with the expert action $x^{v,\dagger}_{2}=x^{v,\dagger}_{1}$, it follows that $g_2^v(x^v_{2})>(1+\lambda)g_2^v(x^{v,\dagger}_{2})=0$ for any $x^v_{2}\in\mathcal{X}$, thus violating the local constraint \eqref{eqn:easyconstraint} for agent $v$. By the same reasoning, the $\lambda$-competitiveness constraint can be violated for the whole network.

\begin{figure*}[!t]
\begin{equation}\label{eqn:local_constraint}
\begin{aligned}
&\sum_{\tau=1}^t f_\tau^{v}(x_\tau^{v}) + \sum_{\tau=1}^t  c_\tau ^{v}(x_\tau^{v}, x_{\tau-1}^{v})+ \sum_{\tau=1}^{t-1}\sum_{(v, u) \in \mathcal{E}} \kappa_{\tau}^{(v,u)} \cdot {s_{\tau}^{(v, u)}(x_{\tau}^{v}, x_{\tau}^u)} + R(x_t^v,x_t^{v,\dagger}) \\
    \leq&  (1+\lambda) \Bigl(\sum_{\tau=1}^t f_\tau^{v}(x_\tau^{v, \dagger})  + \sum_{\tau=1}^t c_\tau ^{v}(x_\tau^{v, \dagger}, x_{\tau-1}^{v, \dagger}) +  
    \sum_{\tau=1}^{t-1}\sum_{(v, u) \in \mathcal{E}} \kappa_\tau^{(v,u)} \cdot {s_\tau^{(v, u)}(x_\tau^{v, \dagger}, x_\tau^{u, \dagger})}\Bigr) 
\end{aligned}
\end{equation}
\hrule
\end{figure*}

\subsubsection{Robust Action Sets via Reservation Costs}\label{section:robust_action_set_reservation}

To ensure non-empty sets of feasible actions satisfying
the local constraints \eqref{eqn:easyconstraint} for each time step $t$, we propose a \emph{reservation cost}
that safeguards each agent $v$'s action against any possible uncertainties (e.g., connected agent $u$'s current actions and future cost functions).
Compared to  a centralized setting \cite{Shaolei_SOCO_RobustLearning_OnlineOpt_MemoryCosts_Infocom_2023,SOCO_OnlineOpt_UnreliablePrediction_Adam_Sigmetrics_2023_10.1145/3579442},
designing a proper reservation cost 
in a decentralized creates substantial challenges, as it
needs to hedge against both \emph{spatial} and future \emph{temporal}
uncertainties.

With only local online information $I_t^v$ available
to agent $v$,
 the key insight of our added reservation cost at each time
step $t$ is to bound
the maximum possible cost difference between agent
$v$'s cost $\sum_{i=1}^t g_i^v(x^v_{i})$ and its corresponding cost constraint
$(1+\lambda) \sum_{i=1}^t g_i^v(x^{v,\dagger}_{i})$
for future time steps.
More concretely, 
 we use a new constraint
in \eqref{eqn:local_constraint} to define the robust action set for agent $v$ at step $t$.
In constraint 
\eqref{eqn:local_constraint}, 
the weight $\kappa_{\tau}^{(v,u)}$ (attributed
to agent $v$) for adaptively splitting the spatial
cost $s_{\tau}^{(v, u)}$ between agent $v$ and
agent $u$ for $\tau=1,\cdots,t-1$ is 
\begin{equation}\label{eqn:splitting_weight}
    \kappa_{\tau}^{(v,u)} = \frac{\|x_{\tau}^v - x_{\tau}^{v,\dagger} \|^2}{\|x_{\tau}^v - x_{\tau}^{v,\dagger} \|^2 + \| x_{\tau}^u - x_{\tau}^{u,\dagger} \|^2}.
\end{equation}
Additionally, the reservation cost\footnote{Despite the one-step delayed feedback of the neighbors' actions, knowing the spatial cost function $s_t^{v,u}(\cdot,\cdot)$ at the beginning of time $t$ is still helpful. For example, the reservation cost for spatial cost
uncertainties can be reduced if the smoothness constant of $s_t^{v,u}(\cdot,\cdot)$ is smaller than $\ell_s$.} is
\begin{equation}\label{eqn:reservation_function}
 R(x_t^v,x_t^{v,\dagger})=\frac{\ell_T + \ell_S\cdot D_{v}}{2}(1+\frac{1}{\lambda_0})  \|x_t^{v} - x_t^{v, \dagger} \|^2,
\end{equation}
where $\ell_T$ and $\ell_S$ are smoothness parameters
for the temporal and spatial cost functions,
$D_v$ is the degree of agent $v$
(i.e., the number of agents
connected to agent $v$), and $0< \lambda_0\leq \lambda$ is a hyperparameter to adjust the size of the robust action set (and will be optimally chosen as $\lambda_0 = \sqrt{1+\lambda} - 1$ in Theorems~\ref{thm:avg_cost_blackbox} and~\ref{thm:avg_cost_e2e}). In the special case
when both $x_{t}^v = x_{t}^{v,\dagger}$
and $x_{t}^u = x_{t}^{u,\dagger}$, we set $\kappa_{\tau}^{(v,u)}=\frac{1}{2}$ 
in \eqref{eqn:splitting_weight}.

Importantly, the new constraint \eqref{eqn:local_constraint} for agent $v$ 
can be calculated purely based on local online
information $I_t^v$; it
only depends on the cumulative node and temporal costs up to time $t$,
as well as the spatial costs (including the feedback of the connected neighboring agents' actions and their expert actions) up to time $t-1$.
Thus, the overall cost to share information between two connected agents in the network
is small.
Moreover, the reservation cost $R(x_t^v,x_t^{v,\dagger})$ 
safeguards agent $v$'s action not only against uncertainties in future temporal cost functions in  online optimization, but also against delayed spatial costs resulting from decentralized optimization, which we further explain as follows.

$\bullet$ \textbf{Temporal uncertainties}.
The temporal cost couples each agent's actions over time,
but the online action needs to be chosen without knowing
all the future costs. Consequently, 
as shown in the  example in Section~\ref{sec:global_cost_decomposition},
simply satisfying
the $\lambda$-competitiveness in terms of the cumulative cost
up to $t$ does not necessarily ensure $\lambda$-competitiveness in the future.
To hedge against temporal uncertainties,  our
reservation cost $R(x_t^v,x_t^{v,\dagger})$ in 
\eqref{eqn:reservation_function} includes
the term
$\frac{\ell_T}{2}(1+\frac{1}{\lambda_0})  \|x_t^{v} - x_t^{v, \dagger} \|^2$, which bounds the maximum cost disadvantage
for agent $v$: 
 $c_t ^{v}(x_t^{v}, x_{t+1}^{v})-(1+\lambda)c_t ^{v}(x_t^{v, \dagger}, x_{t+1}^{v, \dagger})
 \leq \frac{\ell_T}{2}(1+\frac{1}{\lambda_0})\|x_t^{v}-x_t^{v, \dagger}\|^2$.
Thus, $x_{t+1}^{v}=x_{t+1}^{v,\dagger}$ is always a feasible
robust action for agent $v$ at time $t+1$.

$\bullet$ \textbf{Spatial uncertainties}.
In our decentralized setting, agent $v$ chooses its action based on the local online information $I_t^v$,
which creates spatial uncertainties regarding
its connected neighboring agents' actions and spatial costs.
In our design, with
the splitting weight $\kappa_{t-1}^{(v,u)}$ in \eqref{eqn:splitting_weight} and
the term $\frac{\ell_S\cdot D_{v}}{2}(1+\frac{1}{\lambda_0})  \|x_{t-1}^{v} - x_{t-1}^{v, \dagger} \|^2$ included in the reservation cost in \eqref{eqn:reservation_function} at time $t-1$,
we ensure that our constraint in \eqref{eqn:local_constraint}, if satisfied, can
always guarantee the local
constraint in \eqref{eqn:easyconstraint} and hence
also the $\lambda$-competitiveness constraint, due to the following inequality:
\begin{equation}
\begin{split}
&\sum_{(v, u) \in \mathcal{E}} \kappa_{t-1}^{(v, u)} \left({s_{t-1}^{(v, u)}(x_{t-1}^{v}, x_{t-1}^u)}-(1+\lambda)  {s_{t-1}^{(v, u)}(x_{t-1}^{v, \dagger}, x_{t-1}^{u, \dagger})}\right)\\
\leq& \sum_{(v, u) \in \mathcal{E}} \kappa_{t-1}^{(v,u)} \frac{\ell_S}{2}(1+\frac{1}{\lambda_0})\left(\|x_{t-1}^{v}-x_{t-1}^{v, \dagger}\|^2+\|x_{t-1}^u-x_{t-1}^{u, \dagger}\|^2 \right) = \sum_{v\in \cV}\frac{\ell_S\cdot D_{v}}{2}(1+\frac{1}{\lambda_0})  \|x_{t-1}^{v} - x_{t-1}^{v, \dagger} \|^2.
\end{split}
\end{equation}
Note that, as the degree  $D_v$ of node $v$ increases,
more agents are connected to agent $v$ and hence spatial uncertainties also naturally increase, resulting in an increased
reservation cost in \eqref{eqn:reservation_function}.

In summary, our novel robust action set for agent $v$ at time step $t$ is designed as \begin{equation}\label{eqn:robust_action_set}
\mathcal{X}_{\lambda,t}^v=\{x_t^v\mid x_t^v \text{ satisfies } \eqref{eqn:local_constraint} \text{ for step } t\},
\end{equation}
which, by convexity of cost functions, is convex and 
leads to computationally-efficient projection \eqref{eqn:projection}.
For example, in our experiments, it takes about 1 second
to run inference for 1000+ instances on a laptop.

\section{Performance Bounds for \ouralg}\label{sec:analysis}
We now analyze 
\ouralg in terms
of its competitiveness, average cost, and robustness-consistency, proving that
\ouralg is $\lambda$-competitive against any given expert and simultaneously achieves finite consistency.

\subsection{$\lambda$-Competitiveness}\label{sec:robustness_analysis}

We state $\lambda$-competitiveness  of \ouralg as follows. The proof is provided in Appendix~\ref{sec:robustness_proof}.
\begin{theorem}\label{thm:robustness}
    ($\lambda$-competitiveness of \ouralg) Given any ML policy $\tilde{\pi}$ and expert policy
    $\pi^{\dagger}$, for any $\lambda > 0$ and $\lambda_0 \in (0, \lambda]$ in
    the robust action set in \eqref{eqn:robust_action_set}, the cost of \ouralg  satisfies $cost(\ouralg, g_{1:T}) \leq (1+\lambda)\cdot cost(\pi^{\dagger}, g_{1:T})$ for any problem instance  $g_{1:T}\in\mathcal{G}$.
\end{theorem}

Theorem~\ref{thm:robustness} guarantees that, for any problem instance $g_{1:T}\in\mathcal{G}$, the total global cost of \ouralg is always upper bounded by $(1+\lambda)$ times the global cost of the expert policy $\pi^{\dagger}$,
regardless of the quality of ML predictions. 
This competitiveness guarantee is the first 
in the context of decentralized
learning-augmented algorithms
 and attributed to our novel design of locally computable
 robust action sets in \eqref{eqn:robust_action_set}, based
on which each agent individually safeguards
its own online actions. 
 Moreover, for our setting, there exist online policies
(e.g., localized policy for multi-agent
networks \cite{Decentralized_SOCO_YihengLin_Adam_ICML_2022_pmlr-v162-lin22c})
that have bounded competitive ratios against
the offline oracle and hence can be readily applied
as expert policies in \ouralg. Thus,
their competitive ratios immediately translate
with a scaling factor of $(1+\lambda)$ into 
competitiveness of \ouralg 
against the offline oracle. 

\subsection{Average Cost}\label{sec:avg_cost_analysis}

A key goal of utilizing an ML policy is to improve
the \emph{average} performance over the expert policy. Thus, we first consider the average performance of \ouralg 
under a general ML policy.
We rewrite \ouralg as $\ouralg(\tilde{\pi})$ to highlight its dependency
on $\tilde{\pi}$ when applicable. The results
are shown in Theorem~\ref{thm:avg_cost_blackbox},
whose proof relies on the spatial cost decomposition developed in Section~\ref{sec:global_cost_decomposition} 
and is deferred to Appendix~\ref{sec:cost_ratio_proof}.

\begin{theorem}\label{thm:avg_cost_blackbox}
(Average Cost of $\ouralg(\tilde{\pi})$) 
Given an expert policy $\pi^\dagger$ and any ML policy $\tilde{\pi}$, for the context distribution $\mathcal{P}_{g_{1:T}}$, we define $AVG(\pi^{\dagger})$, $AVG(\tilde{\pi})$ as the average costs of 
the expert policy and ML policy. For any $\lambda > 0$, by optimally setting $\lambda_0 = \sqrt{1+\lambda} - 1$, the average cost of $\ouralg(\tilde{\pi})$ is upper bounded by
\begin{equation}
\nonumber
\begin{aligned}
        AVG(\ouralg(\tilde{\pi}))\leq \min \left\{ (1+\lambda) AVG(\pi^{\dagger}), \biggl( \sqrt{\text{AVG}(\tilde{\pi})}  + \sqrt{\sum_{v\in \mathcal{V}} \omega_v(\lambda, \tilde{\pi}, \pi^{\dagger}) } \biggr)^2 \right\},
    \end{aligned}
\end{equation}
where $\omega_v(\lambda, \tilde{\pi}, \pi^{\dagger}) =  \mathbb{E}_{g_{1:T}}\left\{\sum_{t=1}^T \left[ \frac{\ell_f + 2\cdot\ell_T + \ell_S\cdot D_{v}}{2} \|\tilde{x}^v_t - x_t^{v,\dagger}\|^2 -(\sqrt{1+\lambda} - 1)^2 \cdot  \text{cost}_{v,t}^\dagger  \right]^+\right\} $,
in which $D_v$ is the degree of node  $v$ and $\text{cost}_{v,t}^\dagger$ denotes 
the sum of hitting cost and switching cost for the expert $\pi^\dagger$.
\end{theorem}

Theorem~\ref{thm:avg_cost_blackbox} quantifies the tradeoff
between exploiting the ML policy for average cost performance
and following the expert policy for worst-case competitiveness in a decentralized setting. 
Specifically, 
the average cost bound 
of $\ouralg(\tilde{\pi})$ is a minimum of two terms. The first term holds due to the guaranteed $\lambda$-competitiveness against the expert policy. The second term shows
that, due to the competitiveness requirement, 
$\ouralg(\tilde{\pi})$ can deviate from the ML policy
and hence have
a higher average cost 
than $AVG(\tilde{\pi})$. 
The cost difference 
is primarily driven by the sum of $\omega_v(\lambda, \tilde{\pi},\pi^{\dagger})$ for all nodes in the set $\mathcal{V}$, which measures how well \ouralg follows the ML policy. For each node $v$, $\omega_v(\lambda, \tilde{\pi},\pi^{\dagger})$ is upper bounded by the expected distance between actions made by \ouralg and actions made by the pure ML policy. Naturally, as we impose a less stringent competitiveness constraint (i.e., smaller $\lambda>0$) or the expert policy $\pi^\dagger$ and the ML policy $\tilde{\pi}$ are better aligned (i.e., smaller action distance $\|\tilde{x}_t^v - {x}_t^{v,\dagger}\|$), 
we can better exploit the power of the ML policy $\tilde{\pi}$ with a reduced  $\omega_v(\lambda, \tilde{\pi},\pi^{\dagger})$. 
Another insight 
is that
 $\omega_v(\lambda, \tilde{\pi},\pi^{\dagger})$ decreases when 
the expert policy has a higher cost, which
naturally provides more freedom to \ouralg to follow
ML while still being able to satisfy the $\lambda$-competitiveness requirement.

\textbf{Impact of network topologies.} 
Given the same set of nodes but different numbers of node connections,
the spatial costs can be significantly different. This impact
is also captured by the term 
\begin{equation}
    \omega_v(\lambda, \tilde{\pi}, \pi^{\dagger}) =  \mathbb{E}_{g_{1:T}}\left\{\sum_{t=1}^T \left[ \frac{\ell_f + 2\cdot\ell_T + \ell_S\cdot D_{v}}{2} \|\tilde{x}^v_t - x_t^{v,\dagger}\|^2 -(\sqrt{1+\lambda} - 1)^2 \cdot  \text{cost}_{v,t}^\dagger  \right]^+\right\}
\end{equation}
where $D_v$ is the degree of agent $v$.
\revise{Specifically, when agent $v$ is connected with more nodes (i.e., greater $D_v$) while the other factors are held constant, the spatial costs and uncertainties also increase accordingly. The competitiveness guarantees compel agent $v$ to more conservatively follow the expert policy and potentially deviate more from the ML policy.
In other words, the cost gap bound between \ouralg and the ML policy $\omega_v(\lambda, \tilde{\pi}, \pi^{\dagger})$ increases with the increased node degree. }
Thus, in general, when the graph density increases with more node connections, 
the total cost bound compared
to the ML policy also increases because of more spatial cost uncertainties and hence potentially more perturbations added
to the ML advice.

Interestingly, even given the same number of node connections (i.e., edges) and the same number of nodes,
how the nodes are connected (e.g., linear chain vs. star graphs
in Fig.~\ref{fig:graph_topology}) can play a role in the cost.
For example, when every node has a small degree in a linear chain graph and
the competitiveness constraint $\lambda\geq0$ is not too small,
 $\omega_v(\lambda, \tilde{\pi}, \pi^{\dagger})$ 
 is generally smaller due to the ReLu operation, making it easier to follow the ML policy in \ouralg;
on the other hand, when a node has a very high degree (in a star graph),
the term $\omega_v(\lambda, \tilde{\pi}, \pi^{\dagger})$
for the high-degree node is likely to be positive (unless $\lambda$ is sufficiently large), i.e., this node's action likely deviates significantly from its ML policy.
\revise{
Consequently, when the other factors are held constant, \ouralg can more effectively adhere to the ML policy and achieve better average performance in a linear chain graph compared to a star graph, despite the identical number of spatial connections in both graghs.
This phenomenon is empirically observed and discussed in our experiments, as illustrated in Figure~\ref{fig:graph_topology_main}.
}

To further highlight the impact of node connections, we  extend the design and cost analysis of \ouralg from an undirected graph
to a directed graph.
The results are available in Appendix~\ref{appendix:extension_directed_graph}.

\subsection{Robustness and Consistency}\label{sec:robust_consistency_lado}

We now show the robustness and consistency (Definition~\ref{def:robust_consistency}) for \ouralg as follows.

\begin{theorem}[Robustness-consistency of \ouralg]\label{theorem:robust_consistency_lado}
Define $\rho_{\tilde{\pi}}$,
$\rho_{{\pi}^{\dagger}}$ and
$\rho_{\ouralg}$ as the competitive ratios of
the ML policy $\tilde{\pi}$,
expert policy $\pi^{\dagger}$ and 
\ouralg against
the offline optimal policy $\pi^*$, 
and $\ell=\frac{\ell_f+2\ell_T+\ell_SD_{\max}}{2}$
as the gradient Lipschitz constant of the global
cost function,
respectively.  
When $\rho_{\tilde{\pi}}\to\infty$, 
the robustness of \ouralg is upper bounded by $\rho_{\ouralg}\leq(1+\lambda)\rho_{{\pi}^{\dagger}}$ for any $\lambda>0$;
when $\rho_{\tilde{\pi}}=1$,
the consistency of \ouralg is upper bounded by 
\begin{equation}\label{eqn:consistency_lado}
\begin{split}
\rho_{\ouralg}
&\leq \min \left\{\left[1+ \sqrt{\max_{g_{1:T}\in\mathcal{G}}   \frac{\sum_{v\in\mathcal{V}}\sum_{t=1}^T \left[ \ell\|x^{v,*}_t - x_t^{v,\dagger}\|^2 -  (\sqrt{1+\lambda} - 1)^2  \cdot {cost}_{v,t}^\dagger  \right]^+}{cost(x^*_{1:T},g_{1:T})}}\right]^2, (1+\lambda)\rho_{\pi^{\dagger}}\right\}\\
&\leq  \min\left\{\left(1+ 2\sqrt{\frac{\ell}{\beta}\cdot (\rho_{\pi^{\dagger}}-1)}\right)^2,\;(1+\lambda)\rho_{\pi^{\dagger}}\right\}.
\end{split}
\end{equation}
where $\beta>0$ is the strong convex parameter (Assumption~\ref{assumption:node_cost})
and ${cost}_{v,t}^\dagger $ is the expert's cost of node $v\in\mathcal{V}$ 
at time $t\in[1,T]$. 
\end{theorem}

Theorem~\ref{theorem:robust_consistency_lado} is proved
in Appendix~\ref{sec:proof_robust_consistency_lado} and
highlights that \ouralg can achieve both finite robustness
and consistency simultaneously. In contrast with
\ouralglinear which fails to provide finite robustness,
\ouralg prioritizes the worst-case competitiveness guarantees 
as a constraint parameterized by $\lambda>0$.
Nonetheless, in general cases, \ouralg does not have a better consistency
than \ouralglinear or the best expert policy when the ML predictions are perfect. While it remains an open problem to achieve
the optimal robustness-consistency tradeoff (except for a few specific problems) in the learning-augmented literature
\cite{SOCO_MetricUntrustedPrediction_Google_ICML_2020_pmlr-v119-antoniadis20a},
we note that our result is consistent with the fundamental impossibility
in our problem setting. 
Specifically,  even 
in the special single-agent case for our problem setting, 
the prior studies \cite{SOCO_OnlineOpt_UnreliablePrediction_Adam_Sigmetrics_2023_10.1145/3579442} have shown
that it is \emph{impossible} to achieve finite robustness while still having
 a consistency better than the best expert's competitive ratio $\rho_{{\pi}^{\dagger}}$ without further assumptions.
 As a result,
\ouralglinear achieves a better consistency than 
best expert's competitive ratio
(Theorem~\ref{coro:robust_consistency}) 
and hence cannot guarantee finite robustness;
\ouralg guarantees finite robustness (Theorem~\ref{thm:robustness}) 
and hence cannot offer a better consistency in general cases. Nonetheless,
by making an additional assumption that
  the expert's ${cost}_{v,t}^\dagger $
of each node $v\in\mathcal{V}$ is always strictly positive 
at time $t\in[1,T]$, we see from the first inequality 
in \eqref{eqn:consistency_lado}
that \ouralg can achieve a lower consistency by increasing $\lambda>0$
and even simultaneously $1$-consistency and finite robustness when  $\lambda>0$ is sufficiently large (which pushes
the term $\left[ \ell\|x^{v,*}_t - x_t^{v,\dagger}\|^2 -  (\sqrt{1+\lambda} - 1)^2  \cdot {cost}_{v,t}^\dagger  \right]^+\to0$ in \eqref{eqn:consistency_lado}). 

Importantly, the robustness and consistency analysis is still
for the \emph{worst} case. By utilizing a well-trained ML model
in \ouralglinear and \ouralg,
we can still improve the average cost performance compared
to the pure expert policy, highlighting the key advantage of ML predictions. 
This is discussed 
in Theorem~\ref{thm:linear_bound} and Theorem~\ref{thm:avg_cost_blackbox},
and also empirically demonstrated in Section~\ref{sec:numerical_results}.

Theorem~\ref{thm:avg_cost_blackbox} applies to any ML models, including
 ML models that are trained as standalone optimizers without considering
 the design of \ouralg and hence may have training-testing objective mismatches.
To further improve the average cost performance,
we consider a projection-aware ML policy $\tilde{\pi}^{\circ}_{\lambda}$
that is optimally trained to minimize the actual cost with explicit consideration of the 
downstream projection in \ouralg.
Our performance analysis  formally demonstrates the 
benefits of using $\tilde{\pi}^{\circ}_{\lambda}$ in \ouralg
for average cost reduction and is available in Appendix~\ref{sec:end_to_end_ml}.

\section{Case Study: Decentralized Battery Management for Sustainable Computing}\label{sec:experiment}

To demonstrate the empirical benefits of
\ouralglinear and \ouralg, we carry out experiments with the application of decentralized battery management for sustainable computing, as introduced in Section~\ref{sec:application_example}. Our results show that
with learning augmentation, both \ouralglinear and \ouralg
can empirically achieve a good average cost performance.
Meanwhile, 
 compared to \ouralglinear that lacks guaranteed competitiveness, 
\ouralg is less sensitive to the potentially low quality of ML predictions.

\subsection{Experimental Setup}\label{section:case_study}

The recent surge in computing demands, such as large  AI models
for language services, has placed an urgent emphasis on
decarbonizing data centers for sustainability.
To highlight the potential of \ouralg for managing decentralized batteries in the context of sustainable data centers,
we use a trace-based simulation in our experiments
following the common practice in the literature  \cite{Carbon_GenerativeAI_AndrewChien_Chicago_HotCarbon_2023_10.1145/3604930.3605705,SOCO_DynamicRightSizing_Adam_Infocom_2011_LinWiermanAndrewThereska}.
The data center workload trace is taken from Microsoft Azure \cite{cortez2017resource}, which contains the CPU utilization of 2,695,548 virtual machines (VM) for each 5-minute window. We estimate the energy consumption $P_{d,t}$ by summing up the CPU utilization of all VMs. 

The weather-related parameters, i.e., wind speed, solar radiation and temperature data, are all collected from the National Solar Radiation Database \cite{sengupta2018national}. Based on the weather information, we use empirical equations to model the wind and solar renewables generated at time step $t$.
Specifically, the amount of solar energy generated at step $t$ is given based on \cite{wan2015photovoltaic} by
 $P_{\mathrm{solar},t}=\frac{1}{2}\kappa_{\mathrm{solar}} A_{\mathrm{array}} I_{\mathrm{rad},t}(1-0.05*(\mathrm{Temp}_t-25))$,
where $A_{\mathrm{array}}$ is the solar array area ($m^2$),  $I_{\mathrm{rad},t}$ is the solar radiation ($kW/m^2$), and $\mathrm{Temp}_t$ is the temperature ($^{\circ}$C) at time $t$, and $\kappa_{\mathrm{solar}}$ is the conversion efficiency (\%) of the solar panel.
The amount of wind energy is modeled 
based on \cite{sarkar2012wind} as
 $  P_{\mathrm{wind},t}=\frac{1}{2}\kappa_{\mathrm{wind}}\varrho A_{\mathrm{swept}} V_{\mathrm{wind},t}^3$, where  $\varrho$ is the air density ($kg/m^3$), $A_{\mathrm{swept}}$ is the swept area of the turbine ($m^2$), $\kappa_{\mathrm{wind}}$ is the conversion efficiency (\%) of wind energy, and $V_{\mathrm{wind},t}$ is the wind speed ($kW/m^2$) at time  $t$.
Thus, at time $t$, the total energy generated by the solar and wind renewables
is $P_{\mathrm{r},t}=P_{\mathrm{wind},t}+P_{\mathrm{solar},t}$. By subtracting the renewables $P_{{r},t}$ from the data center's energy demand $P_{d,t}$, we obtain the net demand as $P_{n, t} = P_{d,t} - P_{{r},t}$, which 
is then normalized to $[-1, 1]$.

\revise{In our case study, we evaluate the performance of \ouralg and \ouralglin on a diverse set of experimental settings, including heterogeneous graph nodes with various graph topologies. To represent the range of battery health, we assign different self-degradation coefficients $A_v$ to these battery units. Beyond self-degradation coefficients, we further consider heterogeneity in the storage capacities and rated output powers from real-world energy storage systems. We begin with a fully connected graph of 3 battery units, then expand our experiments to 15-node graphs, exploring representative topologies (e.g., star, linear) and a variety of randomly generated graphs with varying densities. To assess the scalability of our algorithm, we generate fully-connected graphs with up to 120 nodes.}
More details on the extended experiments can be found in Appendix~\ref{appendix:experiment_battery}.

\subsection{Baselines}

We compare \ouralglinear and \ouralg with the following baselines. 
In addition, we also evaluate
\ouralgft that uses the optimal projection-aware ML policy
(Appendix~\ref{sec:end_to_end_ml}). These representative baselines are
closely related 
to our problem and range from the simplest Greedy to the most
powerful oracle OPT.
\\
$\bullet$ \textbf{Offline optimal} (OPT): OPT obtains the offline optimal solution to \eqref{eqn:decen_objective} with the complete information for each problem instance.\\
$\bullet$ \textbf{Expert}: The state-of-the-art online algorithm for our problem
is the localized prediction policy \cite{Decentralized_SOCO_YihengLin_Adam_ICML_2022_pmlr-v162-lin22c}.
Here, we set the prediction window as 1 and refer to it as Expert.\\
$\bullet$ \textbf{ML optimizer} (ML): ML uses the same recursive neural network (RNN) model used by \ouralg, but is trained as a standalone policy without considering \ouralg.\\
$\bullet$ \textbf{Hitting cost optimizer} (HitOnly): HitOnly solely optimizes the node cost for each node, which aims at tracking the nominal SoC value exactly. \\
$\bullet$ \textbf{Single-step cost optimizer} (Greedy): Greedy myopically minimizes the node cost and temporal cost at each time for each node.

\subsection{Results for Networks with 3 Nodes}\label{sec:numerical_results}
Considering a simple 3-node network, we show the empirical average (AVG) and competitive ratio (CR) values in Table~\ref{table:main}, where the best empirical AVG and CR values are marked in bold font. The CR values are the empirically worst cost ratio of
an algorithm's cost to OPT's cost in our testing dataset.
We see that \ouralglinear with $\gamma=0.1$ achieves the best empirical CR, but
unlike \ouralg,
the empirical advantage does not have any theoretical guarantees. This
is also partly because the empirical CR value can be affected
by a single bad problem instance and hence is more volatile.
For example,
when we increase the reliance of  \ouralglinear on the ML policy
by increasing $\gamma\in[0,1]$, the empirical of CR achieved by \ouralglinear
also increases quickly and becomes higher than that of Expert and \ouralgft.
Although the pure ML-based optimizer  achieves better average cost by leveraging historical data, its CR is significantly higher than Expert and even higher
than Greedy. 
\begin{table}[!h]
\centering
\scriptsize
\begin{tabular}{c|c|c|c|c|c|c|c|c} 
\toprule
\multirow{2}{*}{} & \multirow{2}{*}{\textbf{Expert}} & \multirow{2}{*}{\textbf{ML}} & \multirow{2}{*}{\textbf{HitOnly}} & \multirow{2}{*}{\textbf{Greedy}} & \multicolumn{4}{c}{\textbf{\ouralg}}\\ 
\cline{6-9}
    & & & & &  $\lambda=0.2$ & $\lambda=0.5$ & $\lambda=1$ & $\lambda=2$ \\ 
\hline
AVG& 134.61 & 108.09 & 139.07 & 200.61 & 115.17 & 109.59 & 108.11 & 107.92 \\ 
\hline
CR & 1.738 & 5.294& 8.993 & 3.264 & 1.843 & 2.174 & 2.535 & 3.617 \\ 
\cmidrule[\heavyrulewidth]{1-9}
\multirow{2}{*}{} & \multicolumn{4}{c|}{\textbf{\ouralglinear}} & \multicolumn{4}{c}{\textbf{\ouralgft}}\\ 
\cline{2-9}
    & $\gamma=0.1$ & $\gamma=0.3$ & $\gamma=0.5$ & $\gamma=0.9$ & 
 $\lambda=0.2$ & $\lambda=0.5$ & $\lambda=1$ & $\lambda=2$  \\ 
\hline
AVG & 127.25 & 115.68 & 108.29 & 106.04 & 113.76 & 106.96 & 105.33 & \textbf{105.06}  \\ 
\hline
CR & \textbf{1.668} & 1.914 & 2.506 & 4.587 & 1.812 & 1.972 & 2.451 & 3.032  \\
\bottomrule
\end{tabular}
\caption{Default case for a network with 3 nodes. 
The average cost of OPT in the testing dataset is 88.10. 
}\label{table:main}
\end{table}

By projecting the ML actions into carefully designed robust action sets,  
\ouralg can significantly reduce the CR compared to ML, while improving the average cost performance compared to Expert. 
Our analysis in Theorem~\ref{thm:avg_cost_blackbox} proves
that, with a larger $\lambda$, the average cost of \ouralg is closer to that of ML, while the guaranteed competitiveness becomes weaker.

Interestingly, in Table \ref{table:main} by setting $\lambda=2$, \ouralg can even achieve a lower average cost than ML, while still having a lower CR. The reason is that Expert performs much better than ML for some problem instances. Thus, the inclusion of Expert in \ouralg avoids those instances that would otherwise have a high cost if ML were used, and meanwhile a large $\lambda=2$ also provides
enough flexibility for \ouralg to exploit the benefits of ML in most other cases.
Moreover, by training an ML policy that is explicitly aware of the projection,
\ouralgft can further reduce the average cost compared to \ouralg while having
the same robustness guarantees. Additional results on different testing distributions are available in Appendix~\ref{sec:3_node_testing}.

\subsection{Results for Larger Networks}

\begin{figure*}[htp]
\subfigure[Complete graph]{\includegraphics[width=0.32\textwidth]{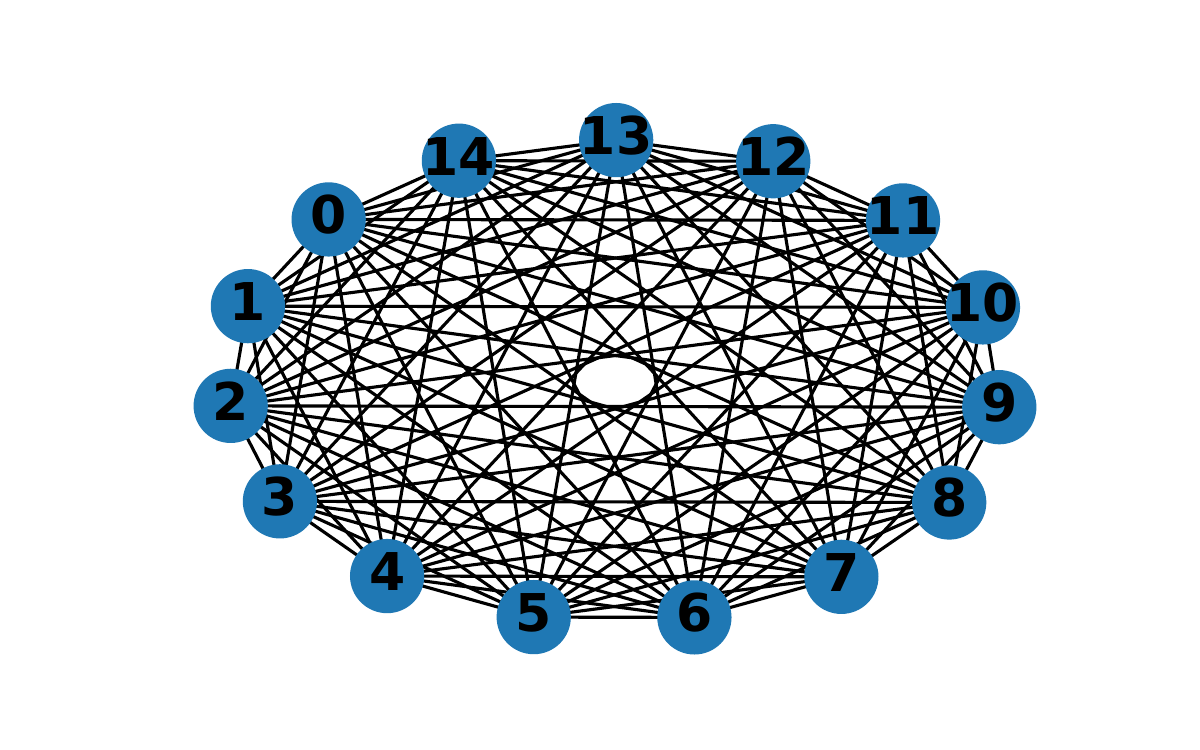}}
\subfigure[Star graph]{\includegraphics[width=0.32\textwidth]{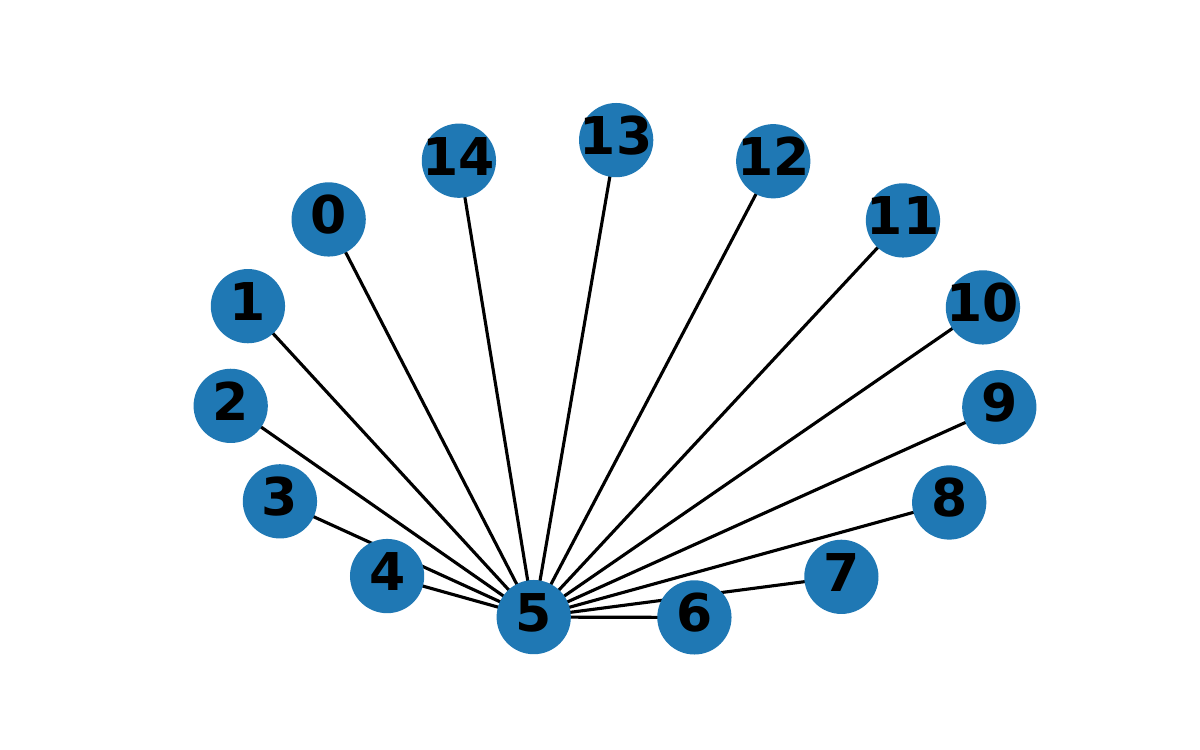}\label{fig:star_topology}}
\subfigure[Linear chain graph]{\includegraphics[width=0.32\textwidth]{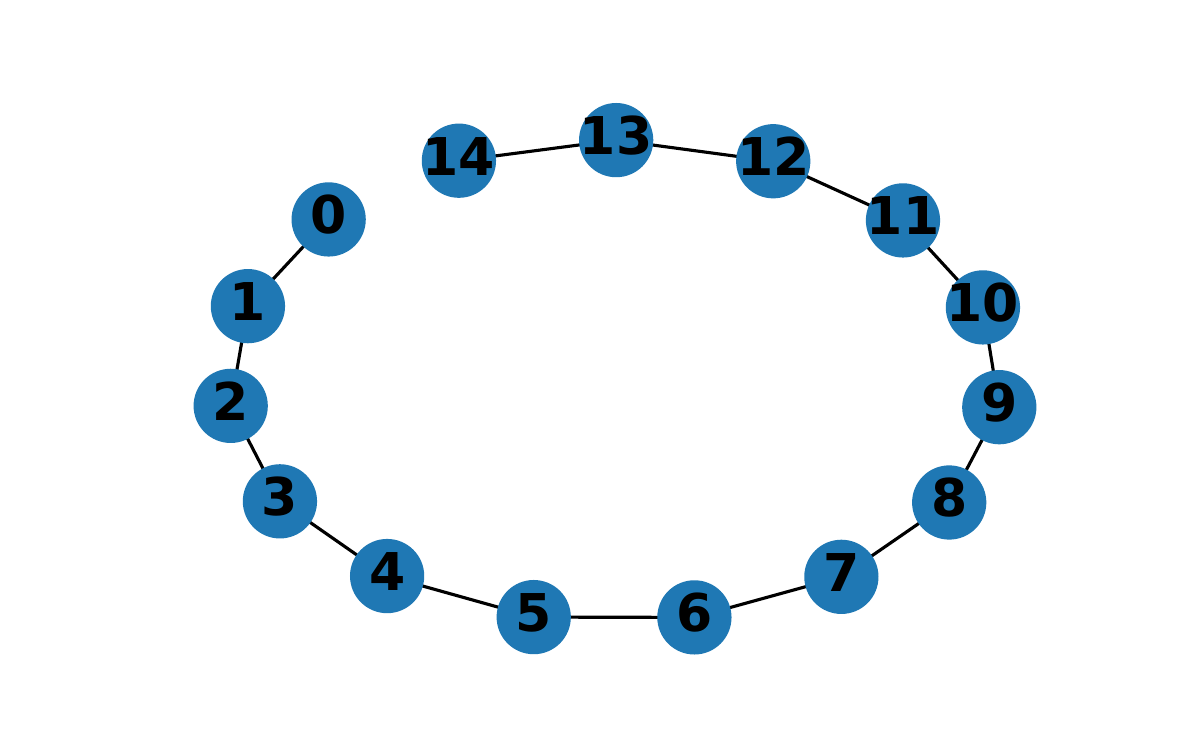}}\\
\subfigure[Individual Costs (complete graph)]{\includegraphics[width=0.31\textwidth]{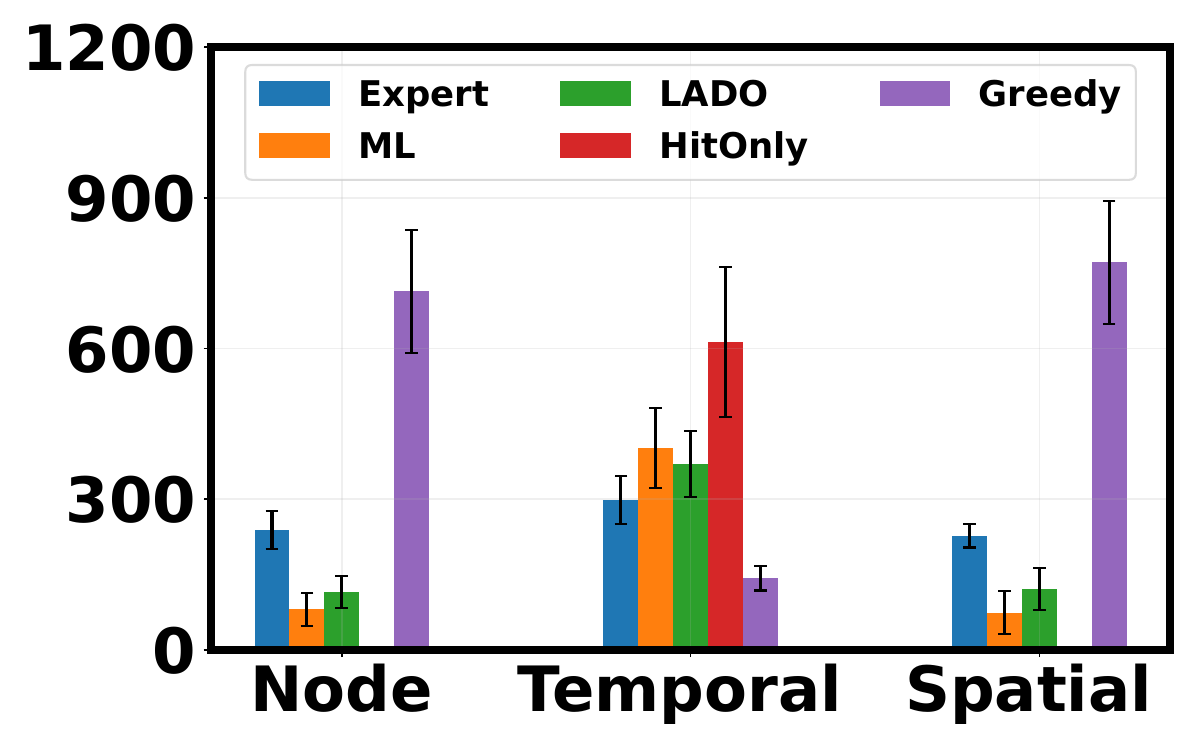}}
\subfigure[Individual Costs (star graph)]{\includegraphics[width=0.31\textwidth]{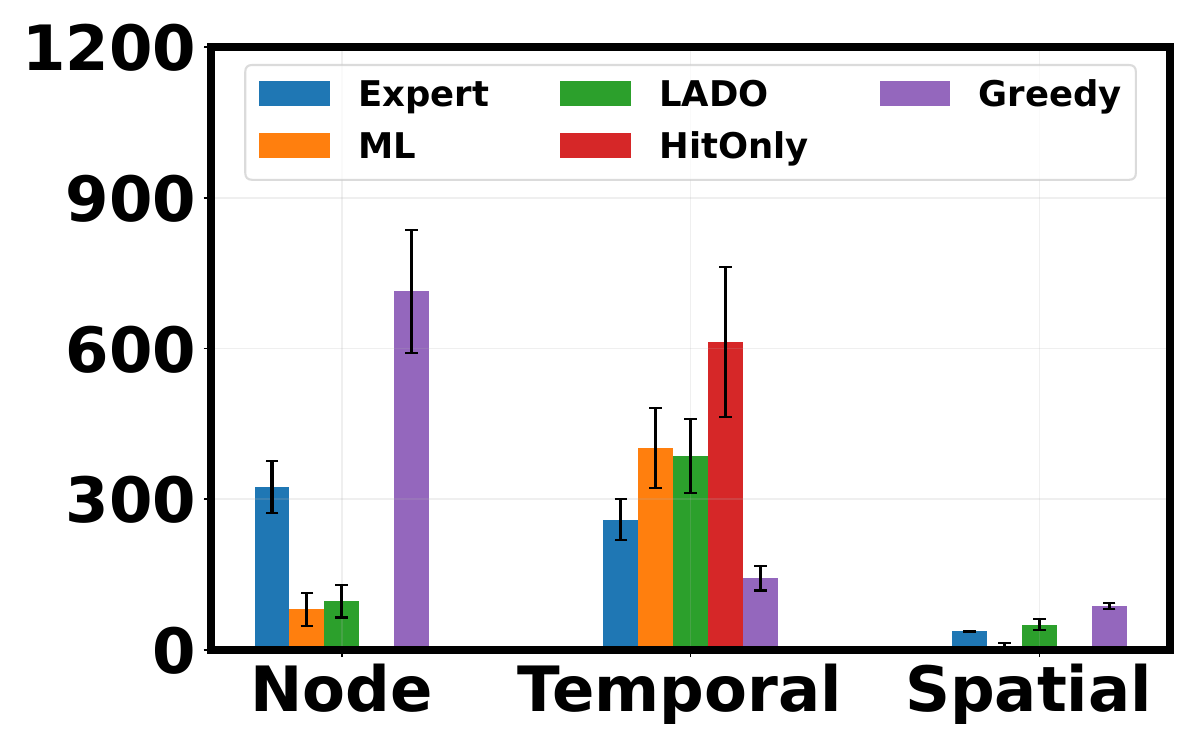}}
\subfigure[Individual Costs (chain graph)]{\includegraphics[width=0.31\textwidth]{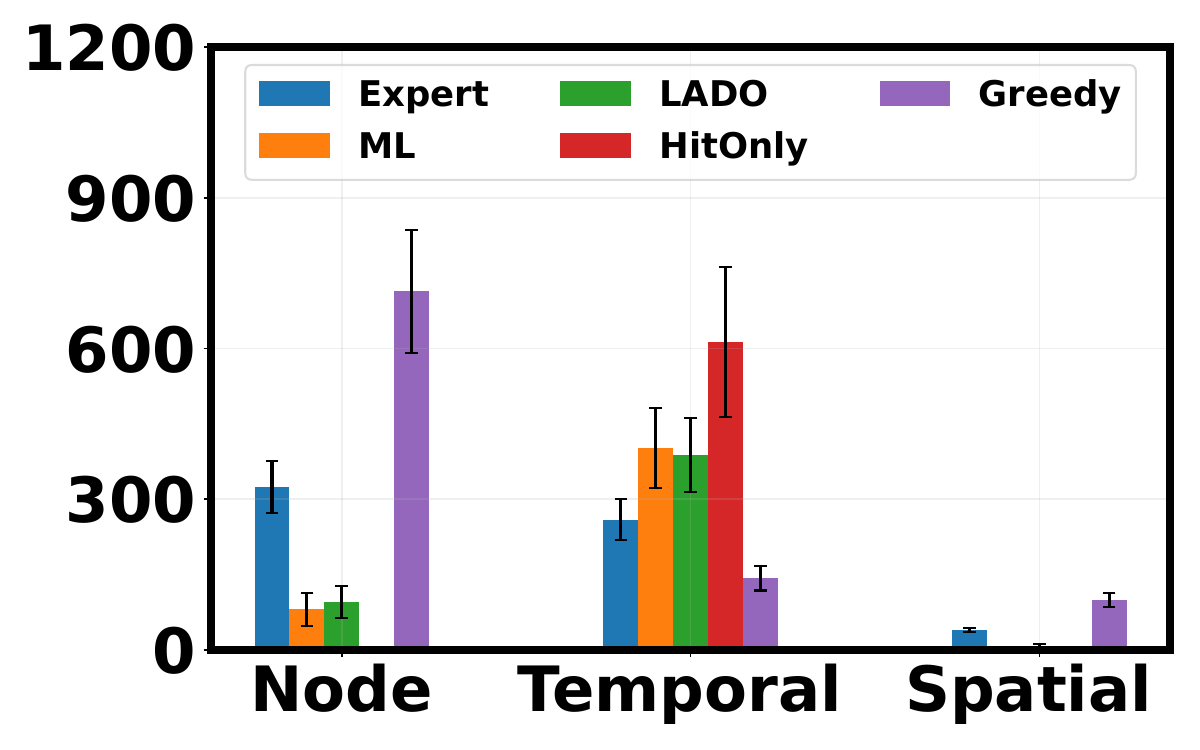}\label{fig:graph_topology_cost_loop_ind_main}}\\
\vspace{-0.4cm}
\caption{ The evaluation of \ouralg and baseline algorithms in terms of the node, temporal and spatial costs, with various graph topologies. By default, the competitiveness requirement $\lambda$ is set to 1  in \ouralg for all the graphs.}\label{fig:graph_topology_main}
\end{figure*}

We now consider larger networks with more nodes to assess \ouralg.
The setup is available in Appendix~\ref{appendix:large_network_setup}.
For three representative graph topologies 
(i.e., complete graph, star graph, and linear chain graph), 
the empirical average node, temporal, and spatial costs for each algorithm
in a 15-node network
are shown in 
Fig.~\ref{fig:graph_topology_main}. 

\revise{Notably, for these three graph topologies, the complete and star graphs have the same maximum node degree, while the star graph and linear chain graph share the same number of edges (or graph density). By comparing algorithm performance on these two pairs, we can gain insights into the impact of graph topologies.
For example, the significantly lower graph density in a star graph than
in a complete graph explains why all the algorithms considered exhibit lower total spatial costs on the star graph. This is consistent with our new theoretical analysis of the cost performance in Theorem~\ref{thm:avg_cost_blackbox}.
Compared to the complete graph, the spatial cost uncertainties are reduced due to fewer connections in the star graph. Thus, \ouralg can more effectively leverage the power of ML predictions, which leads to improved performance in both node cost and temporal cost of \ouralg.

In contrast to the linear chain graph, the star graph concentrates node degrees on a single node, resulting in distinct cost behaviors. As Theorem~\ref{thm:avg_cost_blackbox} indicates, the more uniform node degree distribution in the chain graph affords our algorithm greater flexibility to follow the ML policy by deviating more from the expert policy, ultimately reducing the cost increase term
$\omega_v(\lambda, \tilde{\pi}, \pi^\dagger)$.
Our empirical findings align with this theoretical analysis, demonstrating significantly lower overall costs for our algorithm in the linear chain graph due to its more effective exploitation of the ML policy.} As presented in  Fig.~\ref{fig:graph_topology_cost_loop_ind_main}, the effect of reduced spatial costs is more pronounced for \ouralg in the linear chain graph.

Next, we evaluate \ouralg on graphs with the same number of nodes but varying random graph topologies. 
Starting from the star graph, we gradually add random edges between nodes to increase the graph density until the graph is fully connected. 
Additionally, we also consider different competitiveness requirements in these graph topologies.

\begin{figure*}[h]
\subfigure[Total cost of \ouralg]{\includegraphics[width=0.46\textwidth]{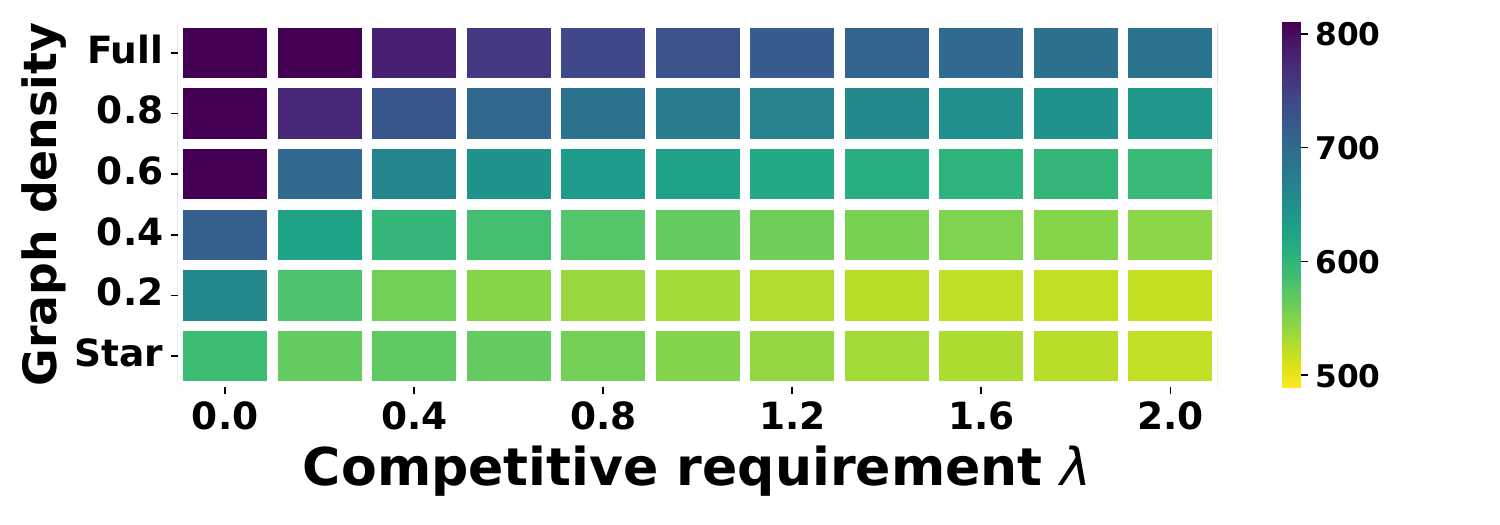}}
\subfigure[Regret of \ouralg compared to ML]{\includegraphics[width=0.46\textwidth]{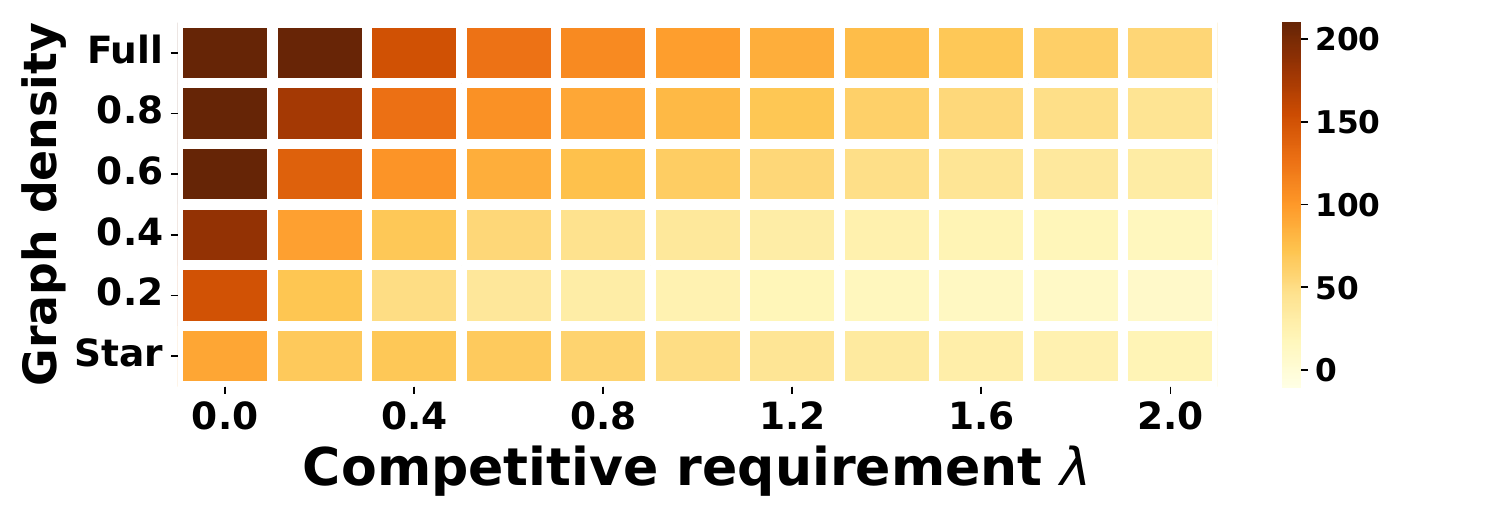}\label{fig:heatmap_response_regret_main}}
\caption{Impact of graph density and competitiveness requirement $\lambda$ on the overall cost of \ouralg, along with the additional cost (regret) associated with the projection process compared to the ML policy.}\label{fig:heatmap_main}
\end{figure*}

\revise{As shown in Fig.~\ref{fig:heatmap_main}, the total cost of \ouralg increases with graph density due to the additional spatial uncertainties introduced by denser connections. As Theorem~\ref{thm:avg_cost_blackbox} suggests, these increased spatial cost uncertainties lead to higher costs for \ouralg compared to the ML policy. To clarify this further, we compare the regret of \ouralg to the ML policy, defined as $\mathbb{E}_{g_{1:T}} \Bigl[ cost(\ouralg(\tilde{\pi})) - cost(\tilde{\pi}) \Bigr]$. This regret quantifies the additional cost incurred due to the projection process required for competitiveness guarantees.
As shown in Fig.~\ref{fig:heatmap_response_regret_main}, the regret of \ouralg typically increases with graph density, as denser graphs introduce more spatial cost uncertainties, making it more challenging to closely follow the ML policy. 
Similarly, with the same graph density (e.g. fully connected), the cost of \ouralg also increases as graph size increase from 3 to 120 nodes, shown as Fig.~\ref{fig:larger_graph_heatmap}.
This is because the growth of graph size naturally leads to more spatial connections, leading to elevated regret and overall cost, where detailed results can be found in Appendix~\ref{sec:larger_graph_exp}.
Moreover, stricter competitiveness requirements necessitate a closer adherence to the expert, further hindering \ouralg's ability to follow the ML policy and resulting in higher regret. }

\section{Related Work}

\textbf{Smoothed online convex optimization.}
Smoothed online convex optimization in a centralized single-agent setting is a classic problem for which many algorithms have been designed to bound the worst-case performance, e.g.,  \cite{SOCO_OBD_LQR_Abstract_Goel_Adam_Caltech_2019_10.1145/3374888.3374892,SOCO_OBD_R-OBD_Goel_Adam_NIPS_2019_NEURIPS2019_9f36407e,SOCO_Revisiting_Nanjing_NIPS_2021_zhang2021revisiting,competitive_control_memory_shi2020online,SOCO_Memory_FeedbackDelay_Nonlinear_Adam_Sigmetrics_2022_10.1145/3508037,SOCO_OBD_Niangjun_Adam_COLT_2018_DBLP:conf/colt/ChenGW18}. Recently, a growing literature has begun to study online convex optimization in a decentralized networked system \cite{Decentralized_OCO_CompressedCommunications_NeurIPS_2022_NEURIPS2022_dececdcb,decentralized_delay_cao2021decentralized,Decentralized_SOCO_YihengLin_Adam_ICML_2022_pmlr-v162-lin22c,Decentralized_NetworkedOnlineOptimization_TSP_7131577}. 
Compared to the centralized setting, the decentralized
setting is significantly more challenging, since an agent has no access to the information of other agents before making its action at each step. 
In this context, a recent work \cite{Decentralized_SOCO_YihengLin_Adam_ICML_2022_pmlr-v162-lin22c} proposes an online algorithm with a bounded competitive ratio and shows the dependency
of the competitive ratio on cost predictions. Several other studies \cite{Decentralized_NetworkedOnlineOptimization_TSP_7131577,decentralized_delay_cao2021decentralized, distributed_OCO_aggregative_li2021distributed,decentralized_OCO_relative_states_cao2021decentralized,distributed_OCO_dynamic_networks_hosseini2016online} propose algorithms with bounded regrets. In all cases, these studies focus on the worst-case performance, which leads to conservative algorithms that may not achieve a low average global cost. 
To address this limitation, in this work we exploit the benefits
of untrusted ML predictions to improve the average cost performance, while leveraging a robust policy  to achieve
the worst-case robustness.

\textbf{ML-based optimizers.}
ML policies have been used for exploiting
the statistical information and improve the average performance
of various (online) optimization problems, including scheduling, resource management, and secretary problems
\cite{L2O_NewDog_OldTrick_Google_ICLR_2019,L2O_Combinatorial_Reinforcement_AAAI_2020,L2O_Scheduling_DRL_Infocom_2019_8737649,Shaolei_LearningRobustCombinatorial_Zhihui_Infocom_2022}. There also exist ML-based optimizers,
such as 
multi-agent learning  \cite{zhang2018drl_collab,multi-agent_RL_overview_zhang2021multi,cooperative_MARL_review_oroojlooy2022review},
in the context of decentralized optimization where agents have limited or delayed communications. 
 However,  
 a crucial drawback of pure ML-based optimizers is
that they may have very high or even unbounded costs in the worst case, making
them unsuitable for mission-critical applications.
We provide an approach to empowering such ML-based algorithms with worst-case robustness guarantees.

\textbf{Learning-augmented algorithms.}
Learning-augmented algorithms have been proposed as a way to add worst-case robustness guarantees on top of ML policies 
 in a variety of settings, e.g., 
\cite{Shaolei_SOCO_RobustLearning_OnlineOpt_MemoryCosts_Infocom_2023,SOCO_ML_ChasingConvexBodiesFunction_Adam_COLT_2022,OnlineOpt_ML_Adivce_Survey_2016_10.1145/2993749.2993766,OnlineOpt_Learning_Augmented_RobustnessConsistency_NIPS_2020,OnlineOpt_ML_Adivce_Survey_2016_10.1145/2993749.2993766,primal_dual_learning_augmented_bamas2020primal,OnlineOpt_ML_Advice_CompetitiveCache_Google_JACM_2021_10.1145/3447579}.
To  guarantee
worst-case competitiveness, it is crucial to address the potential risks
associated with following the ML predictions, which is also the key challenge
for learning-augmented algorithm designs.
 More recently, learning-augmented algorithms have been designed for smoothed online optimization with switching costs
\cite{SOCO_MetricUntrustedPrediction_Google_ICML_2020_pmlr-v119-antoniadis20a,SOCO_OnlineOpt_UntrustedPredictions_Switching_Adam_arXiv_2022,SOCO_ML_ChasingConvexBodiesFunction_Adam_COLT_2022,Shaolei_L2O_ExpertCalibrated_SOCO_SIGMETRICS_2022}.
\revise{
However, learning-augmented algorithm designs in decentralized settings remain largely unexplored and are more challenging due to limited information availability. Thus, our study addresses this gap by introducing a novel, worst-case guaranteed learning-augmented algorithm specifically designed for decentralized environments. 
}

Our work differs from the standard constrained online optimization 
(e.g., \cite{OnlineOpt_Convex_LongTermConstraint_10.5555/2503308.2503322})
in that \ouralg is a {meta} algorithm leveraging
one robust policy to safeguard another policy
which can potentially perform better on average.
 Additionally, besides our novel decentralized setting and algorithm design
 based on spatial cost decomposition,
 \ouralg considers worst-case robustness
 and hence substantially differs from conservative
 bandits/reinforcement learning that focus
 on average or high-probability performance constraints \cite{conservative_bandits_wu2016conservative,conservative_RL_yang2021reduction,Conservative_RL_Facebook_AISTATS_2020_pmlr-v108-garcelon20a}.

\section{Concluding Remarks}

This paper studies 
 learning-augmented decentralized online convex optimization in networks.
We begin with \ouralglinear by linearly combining the ML policy
and the expert policy. It is proved
that, while \ouralglin can exploit the potential of ML to improve
the average cost performance, it does not have guaranteed
worst-case performance. 
Then, we propose \ouralg, a novel algorithm 
that improves the average performance while
guaranteeing worst-case robustness.
\ouralg addresses the key challenges of temporal and spatial
information inefficiency and constructs novel robust action sets
that allow agents to choose individual actions based on local online information.
We prove bounds on
the guaranteed competitiveness and the average performance of \ouralg. 
Finally,
we run an experiment of decentralized battery management
for sustainable computing. Our results highlight the potential of ML augmentation
to improve the average performance in \ouralglinear and \ouralg as well as the guaranteed worst-case performance
of \ouralg.

{
    \bibliographystyle{unsrt}
    \bibliography{main.bbl}
}

\newpage
\appendix
\section{Extension of \ouralg to Directed Graphs}\label{appendix:extension_directed_graph}

In some real-world applications (e.g., wireless networks), the connections between nodes are directional, instead of the bi-directional connections in undirected graphs. Thus, we extend \ouralg
to a directed graph setting. We first show how to modify the design of adaptive spatial cost splitting and reservation cost for a directed graph,
followed by an average cost performance bound.

Consider a network with a finite set $\mathcal{V}$ of nodes. We model
the network as a directed graph, denoted by $(\mathcal{V}, \mathcal{E})$, where $\mathcal{E}$ represents the set of directional edges between nodes in $\mathcal{V}$. For each edge $(v,u) \in \mathcal{E}$, the spatial cost is denoted as $s_{t}^{(v,u)} (x_{t}^v, x_t^{u})$, which depends on the actions of the node $v$ and $u$ at time $t$. Since the edge is directional, the spatial cost $s_{t}^{(v,u)} (x_{t}^v, x_t^{u})$ may not be equal to the cost $s_{t}^{(u,v)} (x_{t}^u, x_t^{v})$ incurred in the opposite direction. 

In a directed graph, the locally computable constraint in Eqn.~\eqref{eqn:local_constraint} for $\lambda$-competitiveness
can be rewritten as below
\begin{equation}\label{eqn:directed_graph_constraint}
\begin{aligned}
&\sum_{\tau=1}^t f_\tau^{v}(x_\tau^{v}) + \sum_{\tau=1}^t  c_\tau ^{v}(x_\tau^{v}, x_{\tau-1}^{v})+ \sum_{\tau=1}^{t-1} \Bigl( \sum_{(v, u) \in \mathcal{E}}\kappa_{\tau}^{(v,u)} \cdot {s_{\tau}^{(v, u)}(x_{\tau}^{v}, x_{\tau}^u)}  + \sum_{(u,v) \in \mathcal{E}}\kappa_{\tau}^{(u,v)} \cdot {s_{\tau}^{(u,v)}(x_{\tau}^{u}, x_{\tau}^{v})} \Bigr) \\
& + R(x_t^v,x_t^{v,\dagger})   \leq  (1+\lambda) \biggl(\sum_{\tau=1}^t f_\tau^{v}(x_\tau^{v, \dagger})  + \sum_{\tau=1}^t c_\tau ^{v}(x_\tau^{v, \dagger}, x_{\tau-1}^{v, \dagger}) +   \sum_{\tau=1}^{t-1} \Bigl( \sum_{(v, u) \in \mathcal{E}} \kappa_{\tau}^{(v,u)} \cdot {s_{\tau}^{(v, u)}(x_{\tau-1}^{v, \dagger}, x_{\tau}^{u, \dagger})}  \\
& + \sum_{(u,v) \in \mathcal{E}} \kappa_{\tau}^{(u,v)} \cdot {s_{\tau-1}^{(u,v)}(x_{\tau}^{u, \dagger}, x_{\tau}^{v, \dagger})}  \Bigr) \biggr),
\end{aligned}
\end{equation}
where the weight $\kappa_{\tau}^{(v,u)}$ for splitting the spatial cost $s_{\tau}^{(v,u)}$ is adaptively chosen as
\begin{equation}
    \kappa_{\tau}^{(v,u)} = \frac{\|x_{\tau}^{v} - x_{\tau}^{v, \dagger} \|^2}{\|x_{\tau}^{v} - x_{\tau}^{v, \dagger} \|^2 + \|x_{\tau}^{u} - x_{\tau}^{u, \dagger} \|^2}
\end{equation}
and the reservation cost is 
\begin{equation}
    R(x_t^v, x_{t}^{v,\dagger}) = \frac{\ell_T + \ell_S \cdot (D_v^{in} + D_v^{out})}{2} (1+ \frac{1}{\lambda_0} \|x_t^v - x_t^{v,\dagger}\|^2)
\end{equation}
where $D_v^{in}$ and $D_v^{out}$ denote the in-degree and out-degree of node $v$. In other words,  $D_v^{in}$ represents the number of edges directed towards node $v$ and $D_v^{out}$ is the number of edges directed from node $v$.
Additionally, based on  Assumptions~\ref{assumption:node_cost} - \ref{assumption:spatial_cost}, the temporal and spatial costs are $\ell_T$- and $\ell_S$-smooth with respect to the actions, respectively.

If there exist bi-directional edges between node $u$ and $v$, the weights for splitting spatial costs $s_{t-1}^{(v,u)}$ and $s_{t-1}^{(u,v)}$ are identical. This is because the weight $\kappa^{(v,u)}_{\tau}$ allocates the spatial cost according to the potential risk of spatial cost increases
due to nodes $v$ and $u$, as measured by the distances between
their actions to the expert advice. Since the risk is independent of the direction, the spatial cost splitting weight $\kappa_{\tau}^{v,u}$ remains the same regardless of the edge direction.

Next, we analyze the average cost of \ouralg in a directed graph.

\begin{corollary}\label{cor:avg_cost_blackbox_direct}
(Average Cost of $\ouralg(\tilde{\pi})$ for directed graph) Given any ML policy $\tilde{\pi}$, for any $\lambda > 0$, by optimally setting $\lambda_0 = \sqrt{1+\lambda} - 1$, the average cost of $\ouralg(\tilde{\pi})$ is  bounded by
\begin{equation}
\nonumber
\begin{aligned}
        AVG(\ouralg(\tilde{\pi}))\leq \min \left\{ (1+\lambda) AVG(\pi^{\dagger}), \biggl( \sqrt{\text{AVG}(\tilde{\pi})}  + \sqrt{    \sum_{v\in \mathcal{V}}\omega_v(\lambda, \tilde{\pi}, \pi^{\dagger})  } \biggr)^2 \right\},
    \end{aligned}
\end{equation}
where $AVG(\pi^{\dagger})$ and $AVG(\tilde{\pi})$ are the average costs of the expert policy and the  ML policy, respectively, and 
$\omega_v(\lambda, \tilde{\pi}, \pi^{\dagger}) = \mathbb{E}_{g_{1:T}}\left\{\sum_{t=1}^T \left[ \frac{\ell_f + 2\cdot\ell_T + \ell_S\cdot (D_v^{in} + D_v^{out})}{2}\|\tilde{x}^v_t - x_t^{v,\dagger}\|^2 -  (\sqrt{1+\lambda} - 1)^2  \text{cost}_{v,t}^\dagger \right]^+ \right\}$ in which 
$\text{cost}_{v,t}^\dagger$ is the expert's node and temporal cost for node $v$ at time $t$. 
\end{corollary}

In Corollary~\ref{cor:avg_cost_blackbox_direct}, 
the average cost of \ouralg is bounded by the expert's average cost scaled by $(1+\lambda)$ and the average cost of learning-based policy $\tilde{\pi}$ along with an additional cost introduced by the projection process, corresponding to the two terms in the min operators, respectively. 
When relaxing the parameter $\lambda$ in the competitiveness guarantee, it grants more freedom for \ouralg to follow the ML policy with less restrictive constraints, resulting in a smaller $\omega_v(\lambda, \tilde{\pi}, \pi^\dagger)$ for all nodes in $\mathcal{V}$. Additionally, the node degree plays an important role, along with the overall action distance between the expert and ML policies. 
Importantly, the spatial cost is shared by connected nodes, according to the adaptive spatial cost splitting weight $\kappa_t^{(v,u)}$. Therefore, the spatial uncertainty of node $v$ is determined jointly by the spatial connections originating from and directed towards that specific node.
For the nodes with greater spatial uncertainties (quantified by the sum of in- and out-degrees) and/or larger policy misalignment in terms of $\|\tilde{x}_t^v - x_t^{v,\dagger}\|$, it is more challenging to adopt the ML policy, potentially incurring a higher cost during the projection. As the graph density increases, more spatial uncertainties are introduced to the connected nodes, thus increasing $\omega_v(\lambda, \tilde{\pi}, \pi^{\dagger})$ incurred by the projection process. However, with the competitiveness guarantee, the average cost of \ouralg can always be bounded by the expert's cost up to a scaling factor of $(1+\lambda)$, regardless of the graph topology or the chosen ML policy.

\section{Optimal Projection-Aware ML Training}\label{sec:end_to_end_ml}

Theorem~\ref{thm:avg_cost_blackbox} applies to any ML models, including
 ML models that are trained as standalone optimizers without considering
 the design of \ouralg and hence may have training-testing objective mismatches.
To improve the average cost performance,
we consider the following ML policy $\tilde{\pi}^{\circ}_{\lambda}$
that is optimally trained with explicit consideration of the 
downstream projection in \ouralg:
\begin{equation}\label{eqn:optimal_constrained_ML_policy}
\tilde{\pi}^{\circ}_{\lambda}=\arg\min_{\pi}\mathbb{E}_{g_{1:T}}\left[cost(\ouralg({\pi}),g_{1:T} )\right],
\end{equation} 
where the projected ML prediction by \ouralg is explicitly used  as
the action in the cost. 
The policy $\tilde{\pi}^{\circ}_{\lambda}$ can be trained offline using 
implicit differentiation (i.e., the added projection in Line~4 for \ouralg in Algorithm~\ref{alg:RP-OBD} 
can be implicitly differentiated based on KKT conditions) \cite{L2O_DifferentiableConvexOptimization_Brandon_NEURIPS2019_9ce3c52f}.
Like in other learning-augmented algorithms \cite{SOCO_ML_ChasingConvexBodiesFunction_Adam_COLT_2022}, 
we consider that $\tilde{\pi}^{\circ}_{\lambda}$ is already available for online inference by individual agents.
Next, we use $\ouralg(\tilde{\pi}^{\circ}_{\lambda})$ to emphasize
the usage of $\tilde{\pi}^{\circ}_{\lambda}$ in \ouralg, and
show its average cost bound.  
The proof is deferred to Appendix~\ref{sec:cost_ratio_e2e_proof}.

\begin{corollary}[Average cost of $\ouralg(\tilde{\pi}^{\circ}_{\lambda})$]\label{thm:avg_cost_e2e}
Given the optimal projection-aware ML policy $\tilde{\pi}^{\circ}_{\lambda}$,
for any $\lambda > 0$, by optimally setting $\lambda_0 = \sqrt{1+\lambda} - 1$, the average cost of $\ouralg(\tilde{\pi}^{\circ}_{\lambda})$ is upper bounded by
\begin{equation}\label{eqn:avgcostbound_new}
    \begin{aligned}
       AVG(\ouralg(\tilde{\pi}^{\circ}_{\lambda}))\leq \min \biggl\{ &(1-\alpha_\lambda)AVG(\pi^{\dagger}) +\alpha_\lambda AVG(\tilde{\pi}^*), \\
       &\biggl( \sqrt{\text{AVG}(\tilde{\pi}^*)}  + \sqrt{  \sum_{v\in \mathcal{V}}\omega_v(\lambda, \tilde{\pi}^*,\pi^{\dagger}) } \biggr)^2 \biggr\}
    \end{aligned}
\end{equation}
where  $AVG(\pi^{\dagger})$ and $AVG(\tilde{\pi}^*)$ are the average costs of the expert and the optimal projection-unaware ML policy 
$\tilde{\pi}^*=\arg\min_{\pi}\mathbb{E}_{g_{1:T}}\left[cost(\pi,g_{1:T} )\right]$, 
respectively,  $\alpha_\lambda = \min\left\{(\sqrt{1+\lambda} - 1)\sqrt{\frac{2}{{\ell_T +  \ell_f +  D_{\max}\cdot\ell_S}} \cdot \hat{C}} , 1\right\} $ 
with $D_{\max}=\max_{v\in\mathcal{V}}D_v$ being the maximum degree in the network
and the expert's minimum single-node cumulative cost normalized by the cumulative expert-ML action distance is defined as $\hat{C} = \min_{g_{1:T} \in \mathcal{G}}\min_{v \in \mathcal{V}, t \in [1,T]}{\frac{\text{cost}_{v}(x_{1:t}^{v,\dagger})}{\sum_{i=1}^t \|x_i^{v,\dagger} - \tilde{x}_i^{v,*} \|^2}}$. 
 Besides, we define 
 \begin{equation}
    \omega_v(\lambda, \tilde{\pi}^*,\pi^{\dagger}) = \sum_{t=1}^T \mathbb{E}_{g_{1:T}}\left\{\left[  \frac{\ell_f + 2\cdot\ell_T + \ell_S\cdot D_{v}}{2} \|\tilde{x}^{v,*}_t - x_t^{v,\dagger}\|^2 - {  (\sqrt{1+\lambda} - 1)^2 }\cdot  \text{cost}_{v,t}^\dagger  \right]^+\right\},  
 \end{equation}
  where $\text{cost}_{v,t}^\dagger$ is the expert's node and temporal cost for node $v$ at time $t$.
\end{corollary}

Corollary~\ref{thm:avg_cost_e2e} formally demonstrates the 
benefits of using the optimal projection-\emph{aware} ML policy  \eqref{eqn:optimal_constrained_ML_policy} compared
to the optimal projection-\emph{unaware} ML policy 
$\tilde{\pi}^*=\arg\min_{\pi}\mathbb{E}_{g_{1:T}}\left[cost(\pi,g_{1:T} )\right]$.
Specifically, 
by the optimality
of $AVG(\tilde{\pi}^*)$
that does not consider 
$\lambda$-competitiveness,
we naturally have $AVG(\tilde{\pi}^*)\leq AVG(\pi^{\dagger})$.
Thus, the first term in \eqref{eqn:avgcostbound_new}
shows that,
the average cost of $\ouralg(\tilde{\pi}^{\circ}_{\lambda})$ 
is no greater than the expert by using the projection-aware
ML policy $\tilde{\pi}^{\circ}_{\lambda}$. This is because 
the expert policy is intuitively a feasible solution
in our $\lambda$-competitiveness ML policy space, while
the policy $\tilde{\pi}^{\circ}_{\lambda}$ in \eqref{eqn:optimal_constrained_ML_policy} 
is the optimal one that specifically
minimizes the average cost of
$\ouralg(\tilde{\pi}^{\circ}_{\lambda})$.
 By contrast,
 even 
 by using the optimal projection-unaware ML 
 policy
 $\tilde{\pi}^*$, 
the average cost of $\ouralg(\tilde{\pi}^*)$ is bounded by 
$(1+\lambda) AVG(\pi^{\dagger})$ in the first term of Theorem~\ref{thm:avg_cost_blackbox}, since the added projection during
actual inference
can void the optimality of $\tilde{\pi}^*$ and 
result in a higher 
average cost up to $(1+\lambda$ times of the expert's cost.
The root reason for the advantage of
 the optimal projection-\emph{aware} 
 ML policy  \eqref{eqn:optimal_constrained_ML_policy} in terms of the average
 cost is that
its ML prediction is specifically customized to  \ouralg.
On the other hand, even though
$\tilde{\pi}^*=\arg\min_{\pi}\mathbb{E}_{g_{1:T}}\left[cost(\pi,g_{1:T} )\right]$
is the optimal-\emph{unconstrained} ML policy on its own,
its optimality can no longer hold
when modified by \ouralg for $\lambda$-competitiveness during actual online inference.

Finally, the second term
inside
 $\min$  in Corollary~\ref{thm:avg_cost_e2e}
shows that  the average
cost of 
$\ouralg(\tilde{\pi}^{\circ}_{\lambda})$ with
the optimal projection-aware ML policy
$\tilde{\pi}^{\circ}_{\lambda}$ in \eqref{eqn:optimal_constrained_ML_policy} is
 upper bounded by that of $\ouralg(\tilde{\pi}^{*}_{\lambda})$,
since $\ouralg(\tilde{\pi}^{*}_{\lambda})$ is a feasible policy
satisfying $\lambda$-competitiveness by our design.
Like in Theorem~\ref{thm:avg_cost_blackbox},
it reinforces the insight
that $\ouralg$ can
better exploit the potential
of ML predictions  for average performance improvement
when $\lambda>0$ increases.

\section{Additional Experiments for Decentralized Battery Management}\label{appendix:experiment_battery}

In this section, we present more results
on different testing distributions in 
3-node network and an extended set of experiments for large networks incorporating a wider variety of battery units. Beyond the self-degradation coefficient $A_v$, each battery unit also exhibits unique characteristics in terms of the storage capacity and maximum continuous discharge current. 
These values for the battery units are derived from publicly available data on energy storage systems. We further investigate the impact of network topology, on the overall cost of \ouralg and other baseline algorithms.

\subsection{ML Model Architecture and More Results for Networks with 3
Nodes}\label{sec:3_node_testing}

\subsubsection{ML Model Architecture}
The ML predictions used in our algorithm are computed using a RNN with 2 recurrent layers, each with 8 hidden features. In all of our experiments, each problem instance spans 24 hours, and
each time step represents one hour. 
For the training processes, we used the net energy demand trace from the first two months of 2015, which contains 1440 hourly data samples and produces a total of 1416 24-hour sequences. The ML model is optimized by Adam with a learning rate of $10^{-3}$ for 60 epochs in total. 
After training, the weights of the ML model are shared between all nodes with different coefficients $A_v$.
On average, the training process takes 3 minutes on a 2020 MacBook Air with 8GB memory, and our testing process takes about 1 second.  
For testing, we use the net demand traces from April to March in the default case. 

\subsubsection{Results}

Based on the setup in Section~\ref{section:case_study}, 
we choose 3 fully connected battery units, where the self-degradation coefficients $A_v$ of the battery units are set as $0.9$, $0.93$, $0.95$, respectively.
We present more results on different testing distributions in 
3-node network.

\textbf{In-distribution testing.} 
First, we consider an ideal case, called in-distribution testing,
where the ML model is trained and tested on the same data distribution.
Naturally, the ML model is expected to perform very well.
Table~\ref{table:dist_aligned} shows that the ML model outperforms
the expert in terms of the empirical CR.
By increasing $\gamma$, \ouralglinear follows the ML more closely
and hence also achieves a better average cost.
Nonetheless, its advantage in terms of the average cost comes
without competitiveness guarantees. By contrast, \ouralg and
\ouralgft can achieve both good average costs and guaranteed competitiveness simultaneously.

\begin{table}[htp]
\centering
\scriptsize
\begin{tabular}{c|c|c|c|c|c|c|c|c} 
\toprule
\multirow{2}{*}{} & \multirow{2}{*}{\textbf{Expert}} & \multirow{2}{*}{\textbf{ML}} & \multirow{2}{*}{\textbf{HitOnly}} & \multirow{2}{*}{\textbf{Greedy}} & \multicolumn{4}{c}{\textbf{\ouralg}}\\ 
\cline{6-9}
    & & & & &  $\lambda=0.2$ & $\lambda=0.5$ & $\lambda=1$ & $\lambda=2$\\ 
\hline
AVG & 127.79 & 95.45 & 119.06 & 192.11 & 106.53 & 98.93 & 96.41 & 95.66 \\ 
\hline
CR & 1.738 & 1.487 & 2.021 & 3.264 & 1.431 & 1.446 & 1.490 & 1.488 \\ 
\cmidrule[\heavyrulewidth]{1-9}
\multirow{2}{*}{} & \multicolumn{4}{c|}{\textbf{\ouralglinear}} & \multicolumn{4}{c}{\textbf{\ouralgft}}\\ 
\cline{2-9}
 & $\gamma=0.1$ & $\gamma=0.3$ & $\gamma=0.5$ & $\gamma=0.9$ & 
 $\lambda=0.2$ & $\lambda=0.5$ & $\lambda=1$ & $\lambda=2$  \\ 
\hline
AVG & {120.20} & 107.91 & 99.50 & \textbf{94.32} & 105.33 & 97.46 & 95.47 & {95.00} \\ 
\hline
CR & {1.668} & 1.546 & 1.449 & 1.446 & 1.420 & \textbf{1.400} & 1.416 & 1.420  \\
\bottomrule
\end{tabular}
\caption{In-distribution testing for a 3-node network. 
The average cost of OPT  is 83.37.  
}\label{table:dist_aligned}
\end{table}

\textbf{Out-of-distribution testing.} Next,
we inject large Gaussian noise into the testing dataset
and consider the out-of-distribution testing case where
the testing and training distributions are different. 
The results are shown in Table~\ref{table:less_align}.
In this case, the ML model has a higher average cost
as well as empirical CR than Expert. While \ouralglinear
can reduce the average cost by slightly incorporating
the ML prediction into its action ($\gamma=0.1$),
this advantage quickly vanishes as $\gamma$ increases.
On the other hand, by training the ML model in a projection-aware
manner, \ouralgft can keep its
average cost low while still offering guaranteed competitiveness.

\begin{table}[htp]
\centering
\scriptsize
\begin{tabular}{c|c|c|c|c|c|c|c|c} 
\toprule
\multirow{2}{*}{} & \multirow{2}{*}{\textbf{Expert}} & \multirow{2}{*}{\textbf{ML}} & \multirow{2}{*}{\textbf{HitOnly}} & \multirow{2}{*}{\textbf{Greedy}} & \multicolumn{4}{c}{\textbf{\ouralg}}\\ 
\cline{6-9}
    & & & & & $\lambda=0.2$ & $\lambda=0.5$ & $\lambda=1$ & $\lambda=2$ \\ 
\hline
AVG & 141.08 & 188.86 & 320.20 & 200.03 & 138.93 & 156.71 & 174.00 & 184.80 \\ 
\hline
CR & 1.623 & 5.452 & 10.488 & 2.748 & 1.808 & 2.188 & 2.857 & 3.969\\ 
\cmidrule[\heavyrulewidth]{1-9}
\multirow{2}{*}{} & \multicolumn{4}{c|}{\textbf{\ouralglinear}} & \multicolumn{4}{c}{\textbf{\ouralgft}}\\ 
\cline{2-9}
    & $\gamma=0.1$ & $\gamma=0.3$ & $\gamma=0.5$ & $\gamma=0.9$ & 
 $\lambda=0.2$ & $\lambda=0.5$ & $\lambda=1$ & $\lambda=2$   \\ 
\hline
AVG & \textbf{136.73} & 134.12 & 139.62 & 174.95 & 137.60 & 145.93 & 150.89 & 152.98\\ 
\hline
CR & \textbf{1.605} & 1.881 & 2.501 & 4.706 & 1.783 & 2.198 & 2.650 & 3.162\\
\bottomrule
\end{tabular}
\caption{Out-of-distribution testing for a 3-node network. The average cost of OPT  is 95.77. } \label{table:less_align}
\end{table}

\subsection{Results for Large Networks}\label{sec:larger_graph_exp}
We present an extended set of experiments for large networks utilizing a variety of battery units.

\subsubsection{Experimental Setup}\label{appendix:large_network_setup}
Following a similar experimental setup in the experiment of a 3-node network, the data center's energy demand $P_{d,t}$ is derived using the hourly workload trace from \cite{Microsoft_Water_2022_how_azure_cloud} and the renewable energy generation $P_{r,t}$ is estimated with the weather-related statistics from \cite{sengupta2018national} . 
The net energy demand of the data center, $P_{n,t} = P_{d,t} - P_{r,t}$ is then served by the energy storage system, powered by a pool of battery units.
Then, we use a sliding window to generate 24-hour net demand sequences as the datasets, where each sequence has 25 successive normalized net demands (from hour 0 to hour 24).
In this experiment, we derive the storage capacity and rated output power from five different commercially available battery storage systems, including Tesla Powerwall+, LG ESS, Solar Edge, Enphase IQ and FranklinWH, where the detailed specifications can be found in Table \ref{table:detailed_specs}. 
We normalize their storage capacity and continuous output power relative to the Tesla Powerwall+ for easy comparison. 
For an easy  comparison, their relative storage capacities are 1, 1.07, 0.72, 0.78, 1.01, and their relative continuous output powers are 1, 1.07, 0.71, 0.55, 0.71, respectively. To account for variations in the battery health of these battery units, we employ three self-degradation coefficients $A_v$, set as 0.9, 0.93, 0.95, consistent with the main experiment. By default, we set $b=5$ and $c=2$ as the weights for the temporal and spatial costs, respectively.  In total, by considering the various battery characteristics along with the self-degradation coefficients, we create 15 distinct battery node configurations for our experiment.

\begin{table}
\centering
\scriptsize
\begin{tabular}{l|c|c|c|c} 
\hline
  & \begin{tabular}[x]{@{}c@{}}Usable battery \\ capacity (kWh) \end{tabular}
   & \begin{tabular}[x]{@{}c@{}}Rated Charging \\ Power (kW) \end{tabular}
    & \begin{tabular}[x]{@{}c@{}} Rated Discharging \\ Power(kW) \end{tabular}
    & \begin{tabular}[x]{@{}c@{}} Peak Discharge \\ Power (kW) \end{tabular}\\ 
\hline
Tesla Powerwall 2 \cite{Tesla_powerwall_spec}     & 13.50  & 5.00   & 5.00 & 7.00     
\\
\hline
LG ESS (Home 8) \cite{LG_ESS_spec} & 14.40   & 5.40   & 7.50      & 9.00   \\ 
\hline
Solar Edge (BAT-10K1P) \cite{SolarEdge_spec} & 9.70   & 5.00    & 5.00 & 7.50   \\
\hline
IQ Battery 10T \cite{IQ_battery_spec}  & 10.08  & 3.84  & 3.84     & 5.76  \\ 
\hline
FranklinWH \cite{FranklinWH_spec}     & 13.60   & 5.00     & 5.00 & 10.00    \\ 
\hline
\end{tabular}
\caption{Specifications of the commercially available home energy storage systems used in the experiment.} \label{table:detailed_specs}
\end{table}

\begin{figure*}[t]
\subfigure[Total cost (complete graph)]{\includegraphics[width=0.31\textwidth]{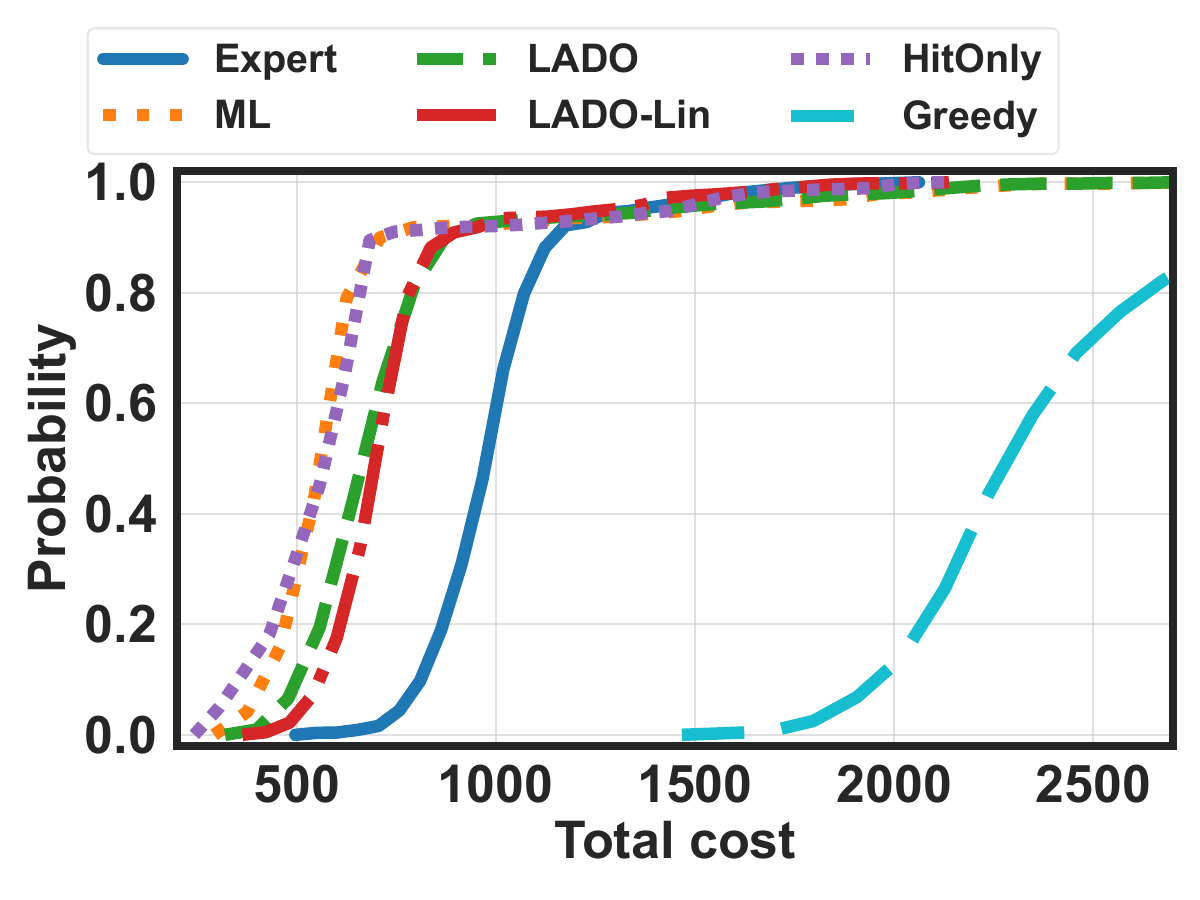}\label{fig:graph_topology_cost_full}}
\subfigure[Total cost (star graph)]{\includegraphics[width=0.31\textwidth]{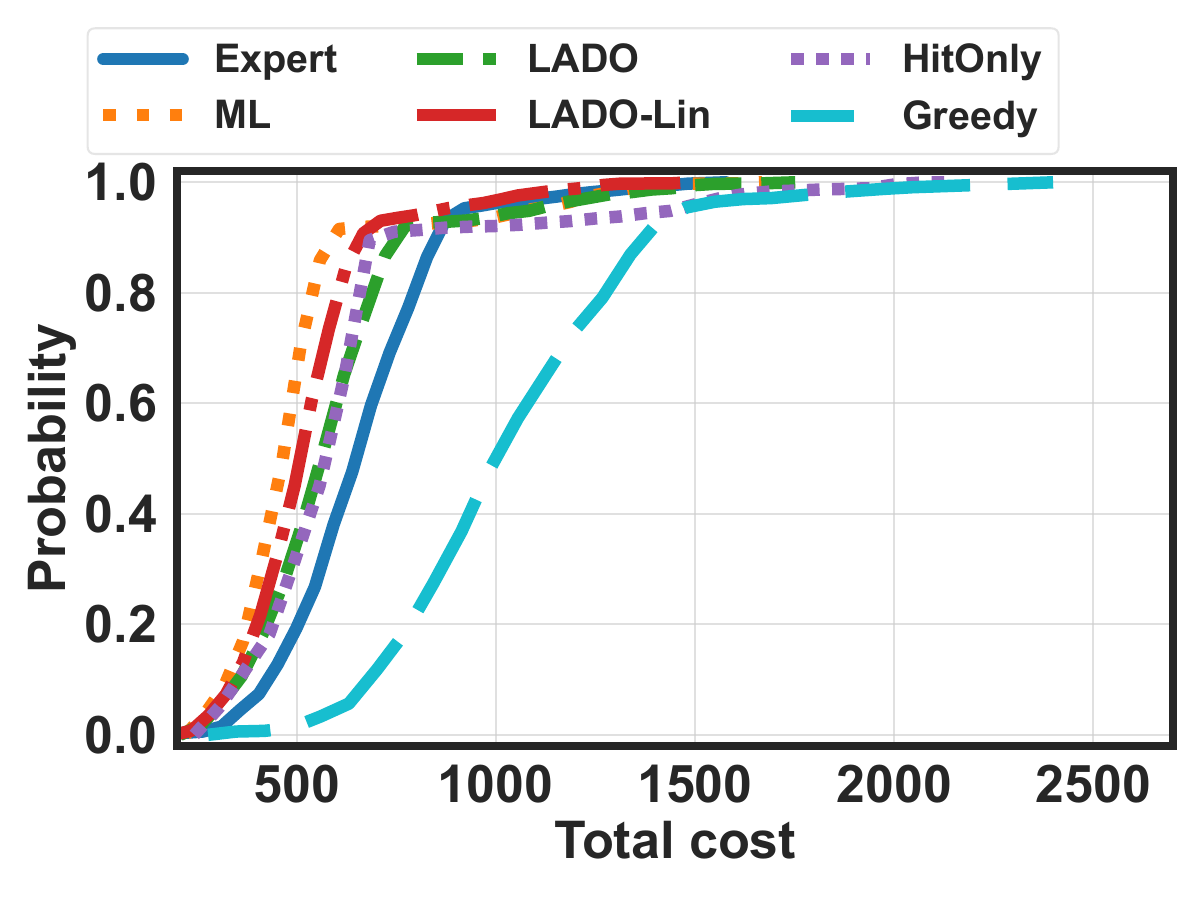}\label{fig:graph_topology_cost_star}}
\subfigure[Total cost (chain graph)]{\includegraphics[width=0.31\textwidth]{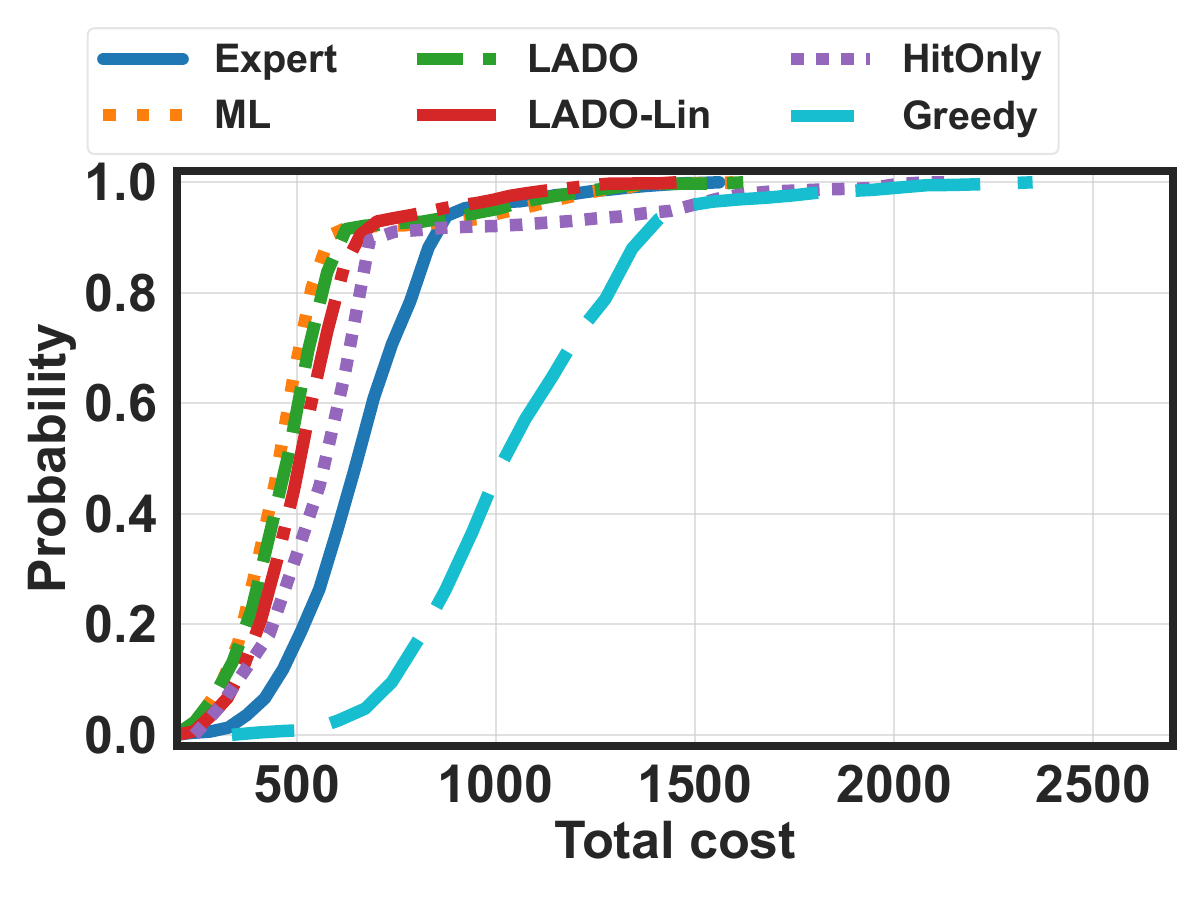}\label{fig:graph_topology_cost_loop}}
\caption{The total cost distribution of \ouralg and baseline algorithms with various graph topologies (15-node network). By default, the competitiveness requirement $\lambda$ is set to 1  in \ouralg for all graphs. }\label{fig:graph_topology}
\end{figure*}

\subsubsection{Results}
For the three representative network topologies 
(i.e., the complete graph, star graph, and linear chain graph), 
the empirical distribution of the total cost for each algorithm compared
is shown in 
Fig.~\ref{fig:graph_topology}. 
Moreover,
Tables~[\ref{table:graph_15_loop},\ref{table:graph_15_star},\ref{table:graph_15_full}]
summarize the average total costs and competitive ratios of all considered algorithms for the three representative topologies, respectively.
Both the complete graph and the star graph have the same maximum node degree, 
whereas the graph density of the star graph is significantly lower. 
This explains why all the algorithms considered exhibit lower costs on the star graph, which is also consistent with our new theoretical analysis of the cost performance in Theorem~\ref{thm:avg_cost_blackbox}.
Furthermore, \ouralg can leverage the power of ML policy more efficiently with the reduced spatial cost uncertainties associated with fewer connections in the star graph. 

\begin{table}[!h]
\centering
\scriptsize
\begin{tabular}{c|c|c|c|c|c|c|c|c} 
\toprule
\multirow{2}{*}{} & \multirow{2}{*}{\textbf{Expert}} & \multirow{2}{*}{\textbf{ML}} & \multirow{2}{*}{\textbf{HitOnly}} & \multirow{2}{*}{\textbf{Greedy}} & \multicolumn{4}{c}{\textbf{\ouralg}}\\ 
\cline{6-9}
    & & & & &  $\lambda=0.2$ & $\lambda=0.5$ & $\lambda=1$ & $\lambda=2$ \\ 
\hline
AVG& 662.38 & 489.03 & 613.73 & 1054.87 & 559.59 & 519.75 & 500.63 & 491.86 \\ 
\hline
CR & 2.065 & 4.653 & 7.380 & 4.227 & 1.948 & 2.234 & 2.422 & 3.068 \\ 
\cmidrule[\heavyrulewidth]{1-9}
\multirow{2}{*}{} & \multicolumn{4}{c|}{\textbf{\ouralglinear}} & \multicolumn{4}{c}{\textbf{\ouralgft}}\\ 
\cline{2-9}
    & $\gamma=0.1$ & $\gamma=0.3$ & $\gamma=0.5$ & $\gamma=0.9$ & 
 $\lambda=0.2$ & $\lambda=0.5$ & $\lambda=1$ & $\lambda=2$  \\ 
\hline
AVG & 618.72 & 548.95 & 502.58 & 480.04 & 555.77 & 509.92 & 488.91 & \textbf{481.59}  \\ 
\hline
CR & 1.972 & 2.021 & 2.382 & 4.042 & \textbf{1.912} & 2.032 & 2.186 & 2.639  \\
\bottomrule
\end{tabular}
\caption{AVG and CR comparison between different algorithms for a 15-node chain graph. The average cost of OPT in the testing dataset is 405.37.}\label{table:graph_15_loop}
\end{table}

\begin{table}[!h]
\centering
\scriptsize
\begin{tabular}{c|c|c|c|c|c|c|c|c} 
\toprule
\multirow{2}{*}{} & \multirow{2}{*}{\textbf{Expert}} & \multirow{2}{*}{\textbf{ML}} & \multirow{2}{*}{\textbf{HitOnly}} & \multirow{2}{*}{\textbf{Greedy}} & \multicolumn{4}{c}{\textbf{\ouralg}}\\ 
\cline{6-9}
    & & & & &  $\lambda=0.2$ & $\lambda=0.5$ & $\lambda=1$ & $\lambda=2$ \\ 
\hline
AVG& 656.98 & 499.12 & 613.73 & 1030.78 & 610.87 & 606.46 & 583.98 & 539.01\\ 
\hline
CR & 1.891 & 4.679 & 7.295 & 3.748 & 1.916 & 2.262 & 2.477 & 3.109 \\ 
\cmidrule[\heavyrulewidth]{1-9}
\multirow{2}{*}{} & \multicolumn{4}{c|}{\textbf{\ouralglinear}} & \multicolumn{4}{c}{\textbf{\ouralgft}}\\ 
\cline{2-9}
    & $\gamma=0.1$ & $\gamma=0.3$ & $\gamma=0.5$ & $\gamma=0.9$ & 
 $\lambda=0.2$ & $\lambda=0.5$ & $\lambda=1$ & $\lambda=2$  \\ 
\hline
AVG & 614.60 & 547.57 & 504.18 & 497.77 & 602.06 & 571.04 & 528.68 & \textbf{497.29}  \\ 
\hline
CR & \textbf{1.802} & 1.889 & 2.309 & 4.614 & 1.882 & 2.039 & 2.225 & 2.666  \\
\bottomrule
\end{tabular}
\caption{AVG and CR comparison between different algorithms for a 15-node star graph, shown as Fig.~\ref{fig:star_topology} . The average cost of OPT in the testing dataset is 411.06.}\label{table:graph_15_star}
\end{table}

\begin{table}[!h]
\centering
\scriptsize
\begin{tabular}{c|c|c|c|c|c|c|c|c} 
\toprule
\multirow{2}{*}{} & \multirow{2}{*}{\textbf{Expert}} & \multirow{2}{*}{\textbf{ML}} & \multirow{2}{*}{\textbf{HitOnly}} & \multirow{2}{*}{\textbf{Greedy}} & \multicolumn{4}{c}{\textbf{\ouralg}}\\ 
\cline{6-9}
    & & & & &  $\lambda=0.2$ & $\lambda=0.5$ & $\lambda=1$ & $\lambda=2$ \\ 
\hline
AVG& 991.99 & 630.41 & 613.73 & 2401.08 & 839.00 & 767.09 & 727.77 & 687.17\\ 
\hline
CR & 3.245 & 3.619 & 4.048 & 11.577 & 2.614 & 3.026 & 3.469 & 3.595 \\ 
\cmidrule[\heavyrulewidth]{1-9}
\multirow{2}{*}{} & \multicolumn{4}{c|}{\textbf{\ouralglinear}} & \multicolumn{4}{c}{\textbf{\ouralgft}}\\ 
\cline{2-9}
    & $\gamma=0.1$ & $\gamma=0.3$ & $\gamma=0.5$ & $\gamma=0.9$ & 
 $\lambda=0.2$ & $\lambda=0.5$ & $\lambda=1$ & $\lambda=2$  \\ 
\hline
AVG & 914.90 & 787.99 & 697.48 & 625.63 & 650.70 & 637.81 & 632.44 & \textbf{622.09} \\ 
\hline
CR & 3.042 & 2.718 & 2.670 & 3.377 & \textbf{2.047} & 2.090 & 2.478 & 2.901  \\
\bottomrule
\end{tabular}
\caption{AVG and CR comparison between different algorithms for a 15-node complete graph. The average cost of OPT in the testing dataset is 524.79.}\label{table:graph_15_full}
\end{table}

Interestingly, the star graph and linear chain graph share the same number of edges (or graph density), while the star graph concentrates node degrees, leading to distinct cost behaviors.
This is evident when comparing the average cost distributions in Fig.~\ref{fig:graph_topology_cost_star} and Fig.~\ref{fig:graph_topology_cost_loop}.
The uniform node degree distribution in the chain graph allows \ouralg to leverage the power of learning-based policy more efficiently and further reduce the total cost in the linear chain graph. 
This empirical observation aligns with Theorem~\ref{thm:avg_cost_blackbox}, 
which suggests 
a more distributed node degree  reduces 
a lower term $\omega_v(\lambda, \tilde{\pi}, \pi^\dagger)$ due to perturbations to the ML policy.

\begin{figure*}[t]
\subfigure[Total cost with different $\lambda$]{
\includegraphics[width=0.45\textwidth]{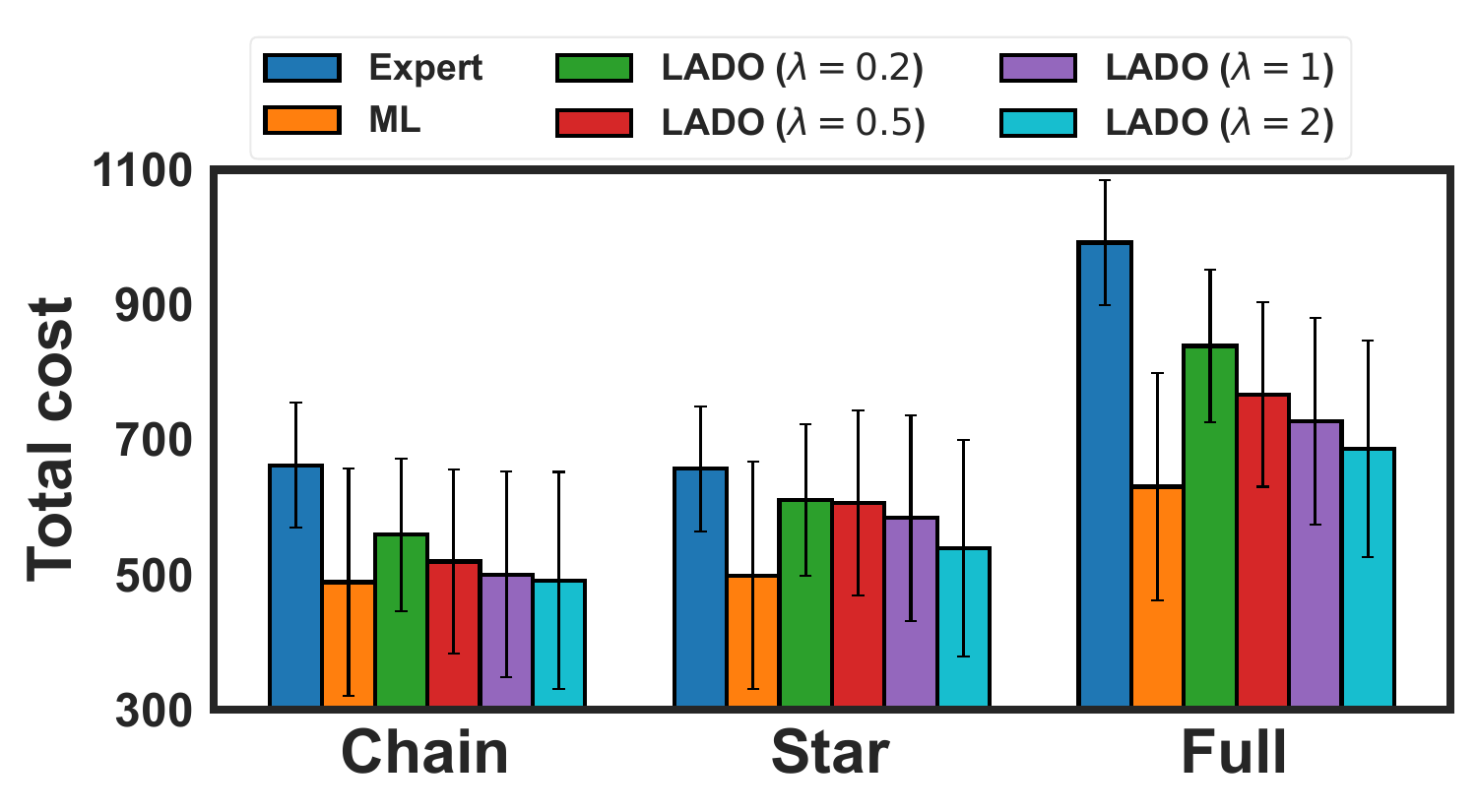}
\label{fig:overall_error_bar}
}
\subfigure[Competitive ratio with different $\lambda$]{
    \raisebox{-1.5mm}{
        \includegraphics[width=0.46\textwidth]{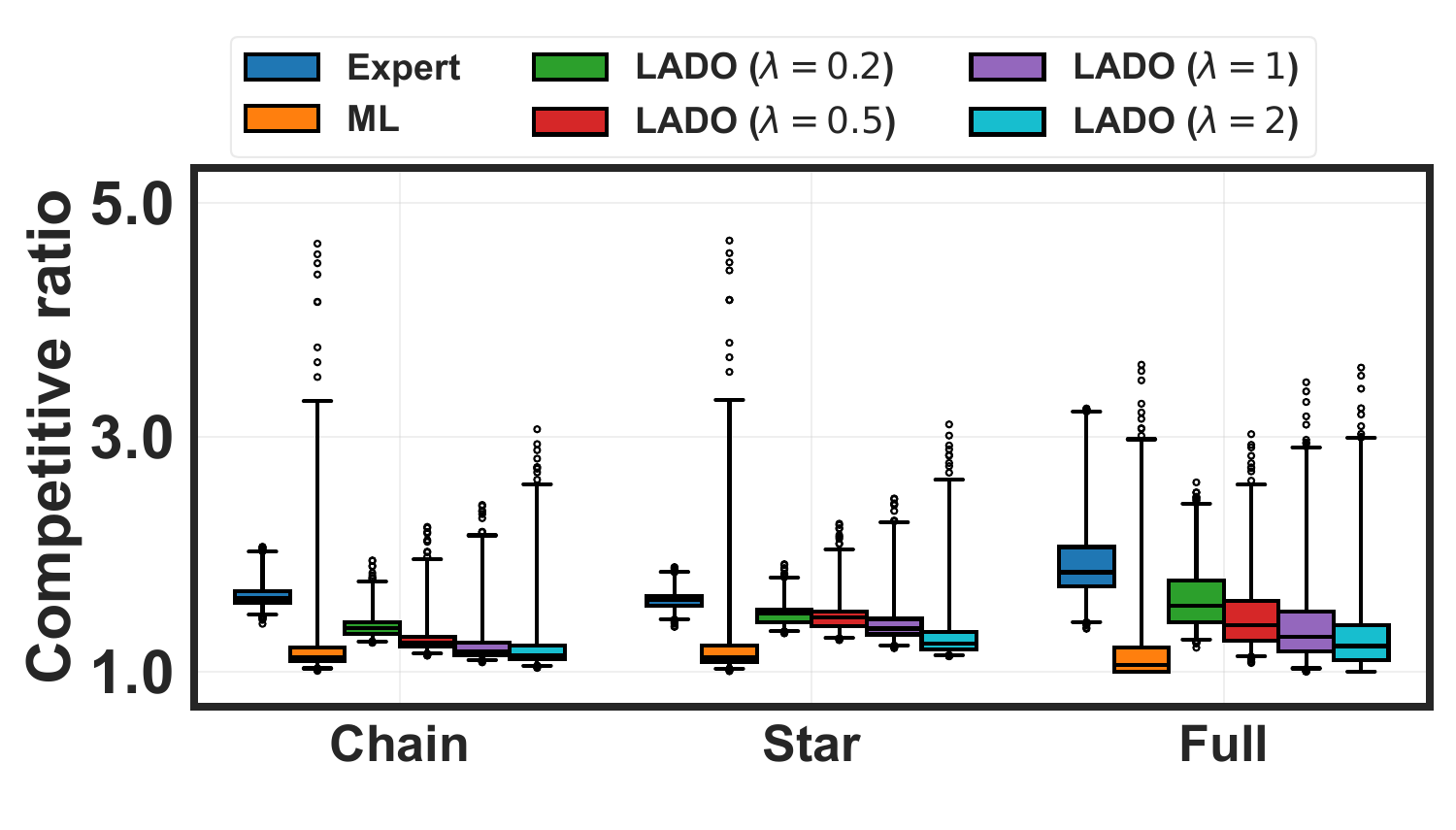}
        \label{fig:overall_CR}
    }
}
\caption{The comparison of total cost and competitive ratio distribution between \ouralg and other baseline algorithms for a 15-node network.}
\end{figure*}

\textbf{Impact of competitiveness requirement $\lambda$}: Next, we empirically evaluate the robustness-consistency trade-off of \ouralg under different graph topologies. For all the three graphs, the empirical average cost of learning-based policy (ML) outperforms the state-of-the-art expert policy, which highlights the importance of reducing the deviation of \ouralg from the ML policy to improve the average cost performance. 
As suggested by Theorem~\ref{thm:avg_cost_blackbox}, 
reducing the spatial cost uncertainty (fewer node connections) or 
relaxing the competitiveness requirement (larger $\lambda$)
leads to a smaller deviation of \ouralg from the ML policy,
which in turn reduces the perturbations to the ML policy because of projection.
Fig.~\ref{fig:overall_error_bar} 
clearly illustrates this trend 
by comparing the average cost 
of \ouralg under different settings.
However, such performance improvement comes at an expense in terms
of the competitiveness guarantees. 
As illustrated in Fig.~\ref{fig:overall_CR}, 
a larger $\lambda$ weakens the competitiveness guarantees of \ouralg,
reducing its worst-case protection and potentially leading to a higher competitive ratio.

\textbf{Impact of network topologies}:
In this experiment, we first evaluate \ouralg on graphs with the same number of nodes but varying random graph topologies. Specifically, we consider 15 battery units with varying characteristics and randomly selected spatial connections between these nodes. The minimum number of edges is set the same as a star graph instead of zero, since otherwise the nodes' decisions would become uncorrelated without any spatial connections. Starting from the star graph, we gradually add random edges between nodes to increase the graph density until the graph is fully connected. 
Additionally, we experiment with different competitiveness requirements under these graph topologies. The total cost and regret of \ouralg compared to the ML policy are shown in Fig.~\ref{fig:heatmap_main}.

As the graph density increases with more spatial connections between nodes, the total cost of \ouralg rises monotonically, showing a direct correlation between increased node connectivity and greater spatial cost uncertainties.
To focus on the impacts solely due to the projection process, we compare the regret of \ouralg against the ML policy, where both algorithms are evaluated under the same graph topologies. 
As shown in Fig.~\ref{fig:heatmap_response_regret_main}, the increased spatial cost uncertainty associated with denser graphs increases the difficulty for \ouralg to follow the ML policy, leading to a larger regret or cost increase compared to the ML policy. Moreover, as the competitiveness requirement becomes more stringent and \ouralg
needs to stay closer to the expert, it is more difficult for \ouralg to closely follow the ML policy.

\begin{figure*}[htp]
\subfigure[Total cost of \ouralg]{\includegraphics[width=0.47\textwidth]{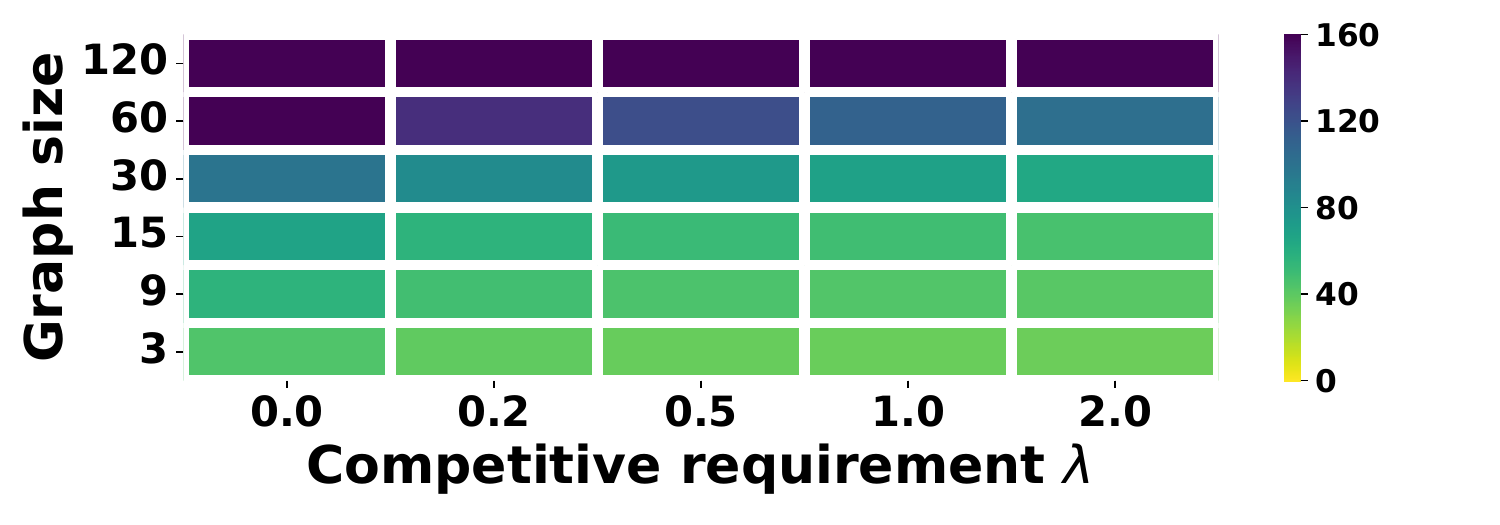}}
\subfigure[Regret of \ouralg compared to ML]{\includegraphics[width=0.47\textwidth]{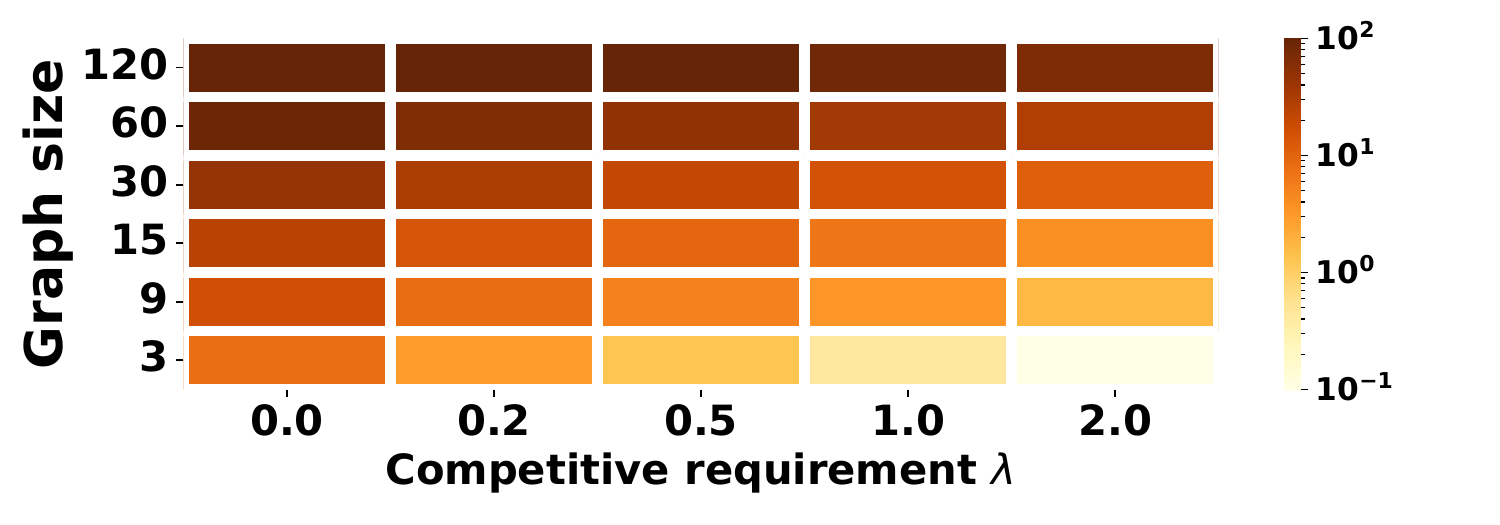}}
\caption{Impact of graph sizes and competitiveness requirement $\lambda$ on the overall cost of \ouralg, along with the additional cost (regret) incurred by the projection process compared to the ML policy. The overall cost and regret are normalized by the number of nodes for a consistent comparison across different graph sizes.}\label{fig:larger_graph_heatmap}
\end{figure*}

\textbf{Impact of graph sizes}: Next, we conduct a comprehensive comparison between the overall cost 
and regret of \ouralg (compared to the ML policy)
over a wide range of graph sizes, ranging from 3 to 120 nodes. For consistency, all the graphs are fully connected. We normalize the overall costs and regrets by the number of nodes. As illustrated in Fig.~\ref{fig:larger_graph_heatmap}, this normalization enables a meaningful comparison between algorithms over different graph sizes. 
Similar to our previous findings, the total cost and regret of \ouralg compared to the ML policy decrease as the competitiveness requirement relaxes with greater $\lambda$, even for the largest graph. In fully-connected graphs, a larger graph implies a higher degree of node connectivity.
As indicated by the term $\omega_v(\lambda, \tilde{\pi}, \pi^\dagger)$ in Theorem~\ref{thm:avg_cost_blackbox}, a larger node degree makes
it more difficult for \ouralg to follow the ML policy.
Consequently, given a fixed competitiveness requirement 
$\lambda$, 
the larger graph sizes, the greater node degrees,
and the higher regret of \ouralg compared to the ML policy
as suggested by Theorem~\ref{thm:avg_cost_blackbox}.

\begin{figure*}[htp]
\subfigure[Complete graph]{\includegraphics[width=0.8\textwidth]{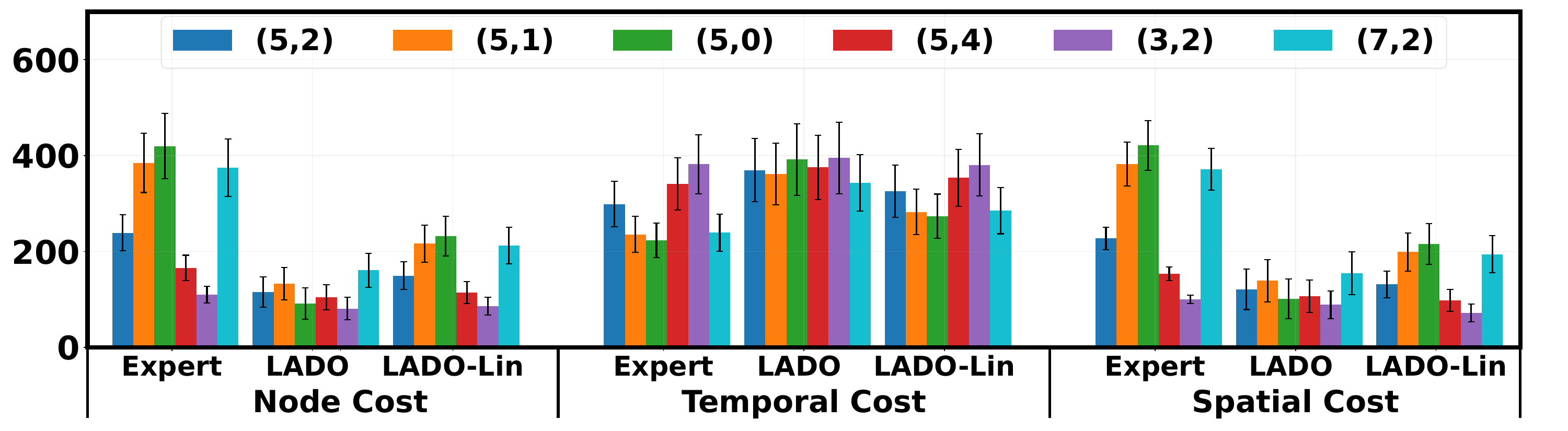}}\\
\vspace{-0.2cm}
\subfigure[Star graph]{\includegraphics[width=0.8\textwidth]{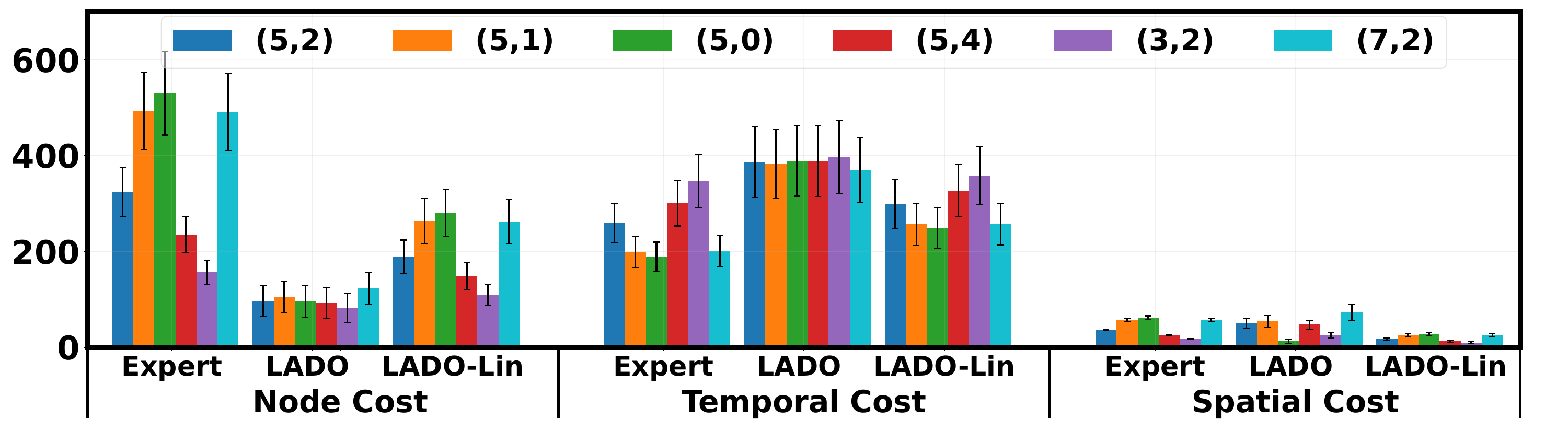}}\\
\vspace{-0.2cm}
\subfigure[Linear chain graph]{\includegraphics[width=0.8\textwidth]{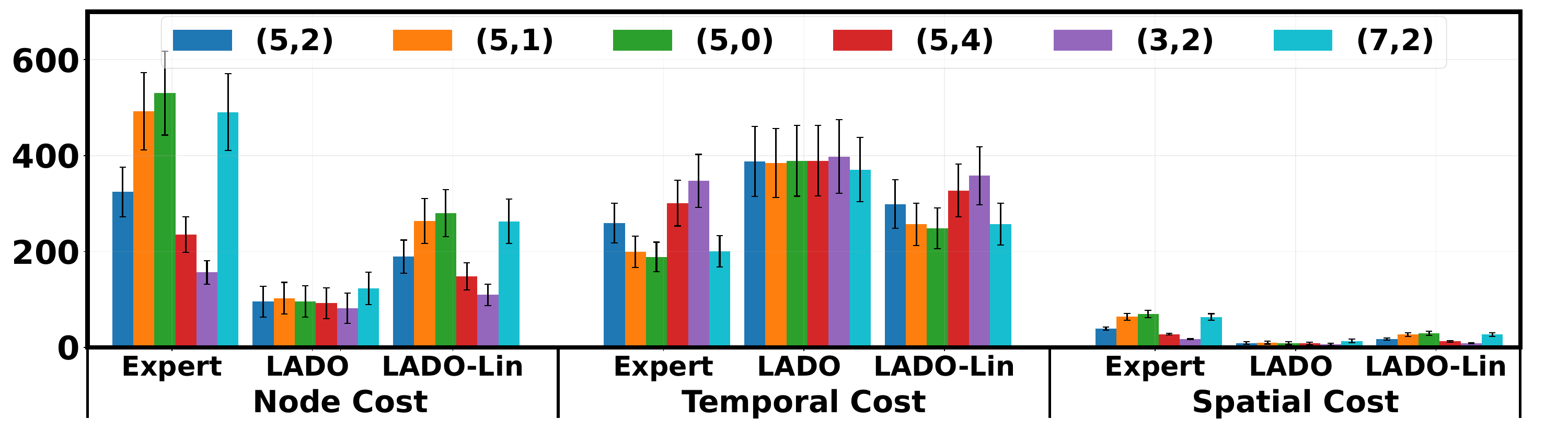}}\\
\caption{Comparison of node, temporal, and spatial costs for \ouralg with varying weights $(b, c)$ in the decentralized battery management formulation (shown as Eqn.~\eqref{eqn:total_cost_battery_original}). By default, 
we consider a 15-node network and set
the competitiveness requirement $\lambda$  as 1. To enhance visual clarity, the temporal costs are scaled up by 10 to align with the scale of node and spatial costs.  Results for \ouralglin and Expert are also included for reference. }\label{fig:different_weights}
\end{figure*}
\textbf{\ouralg with different weights for temporal and spatial costs}: As shown in Eqn.~\eqref{eqn:total_cost_battery_original}, the total cost is parameterized by the weights for the temporal and spatial costs, denoted as $(b,c)$ respectively. 
By normalizing the node cost weight to 1, the magnitudes of $b$ and $c$ directly reflect the relative importance of temporal and spatial decision smoothness versus reducing the node cost for achieving the desired state of charge at each battery node. It is crucial to note that these weights represent relative preferences rather than absolute cost scales.
For instance, even if we set both $b$ and $c$ as 1, the temporal and spatial costs are necessarily equal to the node cost.

We compare the performance of \ouralg, \ouralglin and Expert under a variety of weight combinations, as shown in Fig.~\ref{fig:different_weights}. 
In this experiment, the ML algorithm is not fine-tuned based on the weights
of $b$ and $c$ for the temporal and spatial costs, and the other baseline algorithms are not affected by these weights. 
By keeping the weight of switching cost constant, a greater $c$ prioritizes the spatial cost, thus leading to reduced spatial costs for all the algorithms. Conversely, a smaller $c$ gives more emphasis to the temporal cost. 
Additionally, \ouralg exhibits less sensitivity to weight variations $(b, c)$ compared to \ouralglin and Expert, which is more evident in the complete graph. This robustness stems from \ouralg's design, where the expert advice constructs a robust action set. Consequently, \ouralg's spatial cost is less influenced by the Expert policy changes than the static linear combination in \ouralglin.

\begin{figure*}[htp]
\subfigure[Complete graph]{\includegraphics[width=0.31\textwidth]{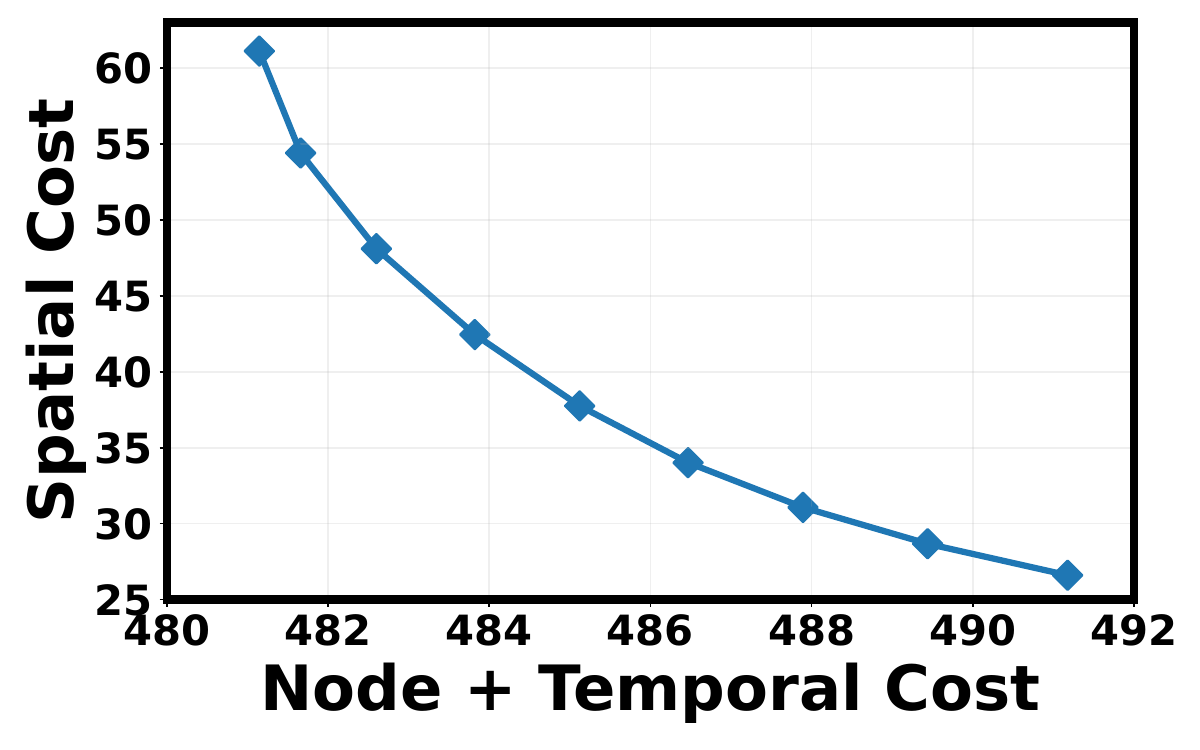}
\label{fig:tradeoff_curve_complete}
}
\subfigure[Star graph]{\includegraphics[width=0.31\textwidth]{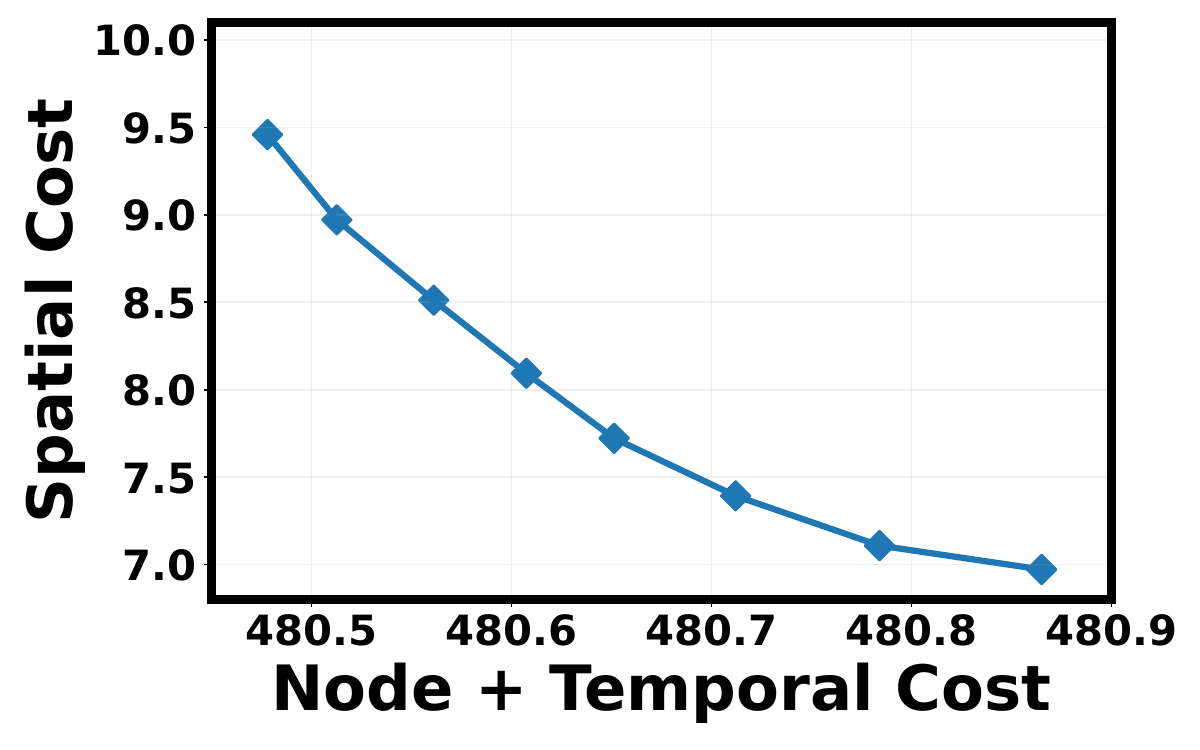}}
\subfigure[Linear chain graph]{\includegraphics[width=0.31\textwidth]{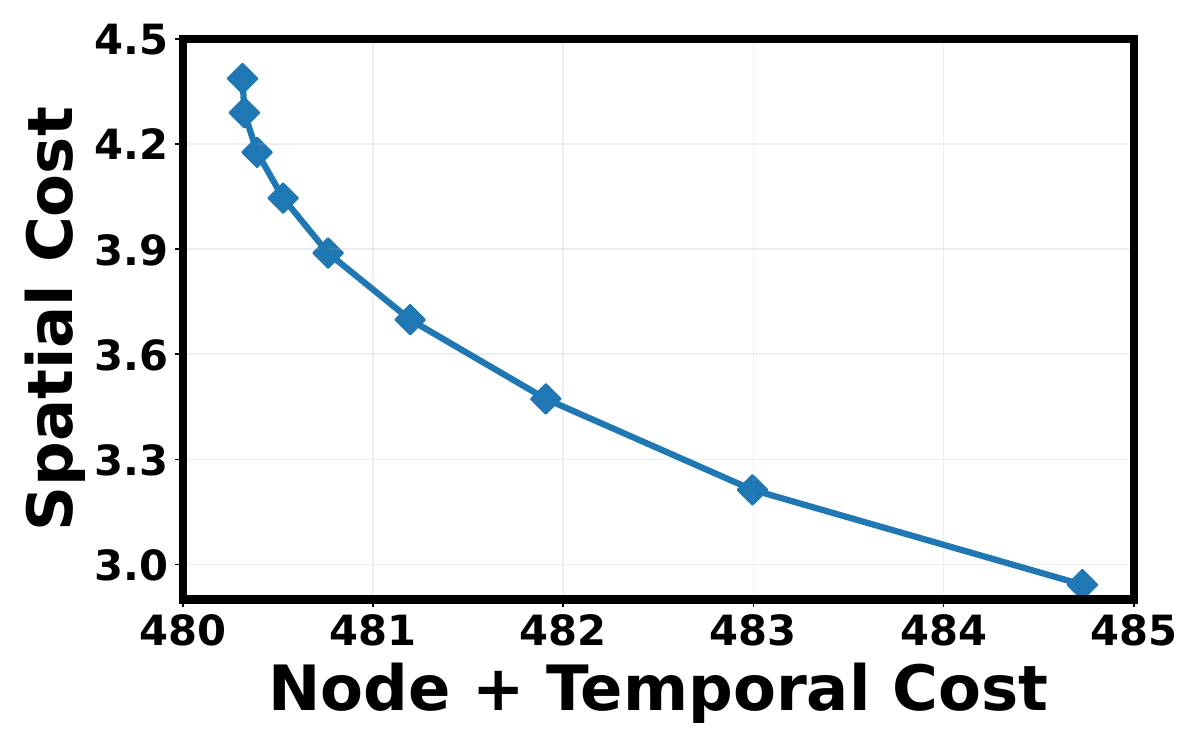}}\\
\caption{{The tradeoff curve between spatial cost and the sum of node and temporal cost of \ouralg by adjusting the coefficient $c$ in spatial cost, which penalizes the SoC difference between connected battery units. The parameter $b$ for the temporal cost is set as 1 by default.} }\label{fig:tradeoff_curve}
\end{figure*}

\revise{In practice, the spatial cost parameter is adjusted to strike a balance between two competing objectives: the local performance of each node (measured by node and temporal costs) and the spatial consistency among connected nodes. For the graphs with heterogeneous nodes, these objectives may conflict. Prioritizing spatial consistency can make it more difficult to achieve optimal local performance for individual nodes compared to the scenario where each node operates independently. 
As shown in Fig.~\ref{fig:tradeoff_curve}, by increasing the spatial cost parameter $c$, \ouralg enhances SoC consistency among connected battery units. To directly compare the actual SoC difference between battery units across different scenarios, spatial costs are evaluated using a constant parameter of $c=1$.
Consequently, for all the three graph topologies, we observe that the local costs (e.g. node and temporal costs) of individual nodes increase as the spatial difference decreases. Moreover, graphs with more spatial connections, such as the complete graph in Fig \ref{fig:tradeoff_curve_complete}, exhibit a greater sensitivity to the spatial cost parameter $c$. This is because introducing additional spatial considerations between nodes amplifies the impact of $c$ on local costs.
}

\section{Proofs of Results in Section~\ref{sec:lado-linear}}\label{appendix:proof_linear}
We begin with a technical lemma.

\begin{lemma}\label{lemma:spatial_reserve}
If the spatial cost is non-negative, convex and $\ell_S$-smooth w,r,t the vector $(x_v, x_u)$, then for any $\lambda>0$, it holds that
\begin{align}
     s_t^{(v, u)}(x_t^v, x_t^u) - (1+\lambda) s_t^{(v, u)}(x_t^{v,\dagger}, x_t^{u,\dagger}) \leq \frac{\ell_S}{2} (1+ \frac{1}{\lambda}) \left( \|x_t^v - x_t^{v,\dagger} \|^2 +  \| x_t^u - x_t^{u,\dagger} \|^2 \right).
\end{align}
\end{lemma} 
\begin{proof}
By the definition of smoothness, we have
\begin{equation}\label{eqn:lemma_1_eq1}
\begin{aligned}
    & s_t^{(v, u)}(x_t^v, x_t^u) \\
    \leq & s_t^{(v, u)}(x_t^{v,\dagger}, x_t^{u,\dagger})  + <\nabla s_t^{(v, u)}(x_t^{v,\dagger}, x_t^{u,\dagger}), (x_t^v - x_t^{v,\dagger}, x_t^u - x_t^{u,\dagger}) > + \frac{\ell_S}{2}\|(x_t^v, x_t^u) - (x_t^{v,\dagger}, x_t^{u,\dagger}) \|^2 \\
    \leq &  s_t^{(v, u)}(x_t^{v,\dagger}, x_t^{u,\dagger})  + \| \nabla s_t^{(v, u)}(x_t^{v,\dagger}, x_t^{u,\dagger})\|\cdot \|(x_t^v - x_t^{v,\dagger}, x_t^u - x_t^{u,\dagger})\| + \frac{\ell_S}{2}\|(x_t^v, x_t^u) - (x_t^{v,\dagger}, x_t^{u,\dagger}) \|^2\\
    \leq & s_t^{(v, u)}(x_t^{v,\dagger}, x_t^{u,\dagger})  + \frac{\lambda}{2\ell_S}\| \nabla s_t^{(v, u)}(x_t^{v,\dagger}, x_t^{u,\dagger})\|^2 +  (1+ \frac{1}{\lambda})\frac{\ell_S}{2}\|(x_t^v, x_t^u) - (x_t^{v,\dagger}, x_t^{u,\dagger}) \|^2 
\end{aligned}
\end{equation}

The second inequality comes from the property of inner product. The third inequality is based on AM-QM inequality. Besides, if $(\hat{x}_t^v, \hat{x}_t^u)$ is a minimizer of the spatial cost, by Lemma 2.9 in \cite{convex_course}, we have
\begin{equation}\label{eqn:lemma_1_eq2}
    s_t^{(v, u)}(x_t^{v,\dagger}, x_t^{u,\dagger}) \geq s_t^{(v, u)}(\hat{x}_t^v, \hat{x}_t^u) + 0 + \frac{1}{2\ell_S}\| \nabla s_t^{(v, u)}(x_t^{v,\dagger}, x_t^{u,\dagger})\|^2 \geq \frac{1}{2\ell_S}\| \nabla s_t^{(v, u)}(x_t^{v,\dagger}, x_t^{u,\dagger})\|^2
\end{equation}
By substituting Eqn~\eqref{eqn:lemma_1_eq2} back to Eqn~\eqref{eqn:lemma_1_eq1}, we have
\begin{equation}
    s_t^{(v, u)}(x_t^v, x_t^u)   \leq  (1+\lambda)\cdot s_t^{(v, u)}(x_t^{v,\dagger}, x_t^{u,\dagger})  +  (1+ \frac{1}{\lambda})\frac{\ell_S}{2}(\|(x_t^v - x_t^{v,\dagger}\|^2 + \| x_t^u - x_t^{u,\dagger}\|^2).
\end{equation}
\end{proof}

\subsection{Proof of Theorem \ref{thm:linear_bound}}\label{appendix:proof_linear_theorem}
Since \ouralglinear selects action as $x_t^v=\gamma \tilde{x}_t^v+(1-\gamma)x_t^{v,\dagger}$ at each round, by convexity of the global cost, we have
\begin{equation}
cost(\ouralglinear, g_{1:T})\leq \gamma cost(\tilde{\pi}, g_{1:T}) + (1-\gamma) cost(\pi^{\dagger},g_{1:T}).
\end{equation}
By taking expectation on both sides, the cost of \ouralglinear is bounded by the first term in the min operator. Next, we prove the cost of \ouralglinear is also bounded by the second term in the min operator. 

First, we can write the norm of the difference between the actions of \ouralglinear and expert $\pi^{\dagger}$ as 
\begin{equation}
\|x_t^v-\tilde{x}_t^{v}\|=(1-\gamma)\|x_t^{v,\dagger}-\tilde{x}_t^v\|.
\end{equation}

Based on Lemma \ref{lemma:spatial_reserve}, for any $\lambda_2 > 0$, for any $v\in\mathcal{E}$, we have 
\begin{equation}\label{eqn:hitting_temperal_2}
\begin{aligned}
    &\left(\sum_{\tau=1}^{T} f_\tau^{v}(x_\tau^{v}) +
    \sum_{\tau=1}^{T}  c_\tau ^{v}(x_\tau^{v}, x_{\tau-1}^{v})\right) -     (1+\lambda_2)\left(\sum_{\tau=1}^{T} f_\tau^{v}(\tilde{x}_\tau^{v}) +
    \sum_{\tau=1}^{T}  c_\tau ^{v}(\tilde{x}_\tau^{v}, \tilde{x}_{\tau-1}^{v})\right) \\
    \leq& (1+\frac{1}{\lambda_2}) \left(\frac{\ell_f}{2} \sum_{\tau=1}^T \|x_\tau^{v} - \tilde{x}_\tau^{v} \|^2  + \frac{\ell_T}{2} \sum_{\tau=1}^T \|x_\tau^{v} - \tilde{x}_\tau^{v} \|^2 +   \frac{\ell_T}{2} \sum_{\tau=0}^{T-1} \|x_\tau^{v} - \tilde{x}_\tau^{v} \|^2\right)\\
    \leq & (1+\frac{1}{\lambda_2}) \frac{\ell_f + 2\cdot \ell_T}{2} \sum_{\tau=1}^T \|x_\tau^{v} - \tilde{x}_\tau^{v} \|^2\leq (1-\gamma )(1+\frac{1}{\lambda_2}) \frac{\ell_f + 2\cdot \ell_T}{2} \sum_{\tau=1}^T \|x_t^{v,\dagger}-\tilde{x}_t^v\|^2
\end{aligned}
\end{equation}

By summing up the spacial costs over time, we have 
\begin{equation}\label{eqn:spacial_cost_sum_2}
\begin{aligned}
    &\sum_{(v, u) \in \mathcal{E}} \sum_{\tau=1}^{T}{s_\tau^{(v, u)}(x_\tau^{v}, x_\tau^u)}  - (1+\lambda_2)\sum_{(v, u) \in \mathcal{E}}\sum_{\tau=1}^{T} {s_\tau^{(v, u)}(\tilde{x}_\tau^{v}, \tilde{x}_\tau^{u})} \\
    \leq & (1+\frac{1}{\lambda_2}) \frac{\ell_S}{2} \sum_{(v, u) \in \mathcal{E}}\sum_{\tau=1}^T \left(\|x_\tau^{v} - \tilde{x}_\tau^{v} \|^2+\|x_\tau^{v} - \tilde{x}_\tau^{v} \|^2\right)\\
    =& (1+\frac{1}{\lambda_2}) \sum_{v\in\mathcal{V}}\frac{D_v\cdot \ell_S}{2} \sum_{\tau=1}^T \|x_\tau^{v} - \tilde{x}_\tau^{v} \|^2\leq (1-\gamma)^2(1+\frac{1}{\lambda_2}) \sum_{v\in\mathcal{V}}\frac{D_v\cdot \ell_S}{2} \sum_{\tau=1}^T \|x_t^{v,\dagger}-\tilde{x}_t^v\|^2
\end{aligned}
\end{equation}

By adding Eqn~\eqref{eqn:spacial_cost_sum_2} to Eqn~\eqref{eqn:hitting_temperal_2}, we can bound the sum cost error as
\begin{equation}
\begin{aligned}
cost (\ouralglinear)  -  (1+ \lambda_2)cost (\tilde{\pi}) &\leq (1-\gamma )^2(1+\frac{1}{\lambda_2}) \sum_{v\in\mathcal{V}}\frac{\ell_f + 2\cdot \ell_T + D_{v}\cdot \ell_S}{2} \sum_{\tau=1}^T \|x_t^{v,\dagger}-\tilde{x}_t^v\|^2
    \end{aligned}
\end{equation}

By taking expectation on both sides, we have
\begin{equation}
\begin{aligned}
AVG (\ouralglinear)  -  (1+ \lambda_2)AVG (\tilde{\pi}) &\leq (1-\gamma )^2(1+\frac{1}{\lambda_2}) \mathbb{E}_{g_{1:T}}\left[\sum_{v\in\mathcal{V}}\frac{\ell_f + 2\cdot \ell_T + D_{v}\cdot \ell_S}{2} \sum_{\tau=1}^T \|x_t^{v,\dagger}-\tilde{x}_t^v\|^2\right]
    \end{aligned}
\end{equation}

 By optimally setting $\lambda_2 = \sqrt{(1-\gamma)^2\mathbb{E}_{g_{1:T}}\left[\sum_{v\in\mathcal{V}}\frac{\ell_f + 2\cdot \ell_T + D_{v}\cdot \ell_S}{2} \sum_{\tau=1}^T \|x_t^{v,\dagger}-\tilde{x}_t^v\|^2\right]\frac{1}{AVG (\tilde{\pi})}}$, we can bound the sum cost as 
 \begin{equation}
\begin{aligned}
cost (\ouralglinear) \leq  \left(\sqrt{AVG(\tilde{\pi})}+ (1-\gamma )\sqrt{\mathbb{E}_{g_{1:T}} \Bigl[ \sum_{t=1}^T \sum_{v\in \cV} \frac{\ell_f + 2\ell_T  + D_v\ell_S}{2} \| \tilde{x}_t^v - {x}^{v, \dagger}_t \|^2} \Bigr]\right)^2
    \end{aligned}
\end{equation}

\subsection{Proof of  Proposition~\ref{coro:robustness_violation}}\label{sec:proof_proposition3.2}
If \ouralglinear satisfies the $\lambda-$competitiveness constraint, we have
\begin{equation}\label{eqn:linear_cost_bound_proof2}
cost(\ouralglinear)\leq (1+\lambda) cost(\pi^{\dagger})\leq (1+\lambda) \rho_{\pi^{\dagger}} cost(\pi^*)
\end{equation}
Since the cost function $cost$ is $\beta-$strongly convex,
the gradient of $cost$ at $x^*$ is $\nabla cost(x^*)=0$ and it holds that
\begin{equation}\label{eqn:linear_cost_bound_proof1}
\begin{split}
cost(\ouralglinear)&\geq cost(\pi^*)+\frac{\beta}{2}\|\gamma \tilde{x}_t+(1-\gamma)x_t^{\dagger}-x^*\|^2\\
&= cost(\pi^*)+\frac{\beta}{2}\|\gamma (\tilde{x}_t-x^*)+(1-\gamma)(x_t^{\dagger}-x^*)\|^2
\end{split}
\end{equation}
Substituting \eqref{eqn:linear_cost_bound_proof1} to \eqref{eqn:linear_cost_bound_proof2}, we have
\begin{equation}
\frac{\beta}{2}\|\gamma (\tilde{x}_t-x^*)+(1-\gamma)(x_t^{\dagger}-x^*)\|^2 \leq \left((1+\lambda) \rho_{\pi^{\dagger}} -1\right)cost(\pi^*).
\end{equation}
By taking the squared root for both sides and applying triangle inequality for the left hand side, it holds that
\begin{equation}\label{eqn:bound_expert_action_2}
\|\gamma (\tilde{x}_t-x^*)\|-\|(1-\gamma)(x_t^{\dagger}-x^*)\| \leq \sqrt{\frac{2}{\beta}\left((1+\lambda) \rho_{\pi^{\dagger}} -1\right)cost(cost(\pi^*)}.
\end{equation}
Since the cost function is $\beta-$strongly convex, we have
\begin{equation}\label{eqn:bound_expert_action_1}
cost(\pi^*)+\frac{\beta}{2}\|x^{\dagger}-x^*\|^2\leq cost(x^{\dagger})\leq \rho_{\pi^{\dagger}}cost(\pi^*).
\end{equation}
Substituting \eqref{eqn:bound_expert_action_1} into \eqref{eqn:bound_expert_action_2}, we have
\begin{equation}
\begin{split}
\|\tilde{x}_t-x^*\|\leq \left(\frac{1-\gamma}{\gamma}\sqrt{(\rho_{\pi^{\dagger}}-1)\frac{2}{\beta}} + \frac{1}{\gamma}\sqrt{\frac{2}{\beta}\left((1+\lambda) \rho_{\pi^{\dagger}} -1\right)}\right)\sqrt{cost(\pi^*)}.
\end{split}
\end{equation}
The proposition is proved by moving items in the above inequality.

\section{Proof of Robustness in Theorem~\ref{thm:robustness}}\label{sec:robustness_proof}
To prove  Theorem \ref{thm:robustness}, the key point is to guarantee the robust action set \eqref{eqn:robust_action_set} is non-empty. We will prove this through induction. For $t=1$, it is obvious that $x_1^{v} =  x_1^{v, \dagger}$ satisfies the constraint. We assume that the robustness constraint is satisfied up to time step $t-1$, which is
\begin{equation}\label{eqn:robust_proof_eqn1}
\begin{aligned}
    &\sum_{\tau=1}^{t-1} f_\tau^{v}(x_\tau^{v}) + \sum_{\tau=1}^{t-2}\sum_{(v, u) \in \mathcal{E}} \kappa_\tau^{(v,u)} \cdot {s_\tau^{(v, u)}(x_\tau^{v}, x_\tau^u)} +  \frac{\ell_T + \ell_S\cdot D_{v}}{2}(1+\frac{1}{\lambda_0})\cdot \|x_{t-1}^{v} - x_{t-1}^{v, \dagger} \|^2  \\
     +&\sum_{\tau=1}^{t-1}  c_\tau ^{v}(x_\tau^{v}, x_{\tau-1}^{v}) \leq  (1+\lambda) \Bigl(\sum_{\tau=1}^{t-1} f_\tau^{v}(x_\tau^{v, \dagger})  + \sum_{\tau=1}^{t-1} c_\tau ^{v}(x_\tau^{v, \dagger}, x_{\tau-1}^{v, \dagger}) +      \sum_{\tau=1}^{t-2}\sum_{(v, u) \in \mathcal{E}} \kappa_\tau^{(v,u)} \cdot {s_\tau^{(v, u)}(x_\tau^{v, \dagger}, x_\tau^{u, \dagger})}\Bigr) 
\end{aligned}
\end{equation}

Based on Lemma \ref{lemma:spatial_reserve} and $\kappa_{t-1}^{(v,u)} = \frac{\|x_{t-1}^v - x_{t-1}^{v,\dagger} \|^2}{\|x_{t-1}^v - x_{t-1}^{v,\dagger} \|^2 + \| x_{t-1}^u - x_{t-1}^{u,\dagger} \|^2}$, we have 
\begin{equation}\label{eqn:spatial_ineq}
    \kappa_{t-1}^{(v, u)} \cdot \left({s_{t-1}^{(v, u)}(x_{t-1}^{v}, x_{t-1}^u)} - (1+\lambda)s_{t-1}^{(v, u)}(x_{t-1}^{v,\dagger}, x_{t-1}^{u,\dagger}) \right) \leq \frac{\ell_S}{2}(1+\frac{1}{\lambda})  \|x_{t-1}^{v} - x_{t-1}^{v, \dagger} \|^2
\end{equation}
For time step $t$, if we choose  $x_t^{v} =  x_t^{v, \dagger}$, by
the smoothness assumption, we have
\begin{equation}\label{eqn:time_ineq}
    \begin{aligned}
    c_t ^{v}(x_t^{v, \dagger}, x_{t-1}^{v}) - (1+\lambda) c_t ^{v}(x_t^{v, \dagger}, x_{t-1}^{v, \dagger}) &\leq \frac{\ell_T}{2}(1+\frac{1}{\lambda})  \|x_{t-1}^{v} - x_{t-1}^{v, \dagger} \|^2
    \end{aligned}
\end{equation}

Since the node cost is non-negative, by \eqref{eqn:spatial_ineq} and \eqref{eqn:time_ineq}, we have
\begin{equation}\label{eqn:robust_proof_eqn2}
\begin{split}
     &f_t^{v}(x_t^{v, \dagger}) + \sum_{(v, u) \in \mathcal{E}} \kappa_{t-1}^{(v,u)} \cdot {s_{t-1}^{(v, u)}(x_{t-1}^{v}, x_{t-1}^u)} + c_t ^{v}(x_t^{v,\dagger}, x_{t-1}^{v}) - \frac{\ell_T + \ell_S\cdot D_{v}}{2}(1+\frac{1}{\lambda_0})  \|x_{t-1}^{v} - x_{t-1}^{v, \dagger} \|^2\\
    &-  (1+\lambda) \Bigl( f_t^{v}(x_t^{v, \dagger})  + c_t ^{v}(x_t^{v, \dagger}, x_{t-1}^{v, \dagger}) +  \sum_{(v, u) \in \mathcal{E}} \kappa_{t-1}^{(v,u)} \cdot {s_{t-1}^{(v, u)}(x_{t-1}^{v, \dagger}, x_{t-1}^{u, \dagger})}\Bigr)\\
    \leq& \frac{\ell_T + \ell_S\cdot D_{v}}{2}(1+\frac{1}{\lambda})  \|x_{t-1}^{v} - x_{t-1}^{v, \dagger} \|^2- \frac{\ell_T + \ell_S\cdot D_{v}}{2}(1+\frac{1}{\lambda_0})  \|x_{t-1}^{v} - x_{t-1}^{v, \dagger} \|^2\\
    \leq & 0,
\end{split}
\end{equation}
where the last inequality holds by $\lambda\geq \lambda_0$.
By adding Eqn~\eqref{eqn:robust_proof_eqn2} back to Eqn~\eqref{eqn:robust_proof_eqn1} and moving items, we recover the robustness constraint for time step $t$ if $x_t^{v} =  x_t^{v, \dagger}$,
\begin{equation}
\begin{gathered}
    \sum_{\tau=1}^{t} f_\tau^{v}(x_\tau^{v}) + \sum_{\tau=1}^{t-1}\sum_{(v, u) \in \mathcal{E}} \kappa_\tau^{(v,u)} \cdot {s_\tau^{(v, u)}(x_\tau^{v}, x_\tau^u)} +
    \sum_{\tau=1}^{t}  c_\tau ^{v}(x_\tau^{v}, x_{\tau-1}^{v})+\frac{\ell_T + \ell_S\cdot D_{v}}{2}(1+\frac{1}{\lambda_0})  \|x_{t}^{v} - x_{t}^{v, \dagger} \|^2 \\
    \leq  (1+\lambda) \Bigl(\sum_{\tau=1}^{t} f_\tau^{v}(x_\tau^{v, \dagger})  + \sum_{\tau=1}^{t} c_\tau ^{v}(x_\tau^{v, \dagger}, x_{\tau-1}^{v, \dagger}) +      \sum_{\tau=1}^{t-1}\sum_{(v, u) \in \mathcal{E}} \kappa_\tau^{(v,u)} \cdot {s_\tau^{(v, u)}(x_\tau^{v, \dagger}, x_\tau^{u, \dagger})}\Bigr) 
\end{gathered}
\end{equation}
In other words, the expert's action $x_t^{v,\dagger}$ is always an action in the corresponding robust action set \eqref{eqn:robust_action_set}. Thus the robust action set is non-empty.

Since $\kappa_t^{(v,u)}+\kappa_t^{(u,v)}=1$ holds for $(v,u)\in\mathcal{E}$, if all the nodes select actions from the robust action set \eqref{eqn:robust_action_set} at each step, we can guarantee that $cost(\ouralg, g_{1:T}) \leq (1+\lambda)\cdot cost(\pi^{\dagger}, g_{1:T})$ is satisfied.

\section{Proof of Average Cost Bounds and Robust-Consistency of \ouralg}

\subsection{Proof of Theorem \ref{thm:avg_cost_blackbox}}\label{sec:cost_ratio_proof}

We begin by stating and proving a technical lemma and then move to the proof of Theorem \ref{thm:avg_cost_blackbox}.
\begin{lemma}\label{lemma:distance} 
We denote the actual actions from \ouralg as $x_{1:T}^{v} = (x_{1}^{v}, \cdots, x_{T}^{v})$, the squared distance between actual action and ML advice is bounded by 
\begin{equation*}
    \sum_{t=1}^T \|x_t^{v} - \tilde{x}_t^{v} \|^2 \leq  \sum_{t=1}^T \left[ \|\tilde{x}^v_t - x_t^{v,\dagger}\|^2 - {\frac{\lambda - \lambda_0}{1+\frac{1}{\lambda_0}} \cdot \frac{2}{\ell_f + 2\cdot\ell_T + \ell_S\cdot D_{v}} \cdot \text{cost}_{v,t}^\dagger } \right]^+ 
\end{equation*}
where  $\lambda$, $\lambda_0$, $D$, and $\rho$ are defined in Theorem \ref{thm:avg_cost_blackbox}.
\end{lemma}
\begin{proof}
To prove this lemma, we first construct a sufficient condition to satisfy the original constraint in Eqn~\eqref{eqn:local_constraint}. Then we prove a distance bound in this sufficient condition, where the bound still holds for the original problem.

At time step $t$, we know the constraint in time step $t-1$ is already satisfied, so we obtain the following sufficient condition of the satisfaction of \eqref{eqn:local_constraint} as
\begin{equation}
    \begin{gathered}\label{eqn:sufficient_1}
     f_t^{v}(x_t^{v})  + \sum_{(v, u) \in \mathcal{E}} \kappa_{t-1}^{(v,u)} \cdot {s_{t-1}^{(v, u)}(x_{t-1}^{v}, x_{t-1}^u)}  +  \frac{\ell_T + \ell_S\cdot D_{v}}{2}(1+\frac{1}{\lambda_0})  \left( \|x_t^{v} - x_t^{v, \dagger} \|^2  - \|x_{t-1}^{v} - x_{t-1}^{v, \dagger} \|^2 \right)\\
    + c_t ^{v}(x_t^{v}, x_{t-1}^{v}) \leq (1+\lambda) \Bigl( f_t^{v}(x_t^{v, \dagger})  + c_t ^{v}(x_t^{v, \dagger}, x_{t-1}^{v, \dagger}) +  \sum_{(v, u) \in \mathcal{E}} \kappa_{t-1}^{(v,u)} \cdot {s_{t-1}^{(v, u)}(x_{t-1}^{v, \dagger}, x_{t-1}^{u, \dagger})}\Bigr) 
    \end{gathered}
\end{equation}
With the convexity and smoothness assumptions, we have
\begin{equation}
    \begin{aligned}
    c_t^{v}(x_t^{v, \dagger}, x_{t-1}^{v}) - (1+\lambda_0) c_t ^{v}(x_t^{v, \dagger}, x_{t-1}^{v, \dagger}) &\leq \frac{\ell_T}{2}(1+\frac{1}{\lambda_0}) \left( \|x_{t}^{v} - x_{t}^{v, \dagger} \|^2 + \|x_{t-1}^{v} - x_{t-1}^{v, \dagger} \|^2   \right)\\
    f_t^{v}(x_t^{v}) - (1+\lambda_0) f_t^{v}(x_t^{v, \dagger}) &\leq \frac{\ell_f}{2}(1+\frac{1}{\lambda_0})  \|x_{t}^{v} - x_{t}^{v, \dagger} \|^2 
    \end{aligned}
\end{equation}
Then a sufficient condition that \eqref{eqn:sufficient_1} holds becomes
\begin{equation}\label{eqn:sufficient_2}
    \begin{gathered}
      \sum_{(v, u) \in \mathcal{E}} \kappa_{t-1}^{(v,u)} \cdot {s_{t-1}^{(v, u)}(x_{t-1}^{v}, x_{t-1}^u)}  + (1+\frac{1}{\lambda_0})\left( \frac{\ell_f + 2\cdot\ell_T + \ell_S\cdot D_{v}}{2}  \|x_t^{v} - x_t^{v, \dagger} \|^2  - \frac{ \ell_S\cdot D_{v}}{2} \|x_{t-1}^{v} - x_{t-1}^{v, \dagger} \|^2 \right)  \\
      \leq (\lambda - \lambda_0) \Bigl( f_t^{v}(x_t^{v, \dagger})  + c_t ^{v}(x_t^{v, \dagger}, x_{t-1}^{v, \dagger}) \Bigr) + (1+\lambda)  \sum_{(v, u) \in \mathcal{E}} \kappa_{t-1}^{(v,u)} \cdot {s_{t-1}^{(v, u)}(x_{t-1}^{v, \dagger}, x_{t-1}^{u, \dagger})}
    \end{gathered}
\end{equation}
From Eqn~\eqref{eqn:spatial_ineq}, we can further cancel out the spatial costs and get the sufficient condition of Eqn~\eqref{eqn:sufficient_2}, shown as below
\begin{equation}\label{eqn:sufficient_3}
    \begin{gathered}
      (1+\frac{1}{\lambda_0})\left( \frac{\ell_f + 2\cdot\ell_T + \ell_S\cdot D_{v}}{2}  \|x_t^{v} - x_t^{v, \dagger} \|^2 \right)  \\
      \leq (\lambda - \lambda_0) \Bigl( f_t^{v}(x_t^{v, \dagger})  + c_t ^{v}(x_t^{v, \dagger}, x_{t-1}^{v, \dagger}) + \sum_{(v, u) \in \mathcal{E}} \kappa_{t-1}^{(v,u)} \cdot {s_{t-1}^{(v, u)}(x_{t-1}^{v, \dagger}, x_{t-1}^{u, \dagger})} \Bigr)
    \end{gathered}
\end{equation}

For the expert policy $\pi^{\dagger}$ at time $t$, we define the sum of node cost and temporal cost as $\text{cost}_{v,t}^\dagger =  f_t^{v}(x_t^{v, \dagger})  + c_t ^{v}(x_t^{v, \dagger}, x_{t-1}^{v, \dagger}) $. Therefore,  a sufficient condition of Eqn~\eqref{eqn:sufficient_2} is 
\begin{equation}\label{eqn:sufficient_final}
    \begin{gathered}
        {\|x_t^{v} - x_t^{v, \dagger} \|^2}       \leq \frac{\lambda - \lambda_0}{1+\frac{1}{\lambda_0}} \cdot \frac{2}{\ell_f + 2\cdot\ell_T + \ell_S\cdot D_{v}} \cdot \text{cost}_{v,t}^\dagger
    \end{gathered}
\end{equation}
By summarizing the sufficient condition in Eqn (\ref{eqn:sufficient_1}-\ref{eqn:sufficient_final}), we  conclude that for any $x_t^{v}$ satisfying Eqn~\eqref{eqn:sufficient_final} must satisfy the original constraint Eqn~\eqref{eqn:local_constraint}. If the ML advice $\tilde{x}^v_t$ satisfies the inequality in Eqn~\eqref{eqn:sufficient_final}, we can completely follow ML advice without any modification for node $v$. 
Otherwise,  we construct a $\hat{x}_t^{v}$ achieves equity in Eqn~\eqref{eqn:sufficient_final} and satisfies $\|\hat{x}_t^{v} - \tilde{x}_t^{v}\| = \|\tilde{x}_t^{v} - x_t^{v,\dagger} \| - \|\hat{x}_t^{v} - \tilde{x}_t^{v}\|$. Therefore, the distance between the constructed action $\hat{x}_t^{v}$ and ML advice is given by
\begin{equation}
    \begin{aligned} 
        \|\hat{x}_t^{v} - \tilde{x}_t^{v} \| = & \left[ {\|\tilde{x}^v_t - x_t^{v,\dagger}\|} - \sqrt{\frac{\lambda - \lambda_0}{1+\frac{1}{\lambda_0}} \cdot \frac{2}{\ell_f + 2\cdot\ell_T + \ell_S\cdot D_{v}} \cdot \text{cost}_{v,t}^\dagger } \right]^+ 
    \end{aligned}
\end{equation}
Since $x_t^v$ is obtained by minimizing its distance to ML action $\tilde{x}_t^v$ under the original constraint in Eqn~\eqref{eqn:local_constraint}, it's obvious that $\|{x}_t^{v} - \tilde{x}_t^{v} \| \leq \|\hat{x}_t^{v} - \tilde{x}_t^{v} \|$. Besides, we have the following inequality
\begin{equation}\label{eqn:distance_single_lado}
    \begin{aligned} 
        & \left( \left[ {\|\tilde{x}^v_t - x_t^{v,\dagger}\|} - \sqrt{\frac{\lambda - \lambda_0}{1+\frac{1}{\lambda_0}} \cdot \frac{2}{\ell_f + 2\cdot\ell_T + \ell_S\cdot D_{v}} \cdot \text{cost}_{v,t}^\dagger } \right]^+ \right)^2\\
        \leq & \left[ {\|\tilde{x}^v_t - x_t^{v,\dagger}\|}^2 - {\frac{\lambda - \lambda_0}{1+\frac{1}{\lambda_0}} \cdot \frac{2}{\ell_f + 2\cdot\ell_T + \ell_S\cdot D_{v}} \cdot \text{cost}_{v,t}^\dagger } \right]^+
    \end{aligned}
\end{equation}
By summing up the inequalities over time, we complete the proof. 
\end{proof}

\begin{proof}[\textbf{Proof of Theorem \ref{thm:avg_cost_blackbox}}]
Based on Lemma \ref{lemma:spatial_reserve}, for any $\lambda_2 > 0$ we have 
\begin{equation}\label{eqn:cost_ratio_eq1}
\begin{aligned}
    &\left(\sum_{\tau=1}^{T} f_\tau^{v}(x_\tau^{v}) +
    \sum_{\tau=1}^{T}  c_\tau ^{v}(x_\tau^{v}, x_{\tau-1}^{v})\right) -     (1+\lambda_2)\left(\sum_{\tau=1}^{T} f_\tau^{v}(\tilde{x}_\tau^{v}) +
    \sum_{\tau=1}^{T}  c_\tau ^{v}(\tilde{x}_\tau^{v}, \tilde{x}_{\tau-1}^{v})\right) \\
    \leq& (1+\frac{1}{\lambda_2}) \left(\frac{\ell_f}{2} \sum_{\tau=1}^T \|x_\tau^{v} - \tilde{x}_\tau^{v} \|^2  + \frac{\ell_T}{2} \sum_{\tau=1}^T \|x_\tau^{v} - \tilde{x}_\tau^{v} \|^2 +   \frac{\ell_T}{2} \sum_{\tau=0}^{T-1} \|x_\tau^{v} - \tilde{x}_\tau^{v} \|^2\right)\\
    \leq & (1+\frac{1}{\lambda_2}) \frac{\ell_f + 2\cdot \ell_T}{2} \sum_{\tau=1}^T \|x_\tau^{v} - \tilde{x}_\tau^{v} \|^2
\end{aligned}
\end{equation}

By summing up the spacial costs over time, we have 
\begin{equation}\label{eqn:cost_ratio_eq2}
\begin{aligned}
    &\sum_{(v, u) \in \mathcal{E}} \sum_{\tau=1}^{T}{s_\tau^{(v, u)}(x_\tau^{v}, x_\tau^u)}  - (1+\lambda_2)\sum_{(v, u) \in \mathcal{E}}\sum_{\tau=1}^{T} {s_\tau^{(v, u)}(\tilde{x}_\tau^{v}, \tilde{x}_\tau^{u})} \\
    \leq & (1+\frac{1}{\lambda_2}) \frac{\ell_S}{2} \sum_{(v, u) \in \mathcal{E}}\sum_{\tau=1}^T \left(\|x_\tau^{v} - \tilde{x}_\tau^{v} \|^2+\|x_\tau^{v} - \tilde{x}_\tau^{v} \|^2\right)\\
    =& (1+\frac{1}{\lambda_2}) \sum_{v\in\mathcal{V}}\frac{D_v\cdot \ell_S}{2} \sum_{\tau=1}^T \|x_\tau^{v} - \tilde{x}_\tau^{v} \|^2
\end{aligned}
\end{equation}

By adding Eqn~\eqref{eqn:cost_ratio_eq1} and Eqn~\eqref{eqn:cost_ratio_eq2}, we can bound the cost error of node $v$ as
\begin{equation}\label{eqn:cost_ratio_eq3}
\begin{aligned}
   \text{cost} ({x}_{1:T})  -  (1+ \lambda_2)\text{cost}(\tilde{x}_{1:T}) &\leq (1+\frac{1}{\lambda_2})  \sum_{v\in\mathcal{V}} \frac{\ell_f + 2\cdot \ell_T + D_{v}\cdot \ell_S}{2} \sum_{\tau=1}^T \|x_\tau^{v} - \tilde{x}_\tau^{v} \|^2,
    \end{aligned}
\end{equation}
where $D_{\max}=\max_{v\in\mathcal{V}} D_v$ is the maximum degree of nodes.
By substituting Lemma \ref{lemma:distance} into Eqn~\eqref{eqn:cost_ratio_eq3}, we have 
\begin{equation}\label{eqn:consistency_proof_3}
    \begin{aligned}
        & \text{cost} ({x}_{1:T})  -  (1+ \lambda_2) \text{cost} (\tilde{x}_{1:T}) \leq  (1+\frac{1}{\lambda_2})   \sum_{v\in\mathcal{V}}   \sum_{t=1}^T \left[ \frac{\ell_f + 2\cdot\ell_T + \ell_S\cdot D_{v}}{2} \|\tilde{x}^v_t - x_t^{v,\dagger}\|^2 -  {\frac{\lambda - \lambda_0}{1+\frac{1}{\lambda_0}} \cdot  \text{cost}_{v,t}^\dagger } \right]^+
    \end{aligned}
\end{equation}
where the right-hand side captures the cost increase brought by the robustification process, which is minimized by setting $\lambda_0 = \sqrt{1+\lambda} - 1$.
By taking the expectation of Eqn~\eqref{eqn:consistency_proof_3} over the context distribution $\mathbb{P}_{g_{1:T}}$, we have 
\begin{equation}
    {AVG}(\ouralg(\tilde{\pi}))  -  (1+ \lambda_2) {AVG}(\tilde{\pi}) \leq  (1+\frac{1}{\lambda_2}) \sum_{v \in \mathcal{V} }\omega_v(\lambda, \tilde{\pi}, \pi^{\dagger})
\end{equation}
where $\omega_v(\lambda, \tilde{\pi}, \pi^{\dagger}) =  \mathbb{E}_{g_{1:T}}\left\{\sum_{t=1}^T \left[ \frac{\ell_f + 2\cdot\ell_T + \ell_S\cdot D_{v}}{2} \|\tilde{x}^v_t - x_t^{v,\dagger}\|^2 -(\sqrt{1+\lambda} - 1)^2 \cdot  \text{cost}_{v,t}^\dagger  \right]^+\right\} $. By optimally setting $\lambda_2 = \sqrt{  \frac{\sum_{v \in \mathcal{V}} \omega_v (\lambda, \tilde{\pi}, \pi^{\dagger})}{{AVG}(\tilde{\pi})}} $

\end{proof}

\subsection{Proof of Corollary~\ref{cor:avg_cost_blackbox_direct}}

    To prove the Corollary~\ref{cor:avg_cost_blackbox_direct}, we first show the robustness is also guaranteed under the setting of directed graphs. Secondly, we will provide the upper bound of the distance between actual action taken by \ouralg and ML advice. Finally, we translate the action distance to the cost increase associated with the projection. Since most of the proof steps are similar to the problem with undirected graphs, we only highlight the difference here.

    Based on Lemma \ref{lemma:spatial_reserve} and $\kappa_{t-1}^{(v,u)} = \frac{\|x_{t-1}^v - x_{t-1}^{v,\dagger} \|^2}{\|x_{t-1}^v - x_{t-1}^{v,\dagger} \|^2 + \| x_{t-1}^u - x_{t-1}^{u,\dagger} \|^2}$, we have 
    \begin{align}
        \sum_{(v,u) \in \mathcal{E}}\kappa_{t-1}^{(v, u)} \cdot \left({s_{t-1}^{(v, u)}(x_{t-1}^{v}, x_{t-1}^u)} - (1+\lambda)s_{t-1}^{(v, u)}(x_{t-1}^{v,\dagger}, x_{t-1}^{u,\dagger}) \right) &\leq \frac{D_v^{out} \ell_S}{2}(1+\frac{1}{\lambda})  \|x_{t-1}^{v} - x_{t-1}^{v, \dagger} \|^2\\
        \sum_{(u,v) \in \mathcal{E}}\kappa_{t-1}^{(u,v)} \cdot \left({s_{t-1}^{(u,v)}(x_{t-1}^{u}, x_{t-1}^{v})} - (1+\lambda)s_{t-1}^{(u,v)}(x_{t-1}^{u,\dagger}, x_{t-1}^{v,\dagger}) \right) &\leq \frac{D_v^{in} \ell_S}{2}(1+\frac{1}{\lambda})  \|x_{t-1}^{v} - x_{t-1}^{v, \dagger} \|^2
    \end{align}
    By plugging the inequality back to Eqn~\eqref{eqn:robust_proof_eqn2}, we can prove that if action constraint in Eqn~\eqref{eqn:directed_graph_constraint} is satisfied up to time $t-1$, the expert action $x_t^{v,\dagger}$ is a feasible solution for the directed graph $\mathcal{G}$. In other words, the action set constructed by the constraint in Eqn~\eqref{eqn:directed_graph_constraint} is nonempty for all $t\in [1,T]$.

    If the constraint in Eqn.~\eqref{eqn:directed_graph_constraint} is satisfied at time $t-1$, a sufficient condition of the constraint at time $t$ is formulated as 
    \begin{equation}\label{eqn:directed_suff_1}
        \begin{gathered}
          (1+\frac{1}{\lambda_0})\left( \frac{\ell_f + 2\cdot\ell_T + \ell_S\cdot( D_{v}^{in} + D_{v}^{out} )}{2}  \|x_t^{v} - x_t^{v, \dagger} \|^2 \right)  \leq (\lambda - \lambda_0) \Bigl( f_t^{v}(x_t^{v, \dagger})  + \\
          c_t ^{v}(x_t^{v, \dagger}, x_{t-1}^{v, \dagger}) + \sum_{(v, u) \in \mathcal{E}} \kappa_{t-1}^{(v,u)} \cdot {s_{t-1}^{(v, u)}(x_{t-1}^{v, \dagger}, x_{t-1}^{u, \dagger})} + \sum_{(u,v) \in \mathcal{E}} \kappa_{t-1}^{(u,v)} \cdot {s_{t-1}^{(u,v)}(x_{t-1}^{u, \dagger}, x_{t-1}^{v, \dagger})} \Bigr)
        \end{gathered}
    \end{equation}
    
    By defining ${cost}_{v,t}^\dagger = f_t^{v}(x_t^{v, \dagger}) + c_t ^{v}(x_t^{v, \dagger}, x_{t-1}^{v, \dagger})$ as the sum of expert's node cost and temporal cost for node $v$ at time $t$,  a sufficient condition of Eqn~\eqref{eqn:directed_suff_1} is 
    \begin{equation}\label{eqn:direct_sufficient_final}
        \begin{gathered}
            {\|x_t^{v} - x_t^{v, \dagger} \|^2}       \leq \frac{\lambda - \lambda_0}{1+\frac{1}{\lambda_0}} \cdot \frac{2}{\ell_f + 2\cdot\ell_T + \ell_S\cdot (D_{v}^{in} + D_{v}^{out})} \cdot \text{cost}_{v,t}^\dagger
        \end{gathered}
    \end{equation}
    By substituting the action distance to Eqn~\eqref{eqn:distance_single_lado} and summing up the distance over the horizon $t$ and the entire graph, we obtain the total action distance between \ouralg and ML policy in the directed graph. Then we finish the proof by translating the distance to the cost increase ncaccording to the smoothness assumption of these three costs (Assumption \ref{assumption:node_cost} - \ref{assumption:spatial_cost})

\subsection{Proof of Theorem~\ref{theorem:robust_consistency_lado}}\label{sec:proof_robust_consistency_lado}

Denote $\ell=\frac{\ell_f + 2\cdot\ell_T + \ell_S\cdot D_{\max}}{2}$. By \eqref{eqn:consistency_proof_3}, when ML gives the offline-optimal actions, i.e. $\tilde{x}_{1:T}=x_{1:T}^*$, we have for any sequence $g_{1:T}$,
\begin{equation}\label{eqn:consistency_proof_5}
    \begin{aligned}
        & \text{cost} ({x}_{1:T})  -  (1+ \lambda_2) \text{cost} (x_{1:T}^*) \\
    \leq &  (1+\frac{1}{\lambda_2}) \ell \sum_{v\in\mathcal{V}}\sum_{t=1}^T \left[ \|x^{v,*}_t - x_t^{v,\dagger}\|^2 - \frac{1}{\ell} \cdot (\sqrt{1+\lambda} - 1)^2 \cdot  \text{cost}_{v,t}^\dagger  \right]^+
    \end{aligned}
\end{equation}

By optimally setting $\lambda_2$, we have
\begin{equation}
    \begin{aligned}
        \text{cost} ({x}_{1:T}) &\leq  \left(\sqrt{\text{cost} (x_{1:T}^*)} +  \sqrt{\ell \sum_{v\in\mathcal{V}}\sum_{t=1}^T \left[ \|x^{v,*}_t - x_t^{v,\dagger}\|^2 - \frac{1}{\ell} \cdot (\sqrt{1+\lambda} - 1)^2 \cdot  \text{cost}_{v,t}^\dagger  \right]^+}\right)^2
    \end{aligned}
\end{equation}
which translates to a competitive ratio of 
\begin{equation}
\rho_{\ouralg}\leq \min \left\{\left[1+ \sqrt{\max_{g_{1:T}\in\mathcal{G}}   \frac{\sum_{v\in\mathcal{V}}\sum_{t=1}^T \left[ \ell\|x^{v,*}_t - x_t^{v,\dagger}\|^2 -  (\sqrt{1+\lambda} - 1)^2  \cdot {cost}_{v,t}^\dagger  \right]^+}{cost(x^*_{1:T},g_{1:T})}}\right]^2, (1+\lambda)\rho_{\pi^\dagger} \right\}
\end{equation}.

By $\beta-$strongly convexity of the cost function, we have $\nabla cost(x_{1:T}^*)=0$ and
\begin{equation}
cost(\pi^{\dagger})\geq cost(\pi^*) + \frac{\beta}{2} \sum_{v\in\mathcal{V}}\sum_{t=1}^T\|x_t^{v,\dagger}-x_t^{v,*}\|^2.
\end{equation}
Thus, the competitive ratio can be simplified as
\begin{equation}
\begin{split}
\rho_{\ouralg}&\leq \left(1+ \sqrt{\max_{g_{1:T}} 2\ell \frac{\sum_{v\in\mathcal{V}}\sum_{t=1}^T\|x_t^{v,\dagger}-x_t^{v,*}\|^2}{cost(x^*_{1:T},g_{1:T})}}\right)^2\\
&\leq \left(1+ \sqrt{\max_{g_{1:T}} \frac{4\ell}{\beta} \frac{cost(x^{\dagger}_{1:T},g_{1:T})-cost(x^*_{1:T},g_{1:T})}{cost(x^*_{1:T},g_{1:T})}}\right)^2\\
&\leq \left(1+ 2\sqrt{\frac{\ell}{\beta}\cdot (\rho_{\pi^{\dagger}}-1)}\right)^2. 
\end{split}
\end{equation}

\subsection{Proof of Corollary~\ref{thm:avg_cost_e2e}}\label{sec:cost_ratio_e2e_proof}
 In Corollary~\ref{thm:avg_cost_e2e}, we assume that $\pi_{\lambda}^{\circ}$ in Eqn.~\eqref{eqn:optimal_constrained_ML_policy} is used in \ouralg. 
To bound the average cost of $\ouralg(\tilde{\pi}_{\lambda}^{\circ})$, we construct a policy that satisfies the constraint \eqref{eqn:local_constraint} for each step in each sequence. Then the average cost bound of the constructed policy is also the average cost upper bound of $\ouralg(\tilde{\pi}_{\lambda}^{\circ})$ since $\ouralg(\tilde{\pi}_{\lambda}^{\circ})$ is the policy that minimizes average cost while satisfying the constraint \eqref{eqn:local_constraint} for each step in each sequence if we assume that the ML model can represent any policy. 
The feasible policy is constructed as $\hat{\pi}=(1-\alpha)\pi^{\dagger}+\alpha\tilde{\pi}^*$ which gives action $\hat{x}_t^v = (1-\alpha) x_t^{v,\dagger} + \alpha \tilde{x}_t^v,  \alpha \in [0,1]$, where $x_t^\dagger$, $\tilde{x}_t$ denotes expert action and the prediction from 
projection-unaware ML model $\tilde{\pi}^*$, respectively. We need to find the $\alpha$ that guarantees the satisfaction of the constraint \eqref{eqn:local_cost_agent_v}. To do that, we rewrite the constraint as 
\begin{equation}\label{eqn:sufficient_e2e_1}
    \begin{aligned}
        &\sum_{\tau=1}^t \Bigl(f_\tau^v(\hat{x}_\tau^v) - (1+ \lambda_0)f_\tau^v(x_\tau^\dagger)\Bigr) +
        \sum_{\tau=1}^t \left(c^v(\hat{x}_\tau^v, \hat{x}_{\tau-1}^v) - (1+ \lambda_0)c^v(x_\tau^{v,\dagger}, x_{\tau-1}^{v,\dagger})\right) + \sum_{\tau=1}^{t-1}\sum_{(v, u) \in \mathcal{E}} \kappa_\tau^{(v,u)} \cdot \\
        &\Bigl( {s_\tau^{(v, u)}(\hat{x}_\tau^{v}, \hat{x}_\tau^u)} -  (1+ \lambda_0){s_\tau^{(v, u)}(x_\tau^{v, \dagger}, x_\tau^{u, \dagger} )}\Bigr) + (1+\frac{1}{\lambda _0})\frac{\ell_T + D_v\cdot \ell_S}{2}\|x_t^v - x_t^{v,\dagger} \|^2  \\
        \leq&  (\lambda - \lambda_0) \Bigl( \sum_{\tau=1}^t f_\tau^v({x}_\tau^{v, \dagger}) + \sum_{\tau=1}^t c^v(x_\tau^{v,\dagger}, x_{\tau-1}^{v,\dagger}) + \sum_{\tau=1}^{t-1} \sum_{(v, u) \in \mathcal{E}} \kappa_\tau^{(v,u)} s_\tau^{(v, u)}(x_\tau^{v, \dagger}, x_\tau^{u, \dagger} ) \Bigr),  \forall t \in [1, T]
    \end{aligned}
\end{equation}

Based on the smoothness assumption, we have 
\begin{equation}
    \begin{aligned}
         f_t^v(\hat{x}_t^v) - (1+ \lambda_0)f_t^v(x_t^{v,\dagger}) \leq & (1+ \frac{1}{\lambda_0})\frac{\ell_f}{2} \| \hat{x}_t^v - x_t^{v,\dagger} \|^2\\
        c^v(\hat{x}_t^v, \hat{x}_{t-1}^v) - (1+ \lambda_0)c^v(x_t^{v,\dagger}, x_{t-1}^{v,\dagger})  \leq & (1+ \frac{1}{\lambda_0}) \frac{\ell_T}{2} (\| \hat{x}_t^v - x_t^{v,\dagger} \|^2 + \| \hat{x}_{t-1}^v - x_{t-1}^{v,\dagger} \|^2)\\
        \kappa_t^{(v,u)}\left({s_t^{(v, u)}(\hat{x}_t^{v}, \hat{x}_t^u)} - (1+ \lambda_0){s_t^{(v, u)}(x_t^{v, \dagger}, x_t^{u, \dagger} )} \right) \leq &  (1+ \frac{1}{\lambda_0}) \frac{\ell_S}{2} (\| \hat{x}_t^v - x_t^{v,\dagger} \|^2 )
    \end{aligned}
\end{equation}
Then, a sufficient condition of Eqn~\eqref{eqn:sufficient_e2e_1} is 
\begin{equation}\label{eqn:sufficient_e2e_2}
    \begin{gathered}
        (1+\frac{1}{\lambda_0})\frac{\ell_f + 2 \ell_T + D_{v} \ell_S}{2}  \sum_{\tau=1}^t \| \hat{x}_\tau^v - x_\tau^{v,\dagger} \|^2  \leq (\lambda - \lambda_0) \text{cost}_{v}(x_{1:t}^{v,\dagger}) , \forall t \in [1, T]
    \end{gathered}
\end{equation}
Since $D_{\max} = \max_{v\in \mathcal{V}} D_v $ is the maximum node degree in the whole graph, then the sufficient condition becomes
\begin{equation}
     \alpha^2 \sum_{\tau=1}^t \| \tilde{x}_\tau^v - x_\tau^{v,\dagger} \|^2  \leq \frac{2}{\ell_f + 2 \ell_T + D_{\max} \cdot\ell_S} \cdot \frac{\lambda - \lambda_0}{1+\frac{1}{\lambda_0}} \text{cost}_{v}(x_{1:t}^{v,\dagger}) , \forall t \in [1, T]
\end{equation}

We define $\hat{C} = \min_{v \in \mathcal{V}, t \in [1,T]} \frac{\text{cost}_{v}(x_{1:t}^{v,\dagger})}{\sum_{i=1}^t \|x_i^{v,\dagger} - \tilde{x}_i^{v} \|^2}$ as the minimum normalized baseline cost, then we can have
\begin{equation}
    \alpha    \leq \min\left\{1,\sqrt{\frac{2}{\ell_f + 2 \ell_T + D_{\max} \cdot \ell_S} \cdot \frac{\lambda - \lambda_0}{1+\frac{1}{\lambda_0}}\cdot \hat{C}} \right\} = \alpha_\lambda
\end{equation}
In other words, as long as $\alpha \in [0, \alpha_\lambda]$, the robustness constraint is always satisfied. Based on the convex assumption on hitting cost, temporal cost and spatial cost, we have
\begin{equation}
    \text{cost}_{v}(\hat{x}_{1:t}^{v})  =  \text{cost}_{v}((1-\alpha){x}_{1:t}^{v, \dagger} + \alpha \tilde{x}_{1:t}^{v, \dagger} ) \leq (1-\alpha)\text{cost}_{v}({x}_{1:t}^{v, \dagger})  + \alpha \cdot\text{cost}_{v}(\tilde{x}_{1:t}^{v}).
\end{equation}
By setting $\alpha = \alpha_\lambda$ and taking expectation of both side over the data distribution, we finish the proof of the first term in Theorem~\ref{thm:avg_cost_e2e}. \qed

\end{document}